\documentclass[10pt,twocolumn,letterpaper]{article}

\usepackage[pagenumbers]{cvpr} 

\usepackage{currfile} 

\usepackage{caption} 
\usepackage{cuted}

\usepackage{bm}

\newcommand{\synvaBase}{SynVA-V1\textsubscript{base}}      
\newcommand{\synvaDepth}{SynVA-V1\textsubscript{depth}}    
\newcommand{\synvaAlign}{SynVA-V1\textsubscript{aligned}}  

\newcommand{\synvaCombined}{SynVA-V1\textsubscript{combined}} 
\newcommand{\synvaPhys}{SynVA-V1\textsubscript{phys}}         
\newcommand{\synvaPhysTx}{SynVA-V1\textsubscript{phys-Tx}}    

\makeatletter
\let\SuppOrigMakeTitle\@maketitle   
\let\SuppOrigAuthor\author
\let\SuppOrigAnd\and                
\renewcommand{\author}[1]{\SuppOrigAuthor{#1}\gdef\SuppAuthor{#1}}
\newcount\SuppLineOffset
\newcommand{\SupplementaryReferences}{%
  \clearpage
  \setcounter{page}{1}%
  \renewcommand{\thepage}{S\arabic{page}}%
  \def\@title{{\normalfont\large Supplementary Material\par}\vspace{0.5em}\thetitle}%
  \let\@author\SuppAuthor
  \let\and\SuppOrigAnd
  \twocolumn[\SuppOrigMakeTitle]%
  \@ifundefined{c@linenumber}{}{%
    \global\SuppLineOffset=\c@linenumber
    \renewcommand\thelinenumber{%
      \fillzeros[3]{\numexpr\c@linenumber-\SuppLineOffset\relax}}}%
  \setcounter{section}{0}\setcounter{figure}{0}%
  \setcounter{table}{0}\setcounter{equation}{0}%
  \renewcommand{\thesection}{S\arabic{section}}%
  \renewcommand{\thefigure}{S\arabic{figure}}%
  \renewcommand{\thetable}{S\arabic{table}}%
  \renewcommand{\theequation}{S\arabic{equation}}%
  \renewcommand{\theHsection}{supp.\arabic{section}}%
  \renewcommand{\theHfigure}{supp.\arabic{figure}}%
  \renewcommand{\theHtable}{supp.\arabic{table}}%
  \renewcommand{\theHequation}{supp.\arabic{equation}}%
}
\makeatother

\definecolor{cvprblue}{rgb}{0.21,0.49,0.74}
\usepackage[pagebackref,breaklinks,colorlinks,allcolors=cvprblue]{hyperref}

\def\confName{CVPR}
\def\confYear{2027}

\title{SynVA: A Modular Toolkit for Vessel Generation and Aneurysm Editing}

\author{
Marten J. Finck$^{1*}$ \quad Niklas C. Koser$^{1*}$ \quad Sarker M. Mahfuz$^{1}$ \quad Tameem Jahangir$^{2}$ \quad Jon E. Wilhelm$^{1}$\\
Daniel Behme$^{3}$ \quad Naomi Larsen$^{4}$ \quad Wojtek Palubicki$^{5}$ \quad Sylvia Saalfeld$^{2}$ \quad Sören Pirk$^{1}$\\[1ex]
{\small $^1$Visual Computing and Artificial Intelligence, Kiel University, Germany}\\
{\small $^2$Institute for Medical Informatics and Statistics, Kiel University, Germany}\\
{\small $^3$Clinic for Neuroradiology, Medical Faculty -- Magdeburg University, Germany}\\
{\small $^4$Department of Radiology and Neuroradiology, University Hospital Schleswig-Holstein, Kiel, Germany}\\
{\small $^5$Faculty of Mathematics and Computer Science, Adam Mickiewicz University, Poznan, Poland}\\[1ex]
{\tt\small \{mafi, niklas.koser, smm, sp\}@informatik.uni-kiel.de, eike.wilhelm@stu.uni-kiel.de,}\\
{\tt\small saalfeld@medinfo.uni-kiel.de, \{tameem.jahangir, naomi.larsen\}@uksh.de,}\\
{\tt\small daniel.behme@med.ovgu.de, wojciech.palubicki@amu.edu.pl}\\[0.5ex]
{\small $^*$Equal contribution}
}

\begin{document}
\maketitle
\begin{strip}
\centering
\includegraphics[width=1.0\linewidth]{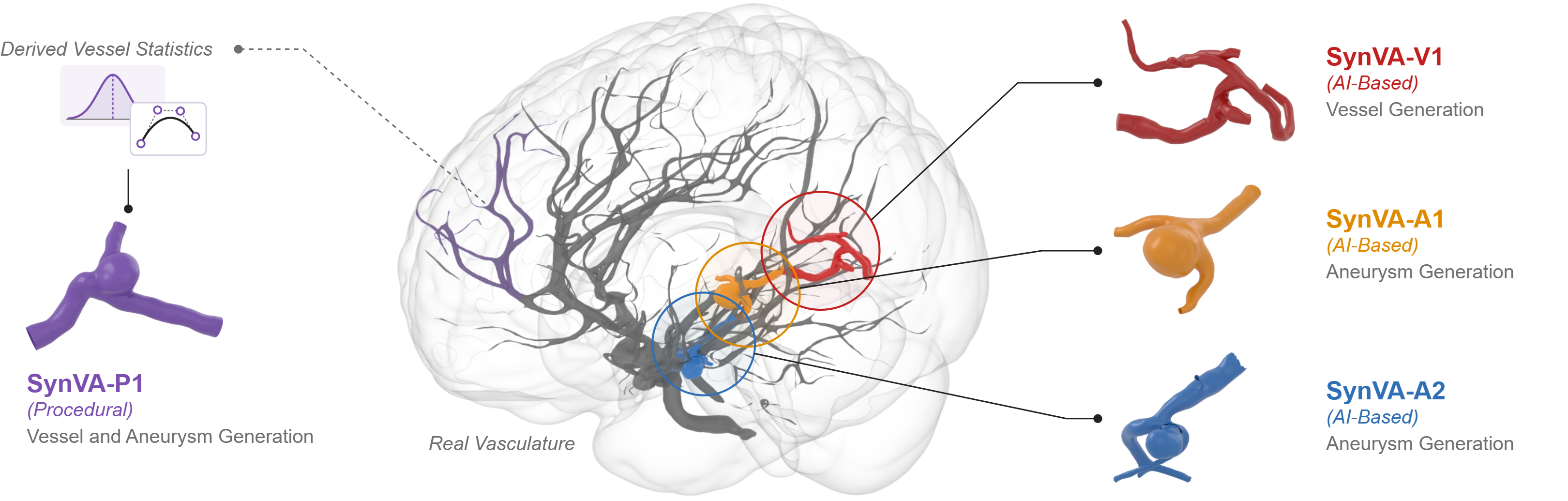}
\captionsetup{hypcap=false}
\captionof{figure}{\textbf{Synthetic vessels and aneurysms generated with SynVA.} Learning-based models generate healthy vessel trees (SynVA-V1) and aneurysms on a given vessel (SynVA-A1, SynVA-A2). Colored circles mark corresponding regions in a real cerebral vasculature. The fully procedural model SynVA-P1 synthesizes labeled vessels with aneurysms from physiological principles and statistical priors.}
\captionsetup{hypcap=true}
\label{fig:teaser}
\end{strip}
\begin{abstract}
Intracranial aneurysms~(IAs), characterized by unpredictable growth and risk of rupture, are a major cause of stroke and can lead to life-threatening hemorrhages with high mortality and long-term disability. With aging populations, the incidence and overall burden of cerebrovascular diseases are expected to increase, highlighting the need for scalable approaches to analyze complex medical data and improve population-level understanding of these conditions. While digital twins and deep learning offer promising avenues for improving diagnosis, prognosis, and treatment, their effectiveness is limited by the scarcity of large-scale, high-quality medical data and corresponding labels. 
We present \textbf{Syn}thetic \textbf{VA}sculature~(\textbf{SynVA}), a modular toolkit for vascular mesh generation and anatomically consistent aneurysm synthesis. 
SynVA combines novel flow-matching-based methods for generating healthy vessel meshes with learning-based approaches for anatomy-conditioned aneurysm mesh generation; it computes aneurysms from pre-existing vascular geometries rather than generating them in isolation. In addition, we introduce the SynVA procedural model for vascular and aneurysm synthesis based solely on physiological principles and statistical priors, which enables the generation of large-scale datasets.
To this end, we release a dataset of $50{,}000$ fully labeled mesh samples for a variety of downstream vision tasks (\href{https://www.kaggle.com/datasets/mrtnfi/synva-procedural-dataset}{Link}). 
Extensive quantitative and qualitative evaluations show that SynVA generates realistic vessel geometries and anatomically plausible aneurysms, with only minor trade-offs between expert-perceived realism and similarity to real aneurysms. Moreover, pretraining on SynVA data improves point-cloud segmentation and morphological parameter estimation on real data. Code will be released upon acceptance.
\end{abstract}  
\section{Introduction}
\label{sec:Introduction}
Intracranial aneurysms (IAs) are acquired pathological dilatations of cerebral arteries that arise from structural alterations of the vessel wall and a resulting imbalance between tensile wall strength and local hemodynamic stress~\cite{schievink1997intracranial,forsting2006intracranial,ringer2018intracranial,frosen2012saccular}.
IAs are typically asymptomatic and often remain undetected, with only $10$--$15\,\%$ of patients experiencing neurological symptoms~\cite{schievink1997intracranial,forsting2006intracranial,keedy2006overview}. Nevertheless, their unpredictable growth and potential for rupture complicate clinical risk assessment and treatment decisions~\cite{ringer2018intracranial,keedy2006overview}. The estimated prevalence of IAs is $5.4\,\%$~\cite{quan2025prevalence}, while approximately $8$--$10$ ruptures occur annually per $100{,}000$ individuals~\cite{ia_statistics_facts}. Rupture is associated with mortality rates of $25$--$50\,\%$, and $50$--$66\,\%$ of survivors experience long-term neurological deficits~\cite{keedy2006overview,frosen2012saccular,ia_statistics_facts}. Given these severe consequences, early detection and accurate risk stratification are essential for guiding patient-specific treatment decisions.
Risk factors for IA formation and rupture include both aneurysm-specific and patient-specific characteristics. These comprise morphological parameters~(e.g., size, aspect ratio, irregularity), anatomical location, hypertension, age, sex (with higher prevalence in females), geographic and ethnic factors, prior SAH, family history, and smoking status~\cite{schievink1997intracranial,ringer2018intracranial,frosen2012saccular,dhar2008morphology,mocco2018aneurysm}.\\
\indent In this context, digital twins (DTs), i.e., virtual counterparts of real anatomical structures that combine statistical, data-driven, and mechanistic models to reflect physiological behavior and dynamics, offer a promising route toward more precise patient-specific prognosis and treatment planning~\cite{sun2023digital,erol2020digital,vallee2023digital}. Since both hemodynamics and aneurysm geometry, individually and in combination, are highly relevant for the dynamic behavior of IAs, there is growing interest in in-silico studies using 3D DTs of vascular systems with IAs, for instance to estimate wall shear stress (WSS) via computational fluid dynamics (CFD)~\cite{schievink1997intracranial,ringer2018intracranial,dhar2008morphology}. 
In parallel, deep learning approaches for accelerated flow prediction or IA segmentation are gaining importance~\cite{sheng2026geometry,li2021prediction,schneider2021medmeshcnn}, further underscoring the need for large, labeled 3D datasets of IAs and their parent vessels.
However, real-world datasets remain limited in size~\cite{juchler2022aneux,yang2020intra,song2024cmha,li2025aneumo,crown_data_2023,de2024time}, and reconstruction from medical imaging data is challenging, requires expert knowledge, and is prone to errors~\cite{lan2018re,mou2024costa,alblas2022deep}. As a result, synthetic 3D data generation is increasingly viewed as a viable strategy to provide labeled data for downstream applications while improving not only quantity but also diversity~\cite{wu2013comparative,van2024synthetic}. Yet, model-driven procedural approaches remain scarce, and existing methods often still rely on real data~\cite{nader2025building,decroocq2023modeling,kim2023tissue,rundfeldt2024cerebral}. More recent data-driven AI methods for vascular synthesis tend to focus either on aneurysm generation~\cite{ding2025two} or on vessel generation~\cite{feldman2025vesselgpt}, but not on a unified end-to-end framework.
Motivated by these challenges, we introduce \textbf{Syn}thetic \textbf{VA}sculature (\textbf{SynVA}), a modular toolkit for the generation of anatomically plausible vascular meshes and subsequently evolving aneurysm geometries, following the physiological formation process in which aneurysms emerge from pre-existing vessel structures~(\cref{fig:teaser}). SynVA supports both data-driven learning-based models and a model-driven procedural approach for synthesizing its two core components: healthy vessel structures and aneurysm geometries. The coupling between vessel generation and aneurysm formation is established via the selection of an ostium, defined as the intersection between the parent vessel and the aneurysm. 
It is placed either manually for controlled editing or automatically for large-scale data generation.
Beyond its clinical motivation, SynVA addresses problems of broad interest to 3D vision: generating structured 3D data with explicit topology and dense part labels, continuing a mesh from a given boundary under partial geometric context, and transferring knowledge from synthetic to real 3D data when annotations are scarce.\\
\indent In summary, SynVA's key contributions are as follows:\\
\textbf{(1)}~A two-stage healthy vessel generation model with autoregressive topology and flow-matching geometry that outperforms three published baselines in maximum mean discrepancy~(MMD), coverage~(COV), and $1$-nearest-neighbor accuracy~(1-NNA) while generating novel anatomies;
\textbf{(2)}~Two methodologically distinct, ostium-conditioned learning-based models that, for the first time, enable both latent-shape (AneuG~\cite{ding2025two}) and autoregressive (MeshAnything V2~\cite{chen2025meshanythingv2}) mesh generation to synthesize aneurysms on existing vessels, turning whole-shape generation into boundary-conditioned continuation of an existing mesh from partial geometric context;
\textbf{(3)}~A first-of-its-kind procedural model for generating labeled vascular meshes and aneurysms based solely on biological and mathematical principles;
\textbf{(4)}~A large-scale synthetic dataset of 3D vessel structures with aneurysms comprising $50{,}000$ samples;
\textbf{(5)}~A synthetic-to-real study showing that SynVA pretraining improves point-cloud segmentation and morphological parameter estimation on real data in low-data regimes.
\section{Related work}
\label{sec:RelatedWork}
\textbf{Procedural Vessel Generation}
Optimization-based approaches minimize a hemodynamically motivated cost function. Most extend Constrained Constructive Optimization~(CCO) with a local per-branch optimization~\cite{karch2000staged,talou2021adaptive,sexton2025rapid,kim2023tissue,hamarneh2010vascusynth}, whereas \citet{jessen2025optimizing} formulate it globally as a nonlinear program. Angiogenesis-based models instead simulate vascular growth as a response to diffusing biochemical signals, reproducing the biological mechanism rather than only its outcome~\cite{milde2008hybrid,schneider2012tissue}. Rule-based and L-system-inspired methods provide procedural control over branching structure, guided by Murray-type assumptions and spatial constraints~\cite{galarreta2013three,ritraksa20213d,rauch2021interactive}. Synthetic approaches generate vascular structures for segmentation, pretraining, or aneurysm-detection tasks~\cite{anso2020synthetic,nader2025building,hamarneh2010vascusynth}. However, existing methods either focus on large-scale vascular trees without aneurysmal pathology~\cite{karch2000staged,talou2021adaptive,jessen2025optimizing,sexton2025rapid,milde2008hybrid,schneider2012tissue,kim2023tissue,hamarneh2010vascusynth,anso2020synthetic}, or model aneurysms only at the voxel/image level rather than as explicit, labeled mesh structures~\cite{nader2025building}. To our knowledge, no prior work produces fully procedural, high-resolution, explicitly labeled meshes of aneurysmal vessels.

\textbf{Mesh Generative AI}
Early autoregressive methods such as PolyGen~\cite{nash2020polygen} model vertices and faces separately, while MeshGPT~\cite{siddiqui2024meshgpt} introduces a tokenization-based approach, where triangle meshes are encoded into sequences of quantized latent embeddings. A Transformer is then trained to autoregressively predict the next token, from which the full mesh can be decoded. This paradigm has been widely adopted and extended by subsequent works, often incorporating alternative architectures and conditioning mechanisms~\cite{chen2024meshxl,wang2024llama,weng2025pivotmesh}. Beyond model design, recent research also focuses on improving mesh representations and tokenization strategies. For instance, MeshAnything~V2~\cite{chen2025meshanythingv2} leverages an adjacent mesh tokenization~(AMT) scheme to significantly increase the number of generatable faces, addressing a key limitation in high-resolution mesh synthesis. Other approaches explore specialized traversal and compression strategies to maximize representation efficiency~\cite{song2026mesh,tang2025edgerunner,weng2025scaling}. More recent models such as MeshTron~\cite{hao2024meshtron}, DeepMesh~\cite{zhao2025deepmesh}, FastMesh~\cite{kim2025fastmesh}, and MeshRipple~\cite{lin2026meshripple} further advance mesh generation.

\textbf{Vascular Mesh Generation}
Generative models for vascular structures can be broadly grouped into topology-oriented approaches, geometry-oriented approaches, and methods that explicitly separate global structure from local geometry. Early work such as VesselVAE \cite{feldman2023vesselvae} generates vascular trees from a hierarchical tree representation, but remains restricted to tree-like topologies and mainly captures coarse morphology through descriptors such as radius, length, and tortuosity. Subsequent topology-oriented methods extend this idea to more general structural representations \cite{prabhakar20243d,kuipers2024generating,batten2025vector,mo2026anatomically}. However, these methods primarily operate at the level of structure, centerlines, or topology rather than directly modeling detailed surface geometry.
In parallel, geometry-oriented methods focus on continuous shape representations. TrIND \cite{sinha2024trind} models anatomical trees with implicit neural fields, while the semantic signed distance field approach of \citet{kuipers2025self} directly generates surface representations of cerebral vascular trees. VesselGPT \cite{feldman2025vesselgpt} further highlights the importance of local cross-sectional modeling through an autoregressive formulation based on B-spline cross-sections. Relatedly, \citet{deo2024few} generate aneurysm-bearing vessel segmentations under few-shot conditions using latent diffusion with SDF-based conditioning. The closest work to our setting is the part-based model of \citet{chen2025hierarchical}, which decomposes generation into a global key-graph, local segment geometry, and assembly; VesselTok \cite{prabhakar2026vesseltok} similarly separates structural and geometric representation learning for vessel-like biomedical graphs. Nevertheless, existing methods either focus on structural generation without precise local surface control, or on continuous geometry without an equally explicit graph-based decomposition. 

\textbf{Aneurysm Mesh Generation}
Generative AI models for IA synthesis are motivated by the scarcity of large, high-quality annotated datasets and, more recently, by the need for simulation-ready aneurysm geometries for downstream hemodynamic analysis \cite{deo2024few,ding2025two,ding2025aneug,li2025aneumo}. Existing approaches mainly differ in whether they generate aneurysm-bearing segmentations, complete vascular geometries, or explicit 3D meshes. 
The closest methodological reference is AneuG \cite{ding2025two}, a two-stage VAE-based model that generates IA meshes by first synthesizing the aneurysm complex and then modeling the parent vessels conditioned on it, with additional control via morphological markers such as aspect ratio, neck width, lobulation index, and volume.
Despite these advances, existing methods either generate aneurysm-bearing segmentations without direct surface-level control or synthesize complete aneurysm--parent-vessel complexes without explicitly modeling the ostium as the primary geometric constraint. This leaves a gap in modeling aneurysm formation as a localized, surface-consistent process on an existing healthy vessel mesh, which SynVA addresses.  
\section{Methodology}
\label{sec:Methodology}

\begin{figure*}[t]
  \centering
  \includegraphics[width=0.8\textwidth]{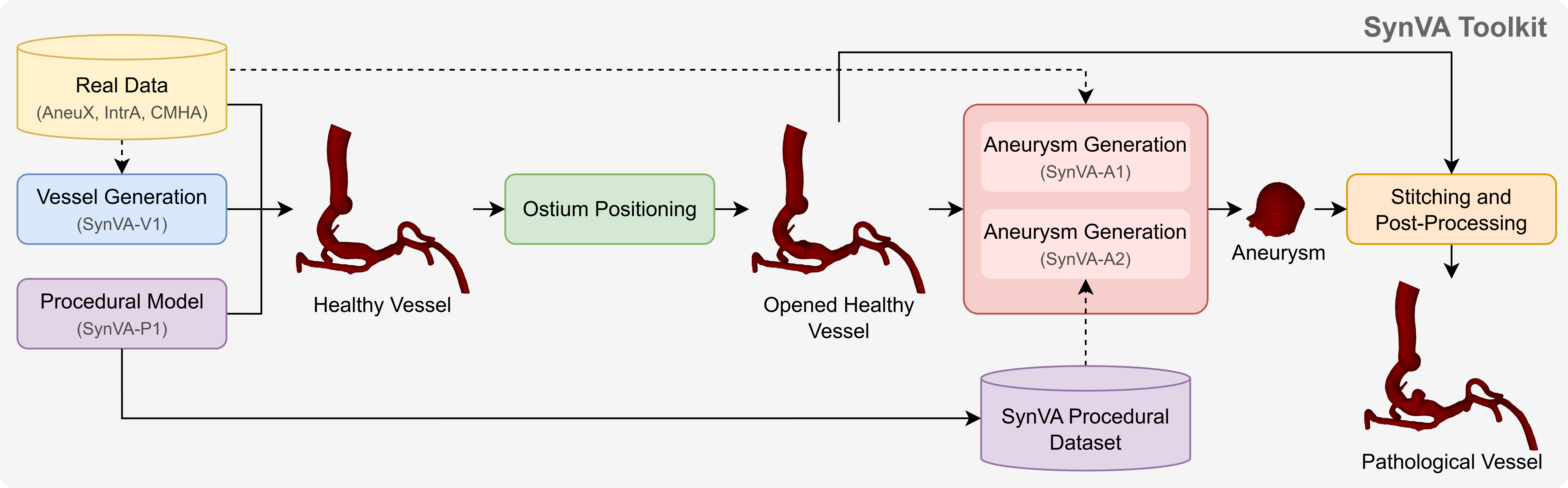}
  \caption{Overview of the SynVA toolkit. SynVA generates pathological vascular meshes by combining healthy vessel generation, ostium selection, and conditioned aneurysm synthesis.}
  \label{fig:overview}
\end{figure*}

SynVA is a modular toolkit (\cref{fig:overview}) for generating fully labeled 3D vascular meshes, including both healthy and pathological geometries with aneurysms. It integrates data-driven and model-driven paradigms within a unified toolkit. On the learning side, an autoregressive topology model and a flow-matching geometry model learn healthy vascular structures~(SynVA-V1). Additionally, we developed the SynVA-P1 procedural model for vessel structures based on physiological principles and statistical priors to generate vascular structures with and without aneurysms. A healthy vessel structure is used to select the position of an ostium -- the ring around an aneurysm opening. 
We then introduce SynVA-A1 and SynVA-A2, which reformulate AneuG~\cite{ding2025two} and MeshAnything V2~\cite{chen2025meshanythingv2} to add aneurysms on existing vessel meshes without full-shape conditioning, substantially extending both models toward mesh continuation.
In a postprocessing step, the generated aneurysm mesh is stitched onto the healthy vessel structure to form a complete pathological vessel with an aneurysm. The following sections describe these components together with the associated interfaces for single-sample generation, editing, and large-scale dataset synthesis. The main focus of this work lies on the learning-based approaches, while SynVA-P1 and the resulting synthetic dataset are detailed in \cref{sec:SM_SynVA-P1,sec:SM_Dataset}.

\subsection{Flow-matching based Vessel Generation}
\label{sec:Methodology_SynVA-V1}
SynVA-V1 generates healthy vascular structures in a two-stage process. In the first stage, only the discrete vessel topology is learned as a rooted directed tree, whose edges encode the branching structure, without observing any continuous geometry. In the second stage, the sampled topology is held fixed and used as a structural condition for geometry generation. Given this tree, SynVA-V1 predicts continuous node features, including parent-relative centerline positions, local cross-section control points, and B-spline knot parameters. The generative model uses a topology-aware Graph Neural Network~(GNN) \cite{gilmer2017neural} backbone that we refer to as \textit{TreeFlowNet}. For each vessel tree, a multigraph is constructed with parent-to-child, child-to-parent, and sibling-to-sibling edges. Separate message-passing functions are used for the different edge types, enabling asymmetric information flow along anatomical directions. Local tree-based message passing is combined with global self-attention every third layer, allowing the model to capture both nearby bifurcation structure and long-range dependencies between distant branches.\\
\indent Training is performed using optimal-transport conditional flow matching \cite{lipman2023flow} on the real dataset splits. Since all available real samples contain aneurysms, we derive healthy vessel data by manually removing aneurysmal regions through mesh sculpting, yielding anatomically plausible vessel-only reconstructions. Starting from Gaussian noise, the model learns a velocity field that transports noisy vessel geometries toward real vessel geometries while remaining conditioned on the fixed tree topology. The loss is computed over valid tree nodes and balances the position, cross-section, and spline-parameter components equally. During inference, a new geometry is sampled by integrating the learned velocity field from noise to the data distribution.\\ 
\indent Validation MMD selects \synvaDepth{}, which warps the flow time of each node with its depth, so that proximal vessels settle before distal branches. A logit-normal time sampler, focal position weighting, ring repair, and physiological auxiliary losses do not improve validation MMD (\cref{sec:SM_SynVA-V1_Results_Models}).
After generation, the predicted node features are denormalized and converted into a spline-based vessel representation. The final healthy vessel mesh is reconstructed by sampling tubular point clouds along the spline segments and applying screened Poisson surface reconstruction with local smoothing at bifurcations. More details can be found in \cref{sec:SM_SynVA-V1}.

\subsection{Procedural Vessel and Aneurysm Generation}
\label{sec:Methodology_SynVA-P1}

SynVA-P1 is a procedural model for generating labeled pathological vascular meshes without relying on patient-specific input geometries. The model combines physiological constraints, statistical priors, and stochastic geometric perturbations. It generates a local bifurcating vessel configuration, optionally augmented with an aneurysm, through two stages: healthy vessel generation and aneurysm synthesis. First, a centerline representation of a bifurcating vessel is sampled. The parent diameter is drawn from a predefined range, while the child-branch diameters are derived according to Murray's law, yielding physiologically plausible vessel proportions. A signed distance field is constructed around the centerlines and converted into a surface mesh using Marching Cubes. To avoid overly idealized tube-like shapes, the mesh is perturbed with multi-scale OpenSimplex noise, smoothed, cleaned, and isotropically remeshed. More details about the SynVA-P1 procedural model can be found in \cref{sec:SM_SynVA-P1}.

\subsection{Ostium Positioning}
\label{sec:Methodology_Ostium}
Ostium positioning can be performed either automatically~(see \cref{sec:SM_SynVA-P1_Ostium}) or interactively via the graphical user interface in SynVA (manual ostium positioning, see \cref{sec:SM_GUI}). The automatic approach ensures consistent and scalable placement driven by geometric priors and statistically informed hemodynamic assumptions, making it well suited for large-scale data generation. In contrast, the manual approach enables precise specification, refinement, and cropping of the ostium, which is particularly useful for single-sample generation and targeted editing of existing vessel geometries.
In automatic ostium selection, two complementary strategies are employed. For bifurcation-based placement, the junction is identified from the centerline topology, and the local flow direction of the parent vessel is approximated from nearby centerline tangents. Within a local neighborhood, surface vertices are evaluated using APSS-based curvature estimation~\cite{guennebaud2007algebraic,guennebaud2008dynamic}, and the ostium is defined as the minimally curved vertex aligned with the flow direction. This choice is motivated by the increased hemodynamic stress typically observed in this region, which has been strongly associated with aneurysm formation~\cite{schievink1997intracranial,forsting2006intracranial,ringer2018intracranial}. For curvature-based placement, a curvature score is computed along the centerline to identify high-curvature regions. Surface vertices in the corresponding neighborhood are then evaluated, and the ostium is selected at either maximal (dominant case) or minimal curvature, capturing both outward- and inward-facing configurations observed in clinical data. Further details are provided in \cref{sec:SM_SynVA-P1_Ostium}.

\subsection{Ostium-Conditioned Aneurysm Generation}
\label{sec:Methodology_SynVA-A}

Aneurysm synthesis in SynVA is formulated as a localized, anatomy-conditioned generation task. Instead of generating aneurysms as isolated objects or complete aneurysm--vessel complexes, all three aneurysm generators operate on a predefined ostium on an existing healthy vessel surface. The ostium defines the attachment boundary, local scale, and growth direction of the aneurysm. This formulation reflects the physiological setting in which aneurysms emerge from pre-existing vascular structures and ensures that the generated geometry can be consistently stitched back onto the parent vessel. We consider three complementary approaches for conditioned aneurysm generation: two learning-based models, SynVA-A1 and SynVA-A2, and the procedural model SynVA-P1. 

\textbf{SynVA-A1} builds on AneuG~\cite{ding2025two} and uses a compact GHD-based latent representation of aneurysm morphology. In contrast to the original configuration, where aneurysm and parent-vessel geometry are modeled jointly in a two-stage generative process, we reformulate the model to focus on the aneurysm sac and neck region conditioned on the local ostium geometry. This allows the model to synthesize aneurysm shapes as localized deformations anchored to the vessel surface, while the surrounding healthy vasculature is provided by the input vessel mesh. To enable this formulation, we condition not only on independent morphological parameters but additionally on the explicit ostium ring geometry, dense samples from the segmented ostium attachment region, compact ostium parameters, and local vessel structure. Furthermore, dedicated loss terms that enforce geometric consistency at the vessel--aneurysm interface are introduced, including consistency with the ostium attachment region, the opening center, and the expected side of the ostium plane. In addition to the standard posterior reconstruction path of the conditional VAE, we explicitly regularize the prior sampling path by decoding samples from $z \sim \mathcal{N}(0,I)$ under the same condition and matching them to case-level and batch-level target statistics. The generated aneurysm is subsequently transformed back into the vessel coordinate system and merged with the parent vessel at the ostium. More details can be found in \cref{sec:SM_SynVA-A1}.

\begin{figure}[htb]
  \centering
  \includegraphics[width=\linewidth]{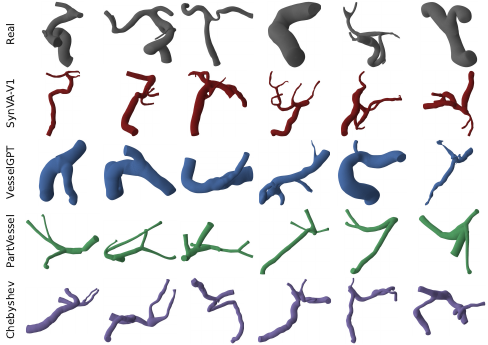}
  \vspace{-6mm}
  \caption{Randomly selected, unconditional samples of healthy vessel meshes from SynVA-V1 and all baseline methods.}
  \label{fig:vessel_structures}
  \vspace{-2mm}
\end{figure}

\textbf{SynVA-A2} extends MeshAnything V2~\cite{chen2025meshanythingv2} from full-shape mesh reconstruction to ostium-conditioned autoregressive aneurysm generation. Since the complete aneurysm geometry is not available at inference time, the model is conditioned only on partial anatomical context: a local vessel point cloud around the ostium and an explicit ostium mesh. The local point cloud provides geometric context about the parent vessel, while the ostium mesh is tokenized and injected as a forced prefix into the autoregressive sequence. The Transformer therefore starts generation from the correct anatomical boundary and predicts the aneurysm mesh as a continuation of the ostium, rather than reconstructing a complete object from a full point cloud. During training, the sequence consists of the ostium prefix followed by the aneurysm continuation, with the loss applied only to the generated aneurysm tokens. More details can be found in \cref{sec:SM_SynVA-A2}.

\textbf{SynVA-P1} provides a procedural model alternative that does not rely on learned shape priors. Given the ostium centroid, ostium normal, and local vessel radius, the aneurysm sac is initialized from a simple geometric primitive, scaled to the vessel size, elongated along the ostium normal, and perturbed with stochastic surface noise to introduce shape variability. Optionally, a secondary protrusion is added to simulate bleb-like irregularities. The resulting aneurysm mesh is merged with the healthy vessel using mesh Boolean operations, followed by remeshing, smoothing, and relabeling. More details can be found in \cref{sec:SM_SynVA-P1}.

For all three approaches, the final step is a post-processing and stitching procedure. Invalid or degenerate mesh elements are removed, normals are recomputed, and the aneurysm is aligned with the ostium in the original vessel coordinate frame. The aneurysm and vessel meshes are then merged at the ostium interface, and the transition region is smoothed to obtain a single pathological vascular mesh. The final output contains a complete surface representation with semantic labels for healthy vessel, aneurysm, and ostium, enabling downstream use in segmentation, morphology analysis, and simulation-oriented workflows.

\subsection{Data and Preprocessing}
\label{sec:Methodology_Data}
We utilize a total of 762 samples from three publicly available datasets containing vascular meshes and IAs. Due to differences in data acquisition, preprocessing pipelines, and mesh representations, each dataset is processed independently to obtain a consistent representation of complete vessels with per-vertex labels. A unified preprocessing pipeline is then applied across all datasets to ensure consistency. This includes normalization of mesh coordinates to the range~$[-1,\,1]$, extraction of labeled point clouds and corresponding sub-point clouds, computation of the ostium centroid and its associated normal vector (oriented toward the aneurysm), submesh extraction, and the computation of relevant morphological parameters (see \cref{sec:SM_MorphologicParameters}). All processed data are stored in a unified and standardized data structure, comprising full meshes, vertex-wise labels, (sub-)point clouds, (sub-)meshes, and additional attributes such as morphological descriptors (see \cref{sec:SM_MorphologicParameters,sec:SM_SynVA-P1}).
Primarily this project uses data from the AneuX morphology database \cite{juchler2022aneux}, an open-access, multi-centric database combining data from three European projects: AneuX project, @neurIST project~\cite{iavindrasana2008neurist,villa2011neurist} and Aneurisk \cite{sangalli2009aneurisk}. It constitutes the largest dataset in our study, comprising 605 patients, 668 vessel trees, and 750 aneurysm meshes with varying cut configurations and mesh resolutions. In addition, clinical and morphological metadata are provided. We use the original mesh resolution and the dome cut configuration (resulting in 564 samples). Following a qualitative inspection, 6 samples were excluded, resulting in a total of 558 samples used in our experiments.
The second dataset is the Intracranial Aneurysm Dataset for Deep Learning (IntrA)~\cite{yang2020intra}, which contains 103 full cerebrovascular models. In addition to automatically generated segmentation meshes, the dataset provides 116 manually annotated segments. After excluding one sample with more than one aneurysm, the remaining annotated segments are incorporated into our standardized data representation.
The third dataset is the CMHA dataset~\cite{song2024cmha}, which includes 99 patients with aneurysms and 44 control subjects, along with morphological and hemodynamic parameters. From this dataset, we use 96 aneurysm cases, excluding three samples with more than one ostium ring.\\
\indent In addition to the real-world datasets, we generate a large-scale synthetic dataset using the SynVA-P1 procedural model~(see \cref{sec:SM_Dataset}). In total, SynVA-P1 is used to create 50,000 pathological vascular meshes with aneurysms. Each sample is generated independently by sampling a bifurcating vessel geometry, selecting an ostium according to the procedural placement strategy, and synthesizing an aneurysm sac.
The same preprocessing used for the real data is then applied to the synthetic samples. Consequently, each synthetic data point contains a complete pathological vessel mesh with vertex-wise labels for vessel, aneurysm, and ostium regions, as well as derived point clouds, submeshes, centerline and radius information, ostium descriptors, and morphological parameters. 
\section{Experiments and Results}
\label{sec:Exp-Results}
For all experiments, we use one train–test split shared by all models. For real data, 100 samples are selected as a stratified test set across the three individual datasets. For the synthetic dataset, 10,000 samples are randomly held out for testing. We train and evaluate SynVA-V1, SynVA-A1, and SynVA-A2 on 2× NVIDIA B200 GPUs (180\,GB VRAM each), 16 cores of an Intel Xeon Platinum 8570 CPU, and Ubuntu~24.04.

\subsection{Healthy Vessel Generation}
\label{sec:Exp-Results_SynVA-V1}
We compare SynVA-V1 with VesselGPT~\cite{feldman2025vesselgpt},
PartVessel~\cite{chen2025hierarchical}, and the Chebyshev curve diffusion model of Mo et al.~\cite{mo2026anatomically}. We retrain the baselines using their official implementations on the
same $388$ training anatomies as SynVA-V1. The validation set contains the remaining $48$ of the $436$ anatomies with a vessel tree, stratified by node count (\cref{sec:SM_SynVA-V1_Training}). \Cref{sec:SM_SynVA-V1_Experiments_Baselines} details the baseline training and selection protocols. Validation MMD selects \synvaDepth{} from the $14$ SynVA-V1 geometry variants (\cref{sec:SM_SynVA-V1}).

Each model generates $1000$ unconditional samples. The test set comprises $89$ of the
$100$ real test samples from AneuX~\cite{juchler2022aneux}, IntrA~\cite{yang2020intra}, and CMHA~\cite{song2024cmha} that yield a valid tree and do not duplicate a training anatomy (\cref{sec:SM_SynVA-V1_Experiments_Test}).

MMD~\cite{gretton2012kernel}, COV~\cite{achlioptas2018learning}, and $1$-NNA~\cite{lopez2017revisiting} compare per-tree geometry features. Cosine similarity compares radius, edge-length, and tortuosity histograms~\cite{feldman2023vesselvae}. The topology metrics compare node-degree distributions ($\text{KL}_{\text{degree}}$) and Laplacian spectra (Spec.)~\cite{chen2025hierarchical}. A generated tree counts as a near-copy if its symmetric Chamfer distance~(CD) to the nearest training tree falls below the fifth percentile of the distances from the test trees to their nearest training tree.

\begin{table*}[htb]
\centering
\small
\setlength{\tabcolsep}{3.5pt}
\caption{Unconditional generation of healthy vessel trees on the test split ($n=89$). All methods train on the same split, and none is selected on the test set. Values are mean $\pm$ standard deviation over $10$ draws (standard deviations of the histogram and topology metrics in \cref{tab:SM_baseline_sd}). A model whose samples lie as close to the training trees as unseen real trees yields about $5\,\%$ near-copies; bold marks the best value per column except near-copies.}
\label{tab:synva_v1_main}
\scalebox{0.7}{
\begin{tabular}{lccccccccc c}
\toprule
Method & Valid
& MMD$\downarrow$
& COV$\uparrow$
& $1$-NNA$\!\to\!.5$
& $\cos_\text{radius}\uparrow$
& $\cos_\text{length}\uparrow$
& $\cos_\text{tortuosity}\uparrow$
& $\text{KL}_\text{degree}\downarrow$
& Spec.$\downarrow$
& Near-copies \\
\midrule
VesselGPT~\cite{feldman2025vesselgpt}
  & $92.3\,\%$ & $0.091 \pm 0.013$ & $0.384 \pm 0.026$ & $0.789 \pm 0.029$
  & $\mathbf{0.944}$ & $0.922$ & $0.919$ & $0.0060$ & $0.0056$ & $39.4\,\%$ \\
PartVessel~\cite{chen2025hierarchical}
  & $100\,\%$ & $0.336 \pm 0.012$ & $0.271 \pm 0.038$ & $0.983 \pm 0.006$
  & $0.671$ & $0.499$ & $0.988$ & $0.0005$ & $0.0018$ & $9.5\,\%$ \\
Chebyshev~\cite{mo2026anatomically}
  & $100\,\%$ & $0.217 \pm 0.009$ & $0.316 \pm 0.032$ & $0.979 \pm 0.012$
  & $0.735$ & $0.345$ & $0.996$ & $0.0038$ & $0.0029$ & $18.6\,\%$ \\
SynVA-V1 (ours)
  & $100\,\%$ & $\mathbf{0.066 \pm 0.007}$ & $\mathbf{0.412 \pm 0.029}$
  & $\mathbf{0.733 \pm 0.021}$
  & $0.935$ & $\mathbf{0.955}$ & $\mathbf{0.999}$
  & $\mathbf{0.0001}$ & $\mathbf{0.0007}$ & $0.1\,\%$ \\
\bottomrule
\end{tabular}}
\end{table*}

\begin{table}[htb]
\centering
\caption{Quantitative comparison of SynVA-A1 and SynVA-A2 on the aneurysm test set. For a fair comparison, results are also shown on the subset of SynVA-A2 successful generations ($n=66$).}
\label{tab:ResultsAneurysm}
\scalebox{0.7}{
\begin{tabular}{lccc}
\toprule
Mean Absolute   & SynVA-A1 & SynVA-A1 & SynVA-A2 \\
Difference & ($n=100$) & ($n=66$) & ($n=66$) \\
\midrule
$A_\mathrm{A}$ & 0.1560 $\pm$ 0.2599 & \textbf{0.1634 $\pm$ 0.2853} & 0.1963 $\pm$ 0.3248 \\
$A_\mathrm{CH}$ & 0.1539 $\pm$ 0.2631 & \textbf{0.1606 $\pm$ 0.2850} & 0.1986 $\pm$ 0.3238 \\
$D_{\max}$ & 0.0842 $\pm$ 0.0992 & \textbf{0.0866 $\pm$ 0.1034} & 0.1028 $\pm$ 0.1085 \\
$H_{\max}$ & 0.0796 $\pm$ 0.0905 & \textbf{0.0826 $\pm$ 0.0962} & 0.1100 $\pm$ 0.1116 \\
$H_{\mathrm{ortho}}$ & 0.0715 $\pm$ 0.0821 & \textbf{0.0764 $\pm$ 0.0906} & 0.1035 $\pm$ 0.0975 \\
$V_\mathrm{A}$ & 0.0214 $\pm$ 0.0449 & \textbf{0.0224 $\pm$ 0.0486} & 0.0330 $\pm$ 0.0608 \\
$V_\mathrm{CH}$ & 0.0214 $\pm$ 0.0449 & \textbf{0.0224 $\pm$ 0.0486} & 0.0257 $\pm$ 0.0553 \\
$W_{\max}$ & 0.0574 $\pm$ 0.0805 & \textbf{0.0581 $\pm$ 0.0856} & 0.0694 $\pm$ 0.0927 \\
$W_{\mathrm{ortho}}$ & 0.0567 $\pm$ 0.0752 & \textbf{0.0572 $\pm$ 0.0798} & 0.0703 $\pm$ 0.0895 \\
\midrule
CD & 0.0864 $\pm$ 0.0652 & \textbf{0.0919 $\pm$ 0.0695} & 0.1001 $\pm$ 0.0710 \\
\bottomrule
\end{tabular}}
\end{table}

\begin{table}[htb]
\centering
\caption{Medical expert evaluation of aneurysm realism on a scale from $0$ to $5$ across $n=40$ cases.}
\label{tab:ratings_summary}
\scalebox{0.7}{
\begin{tabular}{lccc}
\toprule
  & GT & SynVA-A1  & SynVA-A2 \\
  & $(n=40)$ & $(n=40)$ & $(n=40)$\\
\midrule
Medical Expert 1 & 4.675 $\pm$ 0.797 & 4.500 $\pm$ 0.751 & 4.475 $\pm$ 0.640 \\
Medical Expert 2 & 4.350 $\pm$ 1.051 & 3.550 $\pm$ 1.011 & 3.950 $\pm$ 0.783 \\
\midrule
Combined Rating & \textbf{4.513 $\pm$ 0.941} & 4.025 $\pm$ 1.006 & 4.213 $\pm$ 0.758 \\
\bottomrule
\end{tabular}
}
\end{table}

SynVA-V1 reached the lowest MMD, the highest coverage, and the $1$-NNA closest to $0.5$ (\cref{tab:synva_v1_main}). It also showed the highest cosine similarities for the edge-length and tortuosity histograms and the lowest values for both topology metrics. VesselGPT performed best among the baselines on all three feature-based metrics and showed the highest radius histogram similarity across all methods. The $1$-NNA values for PartVessel ($0.983$) and Chebyshev ($0.979$) indicate that the classifier distinguished generated from real trees with high accuracy in the evaluation feature space.

For VesselGPT, the near-copy rate increased from $3.1\,\%$ at epoch $100$ to $35.3\,\%$ at epoch $793$ and reached $39.4\,\%$ at the checkpoint with the lowest training loss (epoch $998$; \cref{tab:synva_v1_main,tab:SM_vgpt_copies}). The corresponding rate for SynVA-V1 was $0.1\,\%$ (\cref{tab:synva_v1_main}), so SynVA-V1 almost never generates a tree that lies closer to a training tree than an unseen test anatomy, and its samples lie slightly farther from their nearest training tree than the test trees do (median CD $0.129$ vs.\ $0.113$). \Cref{fig:vessel_structures} shows qualitative samples of all methods.

\subsection{Aneurysm Generation and Pathological Vessels}
\label{sec:Exp-Results_SynVA-AP}
To evaluate the two ostium-conditioned learning-based models, SynVA-A1 and SynVA-A2, we report
results for their respective best-performing configurations, selected from extensive experiments
exploring architectural choices, training strategies, and data settings
(\cref{sec:SM_SynVA-A1,sec:SM_SynVA-A2}). The quantitative comparison uses mean absolute differences
for independent morphological parameters and the CD on the test set. SynVA-A1 receives these
parameters as part of its condition. Results in \cref{tab:ResultsAneurysm} show that SynVA-A1
performs slightly better on the shared subset of successful generations from SynVA-A2~($66\,\%$
properly terminated and valid meshes, \cref{sec:SM_SynVA-A2}), yielding lower parameter deviations
and a reduced CD.

The quantitative results are complemented by an expert study involving two neuroradiologists, who participated voluntarily and gave informed consent. Since the study collected no personal or patient data, no ethics review was required. For $40$ real vessel cases, each expert was presented with the corresponding ground-truth aneurysm and aneurysms generated by SynVA-A1 and SynVA-A2 side by side, for a total of $120$ evaluated samples. All aneurysms were generated at the same ostium of the respective vessel. The experts rated each pathological vessel on a scale from $0$ (unrealistic) to $5$ (realistic), blinded to whether a sample was real or generated.

\begin{figure}[htb]
  \centering
  \includegraphics[width=\linewidth]{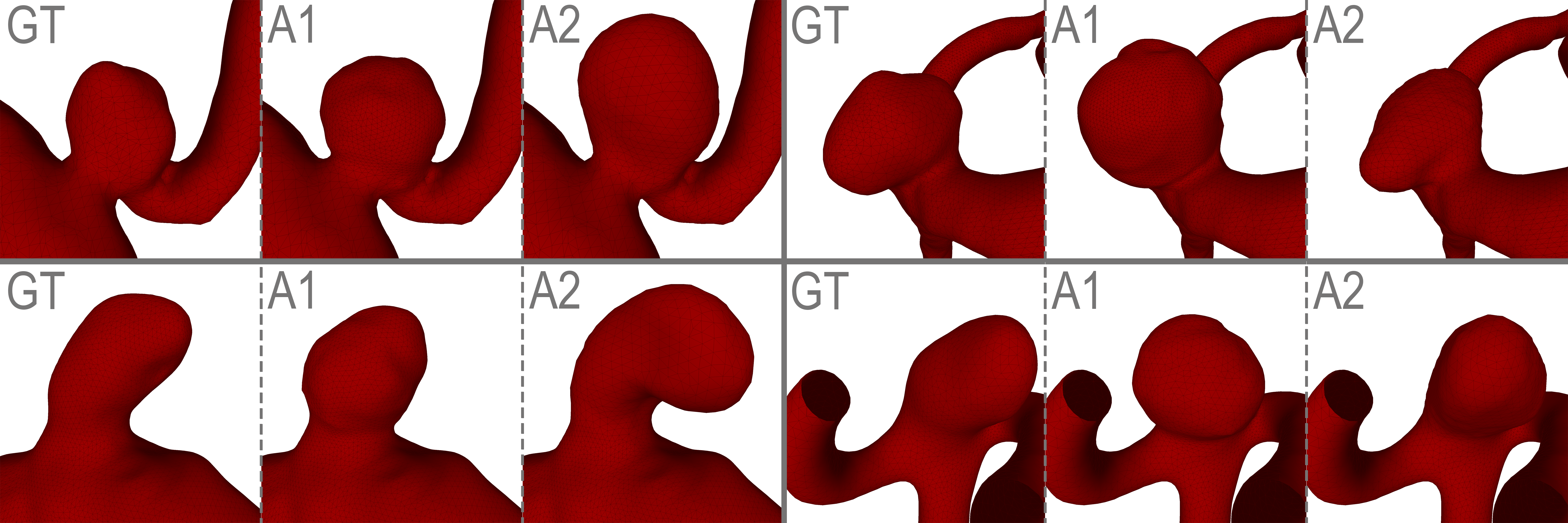}
  \vspace{-6mm}
  \caption{Qualitative comparison of ground-truth and generated aneurysms on real vessel geometries.}
  \label{fig:Aneurysm}
  \vspace{-3mm}
\end{figure}

The mean ratings (\cref{tab:ratings_summary}) show that ground-truth aneurysms are perceived as most realistic, with small but significant differences to both models. Among the generated samples, SynVA-A2 is rated slightly higher than SynVA-A1, although this difference is not statistically significant (inter-rater and significance statistics in \cref{tab:SM_SynVA-A-ExpertAgreement,tab:SM_SynVA-A-ExpertWilcoxon}).
Qualitative feedback from the experts indicates that overly symmetric shapes, irregular proportions between aneurysm neck and sac, and sharp geometric artifacts reduce perceived realism. Certain ground-truth samples—particularly aneurysms with pronounced flask-neck morphologies—were also rated as less realistic, which may indicate potential labeling inconsistencies in the real dataset.
Qualitative examples are shown in \cref{fig:Aneurysm}. Both models consistently generate aneurysms conditioned on the ostium and the local vessel geometry, producing shapes that closely resemble the ground truth while still exhibiting plausible anatomical variation. 
Since SynVA-P1 has no paired real-world counterpart, we evaluate the dataset by transferring it to real data in two downstream tasks, where pretraining on SynVA-P1 improves results across all fractions of real-world data~(\cref{sec:SM_Downstream}). In the three-class point-cloud segmentation of vessels, aneurysms, and ostia, the pretrained model, using only 10\,\% real data, nearly matches the performance achieved by training on all real data (63.88 vs. 64.84 mIoU). For morphological parameter estimation, $R^2$ increases from $-0.07$ (10\,\% real data) to $0.45$ (synthetic pretraining and 10\,\% real data), exceeding the level achieved by training on all real data (0.42). A comparison of size-matched corpora from SynVA-P1, SynVA-A1, and SynVA-A2 shows that all three sources improve transfer to real data, without any single source being superior across all settings (\cref{sec:SM_Downstream_SynDataComparison}). SynVA-A2 becomes usable only after pretraining on SynVA-P1, since it produces no valid generations when trained only on real data.

\section{Discussion}
\label{sec:Discussion}

SynVA demonstrates how learning-based and procedural generation can be combined to synthesize labeled vascular meshes with anatomy-conditioned aneurysms, and how such data transfer to real 3D tasks. A central design choice is to decouple healthy vessel generation from aneurysm synthesis and to connect both through an explicit ostium. This formulation reflects aneurysm formation as a localized process on a pre-existing vessel surface, while enabling both scalable automatic generation and controlled editing.
For healthy vessel generation, SynVA-V1 outperforms three published baselines retrained on the same split on all distribution-level metrics except the radius histogram similarity while producing almost no near-copies of training trees. VesselGPT reproduces the radius distribution slightly better but copies a substantial share of its training trees, whereas PartVessel and Chebyshev generate trees that remain clearly distinguishable from real anatomies. Among the evaluated methods, SynVA-V1 thus best approximates the real vessel distribution while generating new anatomies.
The procedural variant SynVA-P1 provides a robust and fully controllable alternative, but is currently limited in complexity, as it supports only a single bifurcation. 
The aneurysm generators exhibit complementary strengths. SynVA-A1 reliably approximates ground-truth shapes and achieves stronger quantitative performance, benefiting from richer conditioning signals. However, it tends to produce slightly more symmetric geometries, which can reduce perceived realism. In contrast, the experts rated SynVA-A2 slightly higher than SynVA-A1, although not significantly, despite weaker quantitative metrics and a higher mesh-generation failure rate. This highlights that improved numerical accuracy does not necessarily translate to higher clinical realism. Overall, both learning-based models produce plausible aneurysms conditioned on local vessel anatomy, while the procedural approach SynVA-P1 offers controllability at the cost of reduced variability.
Since SynVA-P1 has no real-world counterpart, its value becomes apparent in the utility of the data it generates: As a pretraining source, it transfers geometric prior knowledge to both local and global tasks, compensates for a lack of real-world training data, and makes SynVA-A2 trainable in the first place, though it does not replace real data. As data sources, the generators complement each other without one being superior in all settings: Data from real parent vessels offer anatomical fidelity, while procedural data do not require patient data and can scale arbitrarily.

\section{Conclusion, Limitations and Future Work}
\label{sec:ConclusionLimitationFutureWork}

We presented SynVA, a modular toolkit for synthetic vascular mesh generation and anatomy-conditioned aneurysm synthesis. SynVA combines a two-stage model with autoregressive topology and flow-matching geometry for healthy vessel generation, ostium-conditioned learning-based aneurysm generators, and a fully procedural model for scalable data synthesis. By generating aneurysms from a predefined ostium on an existing vessel surface, SynVA models aneurysm formation as a localized, vessel-conditioned process rather than as isolated shape generation. The resulting pipeline produces complete pathological vascular meshes with semantic labels, morphological descriptors, and consistent vessel--aneurysm interfaces, including a large-scale synthetic dataset of 50,000 samples which enabled the training of our SynVA-A2 model.
SynVA has several limitations. The current vessel model focuses on local tree-like structures and does not yet capture complete patient-level cerebrovascular anatomy or all variants of the Circle of Willis. The procedural component relies on simplified physiological assumptions and statistical priors, while the learning-based models inherit biases from heterogeneous and limited real aneurysm datasets. With SynVA-A2, not every generation results in a valid mesh; however, repeated generation with a perturbed ostium largely mitigates this issue (see \cref{sec:SM_Downstream_SynDataComparison}). In addition, SynVA targets anatomical and morphological plausibility, whereas hemodynamic quantities such as wall shear stress or pressure fields are neither modeled nor validated. 
Some meshes may therefore require further preparation before CFD-based analysis.
Owing to its modular design, individual components of SynVA can be replaced without changing the remaining toolkit. Future work will extend SynVA toward larger and more anatomically complete vascular networks, incorporate vessel-type and location-specific priors, improve SynVA-A2 through constrained termination and recent mesh generators, and condition aneurysm generation on hemodynamic or simulation-derived descriptors. We also plan to evaluate the utility of SynVA-generated data in additional downstream tasks such as rupture-risk assessment and accelerated flow prediction.

\clearpage
\section*{LLM Usage Statement}
The authors take full responsibility for all content. Claude~(Anthropic) was used to assist with revising texts written by the authors, providing review-style feedback on parts of the manuscript, and revising code, analyses, and figures. The overall scientific concept, consisting of research ideas, hypotheses, and conclusions, as well as the literature review, was developed without the use of an LLM. The authors have reviewed all content for which an LLM provided assistance.
{
    \small
    \bibliographystyle{ieeenat_fullname}
    \bibliography{main}
}


\SupplementaryReferences
\section{Supplementary Material: Overview}
\label{sec:SI}
The supplementary material provides additional technical details on the individual components of the SynVA toolkit. It includes further information on the morphological parameters (\cref{sec:SM_MorphologicParameters}), the SynVA-V1 healthy vessel generation model (\cref{sec:SM_SynVA-V1}), the GUI for ostium positioning (\cref{sec:SM_GUI}), and the two ostium-conditioned aneurysm generation models SynVA-A1 (\cref{sec:SM_SynVA-A1}) and SynVA-A2 (\cref{sec:SM_SynVA-A2}). In addition, the procedural pipeline SynVA-P1 (\cref{sec:SM_SynVA-P1}) and the generated synthetic dataset (\cref{sec:SM_Dataset}) are described in detail, followed by the downstream tasks used to validate its usefulness (\cref{sec:SM_Downstream}). 

\section{Morphological Parameters}
\label{sec:SM_MorphologicParameters}
\begin{figure}[ht]
  \centering
  \includegraphics[width=0.45\linewidth]{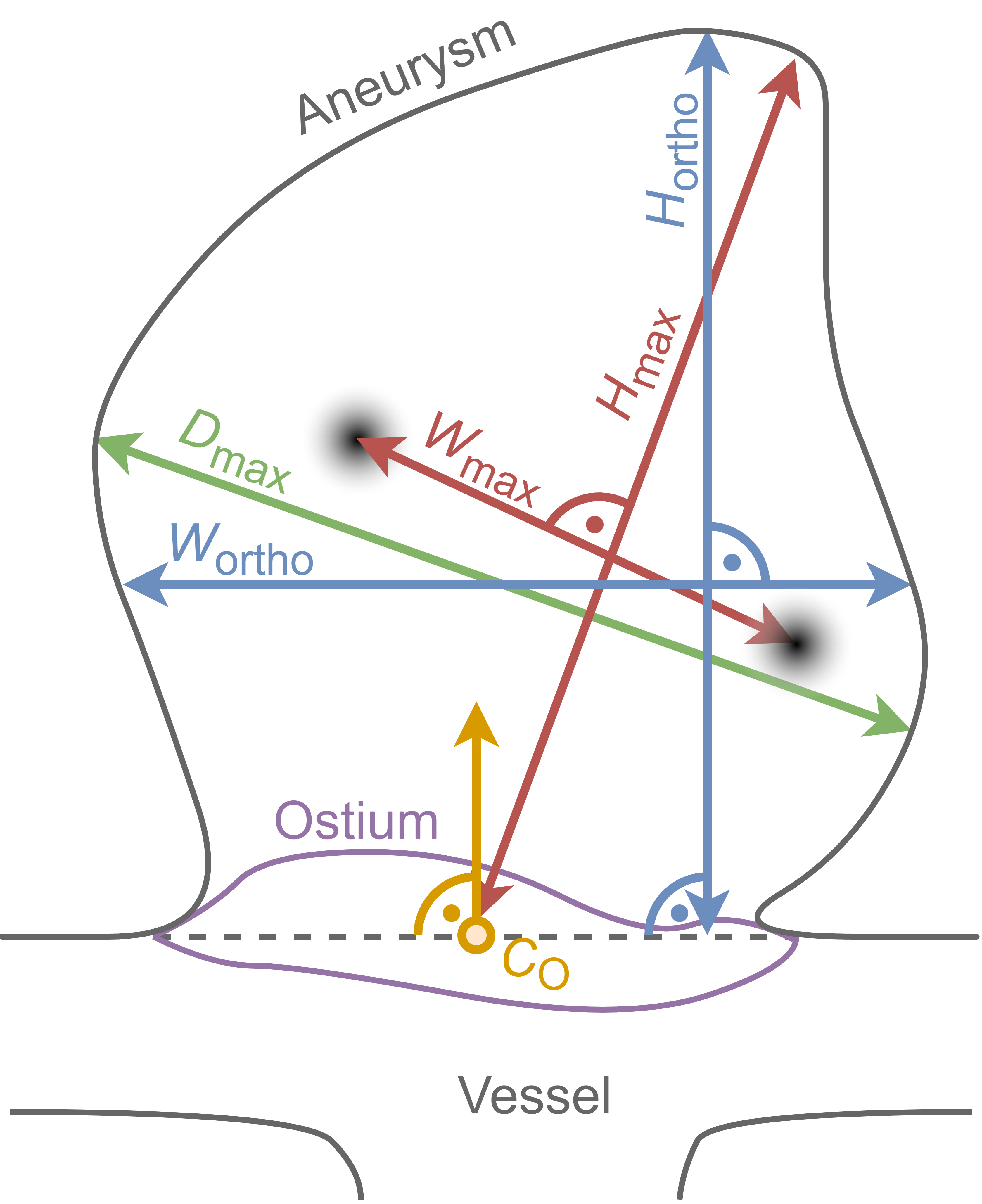}
  \caption{Schematic of the aneurysm and the morphological parameters $H_\mathrm{max}$, $W_\mathrm{max}$, $H_\mathrm{ortho}$, $W_\mathrm{ortho}$, and $D_\mathrm{max}$. The centroid of the ostium ($C_\mathrm{O}$) and its normal vector are also shown. The figure is based on~\citet{niemann2018rupture,saalfeld2018semiautomatic}.}
  \label{fig:SM_MorphologicalParameters}
\end{figure}

The characterization of three-dimensional aneurysm morphology is not only valuable as an additional input for deep learning approaches, but also plays a critical role in the assessment of rupture risk. Consequently, several studies have focused on the automatic extraction of morphological parameters from intracranial aneurysms~\cite{niemann2018rupture,saalfeld2018semiautomatic}. Building upon these works, we re-implemented the computation of the relevant morphological descriptors within our framework. The resulting parameters are illustrated in \cref{fig:SM_MorphologicalParameters} and summarized in \cref{tab:SM_MorphologicalParameters}.

\begin{table}[h]
  \caption{Morphological parameters used in this work and their definitions~\cite{niemann2018rupture,saalfeld2018semiautomatic}. Units are omitted, as the synthetic meshes are not defined in an absolute physical scale. When applied to real-world data, lengths, areas, and volumes correspond to physical units (e.g., mm, mm$^2$, mm$^3$).}
  \label{tab:SM_MorphologicalParameters}
  \centering
  \scalebox{0.78}{
  \begin{tabular}{ll}
    \toprule
    Feature     & Description     \\
    \midrule
    $A_\mathrm{A}$ & Aneurysm area excluding the ostium \\
    $V_\mathrm{A}$ & Aneurysm volume (closed ostium) \\
    $A_\mathrm{O1}$ & Ostium area (variant 1, not flattened) \\
    $A_\mathrm{O2}$ & Ostium area (variant 2, flattened) \\
    $D_\mathrm{max}$ & Maximum aneurysm diameter \\
    $H_\mathrm{max}$ & Maximum aneurysm height from $C_\mathrm{O}$ \\
    $W_\mathrm{max}$ & Maximum aneurysm width perpendicular to $H_\mathrm{max}$ \\
    $H_\mathrm{ortho}$ & Maximum aneurysm height perpendicular to the ostium plane \\
    $W_\mathrm{ortho}$ & Maximum aneurysm width perpendicular to $H_\mathrm{ortho}$ \\
    $N_\mathrm{max}$ & Maximum ostium diameter \\
    $N_\mathrm{avg}$ & Average ostium diameter \\
    $AR_1$ & Aspect ratio: $AR_1=H_\mathrm{ortho}/N_\mathrm{max}$ \\
    $AR_2$ & Aspect ratio: $AR_2=H_\mathrm{ortho}/N_\mathrm{avg}$ \\
    $A_\mathrm{CH}$ & Area of the convex hull of the aneurysm  \\
    $V_\mathrm{CH}$ & Volume of the convex hull of the aneurysm \\
    $EI$ & Ellipticity index: $EI=1-(18\pi)^\frac{1}{3}{V_\mathrm{CH}}^\frac{2}{3}/A_\mathrm{CH}$ \\
    $NSI$ & Non-sphericity index: $NSI=1-(18\pi)^\frac{1}{3}{V_\mathrm{A}}^\frac{2}{3}/A_\mathrm{A}$ \\
    $UI$ &  Undulation index: $UI=1-\frac{V_\mathrm{A}}{V_\mathrm{CH}}$  \\
    \bottomrule
  \end{tabular}}
\end{table}

The aneurysm surface area $A_\mathrm{A}$ is computed based on the triangular faces of the aneurysm mesh with an open ostium. To estimate the enclosed volume $V_\mathrm{A}$, the ostium is first sealed using a hole-closing operation, after which the volume is calculated. In cases where a watertight mesh cannot be obtained, the volume of the convex hull $V_\mathrm{CH}$ is used as a fallback approximation. This approximation directly affects derived indices such as $NSI$ and $UI$, with the latter becoming zero in such fallback scenarios.

The ostium area is computed using two complementary approaches. In the first variant, $A_\mathrm{O1}$ is defined as the sum of triangular faces formed by pairs of adjacent ostium vertices and the ostium centroid $C_\mathrm{O}$. In the second variant, all ostium vertices are projected onto the plane defined by the ostium centroid $C_\mathrm{O}$, and the area $A_\mathrm{O2}$ is subsequently calculated in this planar domain.

Geometric extent measures are defined next. The maximum diameter $D_\mathrm{max}$ corresponds to the largest Euclidean distance between any two vertices of the aneurysm mesh. In contrast, the maximum height $H_\mathrm{max}$ is defined as the largest distance between an aneurysm vertex and the ostium centroid $C_\mathrm{O}$. Based on this principal direction, the maximum width $W_\mathrm{max}$, measured perpendicular to the vector defined by $H_\mathrm{max}$, is computed using a ray-casting approach. Specifically, for each vertex, a ray is cast toward the closest point on the $H_\mathrm{max}$ axis, and the corresponding intersection point on the opposite side of the aneurysm surface is determined. The Euclidean distance between the vertex and this intersection point is then evaluated across all vertices, with the maximum value defining $W_\mathrm{max}$.

The orthogonal measures $H_\mathrm{ortho}$ and $W_\mathrm{ortho}$ are computed analogously. The key distinction is that $H_\mathrm{ortho}$ is defined as the maximum distance from aneurysm vertices to the ostium surface, rather than to the centroid.

Finally, ostium-related size measures are derived. The parameter $N_\mathrm{max}$ is computed analogously to $D_\mathrm{max}$, but restricted to ostium vertices. The average ostium diameter $N_\mathrm{avg}$ is defined as twice the mean distance between all ostium vertices and the centroid $C_\mathrm{O}$.

\Cref{tab:SM_MorphologicalParameters} lists the remaining morphological parameters.
\section{SynVA-V1: Vessel Generation}
\label{sec:SM_SynVA-V1}
\begin{figure*}[ht]
  \centering
  \includegraphics[width=\textwidth]{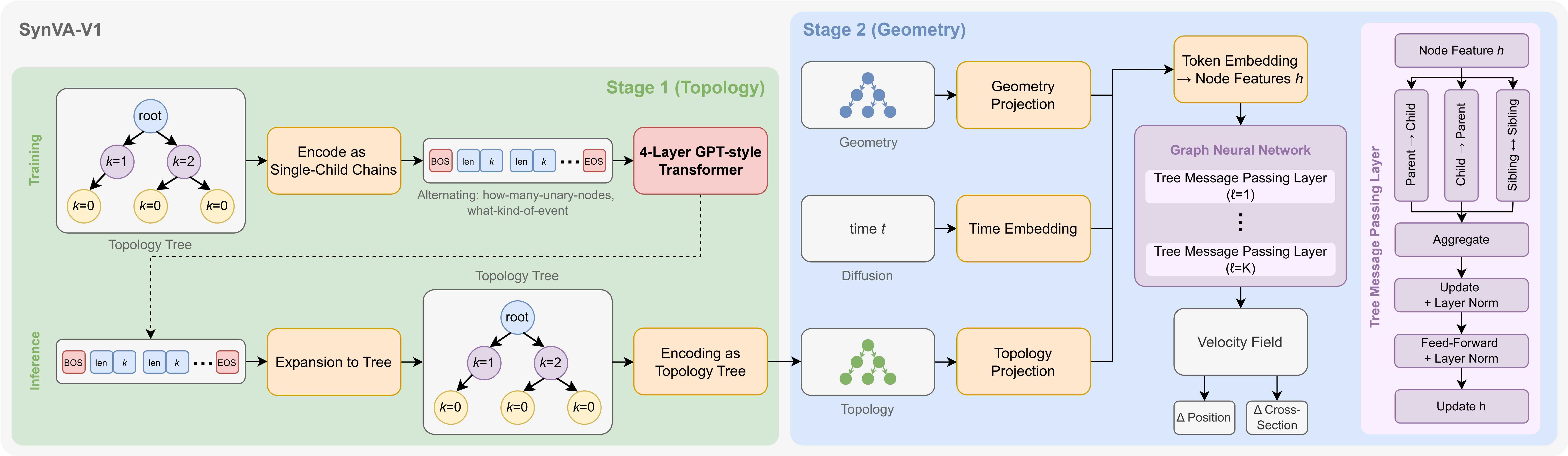}
  \caption{Overview of the SynVA-V1 architecture.
  \textbf{Stage~1 (Topology, green).} The rooted tree is encoded as a sequence of $(\ell_e, \delta_e)$ pairs alternating chain length and event type, and modeled by a 4-layer GPT-style autoregressive Transformer; at inference, the sampled sequence is expanded back into a topology tree.
  \textbf{Stage~2 (Geometry, blue).} Given the topology, the noisy per-node geometry and the flow time $t$ are projected into per-node token embeddings $h$ and processed by $K{=}10$ Tree Message Passing Layers; a final linear head outputs the per-node velocity, split into $\Delta$ position and $\Delta$ cross-section.
  \textbf{Right (pink).} A single Tree Message Passing Layer aggregates parent$\to$child, child$\to$parent and sibling$\leftrightarrow$sibling messages, followed by Update~+~LayerNorm and Feed-Forward~+~LayerNorm.}
  \label{fig:SM_V1Architecture}
\end{figure*}

SynVA-V1 is a two-stage model for the synthesis of healthy vessel meshes that combines autoregressive topology generation with flow-matching-based geometry generation.
This decomposition of topology and geometry enables structured modeling of vascular trees while allowing flexible and precise control over geometric detail. An overview of the architecture is shown in \cref{fig:SM_V1Architecture}.

\subsection{Data Preprocessing}
\label{sec:SM_SynVA-V1_Data}
Our pipeline for converting a raw vessel mesh into a rooted spline tree is adapted from VesselGPT~\cite{feldman2025vesselgpt}. We first center each raw mesh at the midpoint of its bounding box and scale it isotropically so that its largest half-extent equals one, and all lengths below refer to these units. For each mesh, we then extract the vessel centerline graph using the \textit{vmtknetworkextraction} script from VMTK~\cite{antiga2008image}. The script traces the centerline by marching along the vessel in steps proportional to the local vessel radius; the proportionality factor~(the \emph{advancement ratio}) is initially set to $1.05$, and we fall back to the more conservative values $1.02$ and $1.01$ if the previous run does not return a valid centerline network. For test meshes that fail with these settings, we repeat the extraction with a longer time limit and the additional value $1.005$. The extracted branches are stored as polylines, with one vessel radius assigned to each point. To obtain a consistent resolution along the centerline, each branch is resampled with a default point spacing of $0.02$. The spacing is locally reduced in regions where the vessel radius changes strongly and near bifurcations, so that geometrically important areas are represented with higher detail. For every resampled centerline point, we estimate the local tangent from neighboring polyline segments and cut the mesh with the corresponding orthogonal plane. If the cut produces several contour components, only the component closest to the centerline point is kept. Radial outliers are removed before fitting the remaining contour with a closed periodic cubic B-spline. Each fitted cross-section is then projected onto a fixed periodic basis with 8 control points and 12 knots. As a result, each cross-section is represented by $36$ scalar values: $24$ control-point coordinates and $12$ knot values.
The resampled centerline graph is then converted into a rooted tree. The resampled centerline points become nodes, and neighboring points along each polyline are connected by edges. Bifurcation points shared by multiple branches are merged into single graph nodes, and nodes with more than two children are binarized to obtain a binary branching structure. To select a root, we compute all-pairs shortest paths with the Floyd--Warshall algorithm \cite{floyd1962algorithm}, identify the longest graph diameter, and place the root at its midpoint. Starting from this root, a pre-order traversal produces the final ordered tree sequence. Each tree node stores a continuous feature vector $\bm{x}_i = \left(\bm{p}_i, \bm{c}_i, \bm{s}_i\right) \in \mathbb{R}^{39}$, where $\bm{p}_i \in \mathbb{R}^{3}$ denotes the position of node $i$, $\bm{c}_i \in \mathbb{R}^{24}$ contains the cross-section control points, and $\bm{s}_i \in \mathbb{R}^{12}$ contains the corresponding B-spline knots. In addition, each node is assigned a discrete child-count token $k_i \in \{0,1,2\}$, which indicates whether the node has zero, one, or two children. This yields a rooted spline tree with a fixed continuous node representation and an explicit discrete topology.

For SynVA-V1\textsubscript{topology}, only the discrete topology of the rooted spline tree is modeled. We therefore discard the continuous node features $\bm{x}_i$ and use only the child-count tokens $k_i \in \{0,1,2\}$. We traverse the tree in pre-order and compress the resulting child-count sequence by collapsing each chain of consecutive single-child continuation nodes ($k_i=1$) into a single length value, which is attached to the next leaf or bifurcation. The compressed sequence thus consists only of events — leaves, bifurcations, and the possible single-child root continuation — and each event is represented by a pair~$(\ell_e, \delta_e)$, where $\ell_e \in \{0,\dots,L_{\max}\}$ is the length of the preceding $k_i{=}1$ chain and $\delta_e \in \{0,1,2\}$ is the event degree. Here, $\delta_e=0$ corresponds to a leaf, $\delta_e=1$ to a single-child root continuation, and $\delta_e=2$ to a bifurcation.

For TreeFlowNet (main backbone of our geometry model), we keep the full continuous feature vector $\bm{x}_i = (\bm{p}_i, \bm{c}_i, \bm{s}_i)$ of every node and pair it with three discrete structure tokens that describe the conditioning topology $\mathcal{T}$: the child count $k_i$ from above, the node depth $d_i$, and the child slot $o_i \in \{1,2\}$ identifying which sibling position the node occupies under its parent. The position $\bm{p}_i$ is converted from absolute to parent-relative form, so that the model reconstructs the displacement to the parent rather than the global location. Tree edges that are very short relative to the local vessel radius are typically spurious branches introduced by the centerline extraction, and we drop them with a length filter that distinguishes by edge type: an edge into a single-child node (a \emph{continuation edge}) is kept only if its length exceeds the local vessel radius ($\alpha_\mathrm{cont}{=}1.0$), while an edge into a bifurcation is already kept at $0.30{\times}$ the local radius ($\alpha_\mathrm{bif}{=}0.30$), so that genuine branching points are preserved even when locally short. We then shift the position and control-point channels of all trees by a constant offset, the median node position, and divide by one global factor, the $99.9$th percentile of the absolute values ($1.70$), so the trees retain their relative size after per-mesh scaling. The filter and the normalization apply to the training, validation, and test trees of all methods of this work, and we map the samples of every method back to these coordinates before evaluation. TreeFlowNet also $z$-normalizes each of the $39$ channels using the statistics from the training anatomies.
For the \synvaAlign{} and aligned-$45^\circ$ variant only, we also repair cross-section rings whose plane tilts by more than $60^\circ$ ($45^\circ$ for aligned-$45^\circ$) relative to the parent node's ring. Such tilts arise from the spline fit at sharp bends and near bifurcations and do not reflect the anatomy. We project the control points of each such ring onto a plane whose normal lies halfway between the parent normal and the local centerline tangent, keeping its centroid and mean radius. All other variants use the rings as extracted.

\subsection{Architecture}
\label{sec:SM_SynVA-V1_Architecture}
SynVA-V1\textsubscript{topology} (\cref{fig:SM_V1Architecture}, Stage~1) is a small GPT-$2$-style autoregressive Transformer decoder $p_\phi(\mathcal{T})$ \cite{vaswani2017attention}. The vocabulary has $|\mathcal{V}| = 70$ tokens in three groups: $64$ chain-length tokens $\ell_e \in \{0, \dots, L_{\max}{=}63\}$ that count consecutive single-child continuation nodes, $3$ event-degree tokens $\delta_e \in \{0, 1, 2\}$ for the three event types (leaf, single-child root continuation, bifurcation), and the $3$ special tokens BOS, EOS and PAD. The decoder has $4$ Transformer layers, $128$-dim token embeddings, $4$ attention heads, and dropout $0.15$ on the residual, attention and embedding paths.

TreeFlowNet (\cref{fig:SM_V1Architecture}, Stage~2) is a topology-conditioned velocity field $v_\theta(\bm{x}_t, t \mid \mathcal{T}) \in \mathbb{R}^{N \times 39}$, implemented as a typed message-passing GNN~\cite{gilmer2017neural,schlichtkrull2018modeling} with global self-attention~\cite{vaswani2017attention} interleaved every $g{=}3$ layers. Each node is embedded into a token feature $h_i^{(0)} \in \mathbb{R}^{d}$ ($d{=}256$) from three input streams. 
The discrete structure tokens $(k_i, d_i, o_i)$ are passed through three separate learned embedding tables, the results are concatenated and projected by a small GELU MLP into a structural embedding; the noisy geometry $\bm{x}_{i,t}$ at flow time $t$ is mapped to $\mathbb{R}^{d}$ by a linear projection; and the flow time $t$ itself is encoded by a sinusoidal embedding.
The structural and geometry features are concatenated and linearly projected back to $\mathbb{R}^{d}$, the time embedding is then added, and the sum is passed through LayerNorm. $K{=}10$ \texttt{TreeMessagePassingLayer} blocks (\cref{fig:SM_V1Architecture}, right) then update $h_i$ as follows.  Every edge of the tree falls into one of three types read from the parent index $\pi$ — parent$\to$child, child$\to$parent or sibling$\leftrightarrow$sibling — and each type is associated with its own edge-type-specific message MLP. For every edge, the source and destination node features are concatenated and passed through the corresponding MLP to produce a per-edge message. At each node, all incoming messages are summed, concatenated with the current node feature, passed through an update MLP, added back as a residual, and normalized with LayerNorm. A feed-forward $+$ LayerNorm residual closes the block, and every $g{=}3$ layers a global multi-head self-attention block ($8$ heads) aggregates long-range context across the tree.  After the last block, a linear head projects $h_i$ back to $\mathbb{R}^{39}$, giving the per-node velocity $v_\theta(\bm{x}_t, t \mid \mathcal{T})_i$.

To generate new synthetic vessels, SynVA-V1\textsubscript{topology} first samples a topology $\mathcal{T}$ autoregressively with nucleus decoding (temperature $1.0$, top-$p{=}0.95$) under a hard grammar mask. Before every step, the logits of all tokens that would violate the rooted-tree grammar are set to $-\infty$, so that they can never be drawn. Masked cases include a length token where a degree token is expected (and vice versa, since events strictly alternate $\ell_e, \delta_e$), a degree-$1$ event after the first event (the single-child root continuation may only occur once, at the start), and an EOS while child slots are still open. Every sampled sequence therefore expands into a valid rooted binary tree of $N$ nodes.

TreeFlowNet then samples the geometry conditioned on $\mathcal{T}$ by integrating the ODE $\dot{\bm{x}}_t = v_\theta(\bm{x}_t, t \mid \mathcal{T})$ from $\bm{x}_0 = \bm{z} \sim \mathcal{N}(0, I) \in \mathbb{R}^{N \times 39}$ to $\bm{x}_1$ with $100$ Euler steps. For the unconditional comparison in the main paper, SynVA-V1\textsubscript{topology} samples $\mathcal{T}$, and the ablation in \cref{sec:SM_SynVA-V1_Results} additionally conditions on the ground-truth topology of the test trees. With the depth warp of \synvaDepth{} (\cref{sec:SM_SynVA-V1_Experiments_Models}), every node advances along its own warped time. The denoised per-node vectors $\bm{x}_i^{(1)} = (\bm{p}_i, \bm{c}_i, \bm{s}_i)$ are then denormalized with the training-set $z$ statistics, and every ring is regularized: its control points are projected onto their best-fit plane, a periodic spline is refit with a damped radius correction, and the aspect ratio is moved by $35\,\%$ towards $0.786$ at constant area. Then we accumulate the parent-relative positions $\bm{p}_i$ into absolute coordinates via a single traversal from the root, and reconstruct each cross-section ring from its 8 control points $\bm{c}_i$ and 12 B-spline knots $\bm{s}_i$. The result is a complete rooted spline tree, ready for the mesh-reconstruction pipeline below.

Predicted spline trees are converted to watertight surfaces in five deterministic stages: arc-length spline-station interpolation ($\Delta s{=}0.005$) of $64$-point cross-section rings; per-segment screened-Poisson reconstruction~\cite{kazhdan2013screened} at octree depth $D{=}8$ with end caps suppressed at bifurcations; Boolean union of the per-segment tubes; localized Laplacian and Taubin~\cite{taubin1995signal} smoothing in $3{\times}$ local-radius balls around bifurcation centers; and a global re-Poisson pass at depth $D'{=}9$.

\subsection{Training Details} 
\label{sec:SM_SynVA-V1_Training}
All models in this work use the same split of the real vessel data into $662$ training and $100$ test cases, and none uses the test cases for model selection. For SynVA-V1, the training cases yield trees of $436$ anatomies from AneuX~\cite{juchler2022aneux} and IntrA~\cite{yang2020intra}. The centerline extraction needs meshes with open vessel ends. Cases drop out when no healthy vessel mesh with open ends is available or when centerline extraction or conversion to a tree fails. We added the CMHA meshes to our data only after training, so
no CMHA anatomy enters training or validation. Seven of the $436$ anatomies are AneuX cases that the shared split does not list. None of them is in the test split, and the geometric twin check in \cref{sec:SM_SynVA-V1_Experiments_Test} covers them. For SynVA-V1 and all its baselines, we hold out $48$ of the $436$ anatomies, stratified by tree size, as validation anatomies. They serve only for checkpoint and model selection, and we train on the remaining $388$, each in $10$ random rotations. \Cref{sec:SM_SynVA-V1_Experiments_Test} describes the test trees. 

The topology model and the geometry model of SynVA-V1 are trained independently on these trees. SynVA-V1\textsubscript{topology} sees each training anatomy once, since a rotation does not change the topology, and is optimized for $160$ epochs with AdamW (lr $3{\times}10^{-4}$, weight decay $0.02$, cosine schedule, batch size $32$, dropout $0.15$). Training minimizes a next-token cross-entropy loss on the topology sequence of event pairs~$(\ell_e, \delta_e)$, and we keep the checkpoint with the lowest cross-entropy on the validation anatomies. 

TreeFlowNet trains on the $3880$ rotated training trees for up to $400$ epochs with AdamW (lr $2{\times}10^{-4}$, weight decay $0.01$, $500$-step linear warm-up, batch size $64$, seed $42$) under the optimal-transport conditional flow-matching loss~\cite{lipman2023flow, tong2024improving}
\begin{equation}
\mathcal{L}_{\mathrm{FM}}
= \mathbb{E}_{t, \bm{x}_1, \bm{z}}
\bigl\lVert v_\theta\bigl((1{-}t)\bm{z} + t\bm{x}_1,\, t \mid \mathcal{T}\bigr)
              - (\bm{x}_1 - \bm{z}) \bigr\rVert_{2,m}^2 ,
\label{eq:SM_fm}
\end{equation}
where $\bm{x}_1$ is the ground-truth per-node geometry of a training tree under topology $\mathcal{T}$, $\bm{z} \sim \mathcal{N}(0, I)$, and $\lVert\cdot\rVert_{2,m}$ is a masked block-weighted norm over the position, cross-section and knot channels of $\bm{x}_i = (\bm{p}_i, \bm{c}_i, \bm{s}_i)$ with equal weights $(\lambda_p, \lambda_c, \lambda_s) = (1, 1, 1)$. Auxiliary regularizers for knot monotonicity ($0.5$), cross-section planarity ($1.0$) and radius consistency ($3.0$) act on the same blocks. We keep an exponential moving average of the weights (decay $0.999$) for sampling, select the epoch with the lowest flow-matching loss on the $480$ rotated validation trees, and stop training after $60$ epochs without improvement. During training, the conditioning topology $\mathcal{T}$ is the ground-truth topology of the same tree. At inference, SynVA-V1\textsubscript{topology} samples $\mathcal{T}$ for the unconditional comparison in the main paper, and the ablation in \cref{sec:SM_SynVA-V1_Results} additionally conditions on the ground-truth topology of the test trees.

\subsection{Experimental Setting}
\label{sec:SM_SynVA-V1_Experiments}

\subsubsection{Test Split}
\label{sec:SM_SynVA-V1_Experiments_Test}
The test part of the split holds $100$ anatomies from AneuX, IntrA and CMHA~\cite{song2024cmha}. For two of them no healthy vessel mesh is available, and the centerline extraction fails for six, which leaves $92$ test trees. A check by case name alone cannot guarantee that no test anatomy also appears in training, because AneuX stores two aneurysms of one patient as two cases on the same vessel tree. We therefore match every validation and test tree to its nearest training and validation tree by the symmetric Chamfer distance (CD) of the node positions. Distinct anatomies lie at least $0.059$ apart (median $0.12$), whereas three of the $92$ test trees lie within $0.012$ of a training or validation tree. We remove these three and evaluate on the remaining $89$ trees ($62$ from AneuX, $14$ from IntrA and $13$ from CMHA). No validation tree has such a twin in the training set. Eleven test trees come from patients who also contribute a different vessel to training or validation, and we keep them, since their tree geometry differ. The training and validation trees contain no CMHA data, so all methods meet CMHA vessels only at test time.

\subsubsection{Model Variants}
\label{sec:SM_SynVA-V1_Experiments_Models}
We define three base configurations on which all further TreeFlowNet variants build, and the main comparison uses \synvaDepth{}, which validation MMD selects among all $14$ variants (\cref{sec:SM_SynVA-V1_Results_Models}). All variants use the same base flow-matching loss, and all variants except \synvaPhysTx{} also share the backbone and the optimizer settings. They differ in how each node maps the global flow time to its own time, in the distribution from which training draws $t$, in how the loss weights individual nodes, and in whether the cross-section rings of the training data are repaired.

\synvaBase\ is the simplest version. During training, the flow time $t$ is drawn uniformly from~$[0,\,1]$, every node contributes equally to the loss, and the cross-section rings are used in their original orientation.  We use this configuration as the reference against which the other variants are compared.

\synvaDepth\ adds a node-dependent warp of the flow time. Each node $i$ at depth $d_i$ follows its own time $t_i = t^{\,1 + \alpha d_i / d_{\max}}$ with $\alpha = 0.5$, where $d_{\max}$ is the depth of the deepest node of the tree. The interpolation between noise and data, the time embedding, and the Euler steps of the sampler all use this per-node time. Since $t_i \le t$, a deep node stays noisier than the root at every global $t$, so the sampler settles the proximal vessels first and the distal branches later. Every node still runs from $t_i{=}0$ to $t_i{=}1$.

\synvaAlign\ adds three further changes on top of \synvaDepth. First, the flow time is drawn from a logit-normal $t = \sigma(\mu + s\,\varepsilon)$ with $\varepsilon \sim \mathcal{N}(0, 1)$, $\mu{=}0$, $s{=}1$~\cite{esser2024sd3} instead of uniformly, which concentrates training on intermediate times. Second, the position channel of the loss is multiplied by a Gaussian focal weight $w_i^{\mathrm{focal}} = 1 + \lambda_\mathrm{f}\, \exp(-\lVert \bm{p}_i \rVert^2 / 2\sigma_\mathrm{f}^2)$ on the normalized parent-relative position, with $\lambda_\mathrm{f}{=}4$ and $\sigma_\mathrm{f}{=}1$. This weight up-weights short continuation segments whose parent-relative position is close to zero. Third, it trains on the ring-repaired data, in which rings that tilt by more than $60^\circ$ against their parent ring are projected onto a corrected plane (\cref{sec:SM_SynVA-V1_Data}).

All remaining variants build on \synvaDepth{}. \textit{Logit-time} replaces the uniform diffusion-time sampler of \synvaDepth{} with the logit-normal one, without focal weight or ring repair. All further variants of the first block build on logit-time. \textit{Focal-pos $\lambda{=}2/4$} adds the Gaussian focal weight on the position channel with strength $\lambda$, and \textit{focal-pos $+$ cp-bif} additionally up-weights the cross-section channels at bifurcation nodes by $3{\times}$ (with $\lambda{=}2$). \textit{Aligned-$45^\circ$} is the same recipe as \synvaAlign{} but repairs every ring that tilts by more than $45^\circ$ ($16\,\%$ instead of $7\,\%$ of the rings), so more rings are corrected, and \textit{anti-curl-back} adds a sibling-separation hinge loss preventing the two child branches from initially curling back together right after a bifurcation. The physiological block adds the physiological auxiliary losses to logit-time. They act on the bifurcation angle with target $\theta^\star$, the cosine between sibling directions, sibling radius symmetry, and depth-dependent radius shrinkage, along with cross-section continuity terms. We switch off the Murray's law term~\cite{murray1926physiological} in all of them. The angle term and the sibling-cosine term both act on the angle between the two child directions. The angle term tolerates deviations of up to $30^\circ$ ($\theta^\star{=}120^\circ$) or $20^\circ$ ($\theta^\star{=}50^\circ$) from its target, and all four variants keep the sibling-cosine target of $-0.5$, which corresponds to $120^\circ$, at three times the weight of the angle term. \textit{Physio-wide} targets a bifurcation angle of $\theta^\star{=}120^\circ$, \textit{physio-wide, no-warp} additionally drops the depth warp, \synvaPhys{} lowers the target to $\theta^\star{=}50^\circ$, and \synvaCombined{} adds focal-pos $\lambda{=}4$ and anti-curl-back to \synvaPhys{}.

\synvaPhysTx{} replaces the TreeFlowNet GNN backbone with a flat self-attention Transformer and keeps the physiological losses of \synvaPhys{}. It also differs from \synvaPhys{} in its training setup. It uses no depth warp, a wider network, a higher learning rate, a smaller batch, and an earlier version of the training trees with about $1.8$ times as many nodes per tree. The comparison with \synvaPhys{} therefore does not isolate the contribution of typed message passing.

\subsubsection{Baselines}
\label{sec:SM_SynVA-V1_Experiments_Baselines}
We follow the released code of each baseline. Where the code leaves a setting open or a setting requires adaptation to our data, we use the settings documented in this section and listed in \cref{tab:SM_baseline_settings}.

Like SynVA-V1, VesselGPT trains on ten rotations per training anatomy. Its paper leaves the epoch
count unspecified, and the released code trains the Transformer without validation. We train for $1000$ epochs with the released hyperparameters and follow the code in retaining the checkpoint with the lowest
training loss.

For PartVessel, we follow the released code and keep the stage-1 checkpoint with the lowest key-graph CD on the validation trees. Chebyshev expects centerlines in millimeters, so we map our unitless coordinates to $25$\,mm per unit, which validation MMD selects among $25$, $50$ and $100$\,mm. For both methods, we also train on the rotated training anatomies, and validation MMD selects the runs without rotations.

\begin{table*}[htb]
\centering
\small
\setlength{\tabcolsep}{4pt}
\mdseries
\caption{Baseline settings that the released code leaves open or that we adapt to our data and
evaluation protocol. All other settings follow the released code.}
\label{tab:SM_baseline_settings}
\scalebox{0.85}{
\begin{tabular}{l l p{5.4cm} p{6.4cm}}
\toprule
Method & Setting & Released code & Ours \\
\midrule
VesselGPT & Normalization & each tree scaled to $[-1,1]$ & one global factor for all trees, so trees keep the relative size they have after the per-mesh scaling, as for all other methods \\
          & Tree height & cut to $15$ or $20$ levels & full trees, as for all methods \\
          & Training & code trains the Transformer without validation and keeps the checkpoint with the lowest training loss & $1000$
  epochs with the code hyperparameters, checkpoint with the lowest training loss \\
          & Sampling & prompts with the first node of a validation tree
  & unconditional from the start token, single beam (selected on validation) \\
\midrule
PartVessel & Training & no validation split &
  lowest key-graph CD on the validation trees \\
           & Rotations & none & compared on validation, none kept \\
           & Output & dense centerline & resampled at the median node spacing of the training
  trees \\
\midrule
Chebyshev & Units & millimeters & $25$\,mm per unit, selected on validation among $25$, $50$,
  $100$ \\
          & Splits & separate splits of the authors' data per stage & our split for both stages \\
          & Root & leaf with the largest radius & first node of our trees \\
          & Node limit & $20$ key nodes & largest training key graph ($47$) \\
          & Rotations & none & compared on validation, none kept \\
          & Output & dense centerline & resampled at the median node spacing of the training
  trees \\
\bottomrule
\end{tabular}}
\end{table*}

\subsubsection{Evaluation Metrics}
\label{sec:SM_SynVA-V1_Experiments_Metrics}
MMD~\cite{gretton2012kernel}, COV~\cite{achlioptas2018learning} and $1$-NNA~\cite{lopez2017revisiting} compare generated and real trees as sets of per-tree feature vectors. Each vector holds $16$ features, namely node count, maximum depth, number of bifurcations, total branch length, mean, standard deviation and median of the radius, median and standard deviation of the bifurcation angle, median Murray's-law violation, fraction of tapering
violations, mean sibling symmetry ratio, mean and standard deviation of the edge length, spatial
span and compactness. We standardize the features with the statistics of the $388$ unrotated training anatomies and compute MMD with a Gaussian kernel whose bandwidth follows the median heuristic. $\cos_\text{radius}$ is the cosine similarity between the branch-radius histograms pooled over all real and over all generated trees~\cite{feldman2023vesselvae}, and $\cos_\text{length}$ and $\cos_\text{tortuosity}$ apply the same measure to the edge-length and tortuosity histograms. For topology, $\mathrm{KL}_\text{degree}$ compares the node-degree distributions and Spec.\ the Laplacian spectra~\cite{chen2025hierarchical}, both with their published definitions. For PartVessel and Chebyshev, we resample the dense centerlines at the median node spacing of the training trees before we compute any metric.
Each method generates $1000$ unconditional samples, and all metrics are averaged over $10$ random draws of $89$ samples that decode into a valid tree. Model selection uses the same metrics on the $48$ unrotated validation trees.

For the near-copy analysis, we compute symmetric CDs between sets of node positions
and use the $3880$ rotated training trees as the reference set. The fifth percentile of the
distances from the $89$ test trees to their nearest training tree sets the threshold ($0.077$).
Two distinct real anatomies fall below it at a similar rate, since $6.4\,\%$ of the training
anatomies lie closer than $0.077$ to a different training anatomy (own rotations excluded).

\subsection{Additional Results}
\label{sec:SM_SynVA-V1_Results}

\subsubsection{Model Variants}
\label{sec:SM_SynVA-V1_Results_Models}
\Cref{tab:SM_TreeflowAblation} lists all 14 geometry variants we trained. The first block holds the three base configurations and the variants that build on them, the second block the variants with physiological auxiliary losses, and the third block the Transformer variant baseline \synvaPhysTx{} (\cref{sec:SM_SynVA-V1_Experiments_Models}).

\begin{table*}[ht]
\centering
\small
\setlength{\tabcolsep}{4pt}
\caption{All $14$ geometry variants. The first column reports the MMD of each variant’s unconditional samples on the validation trees ($n{=}48$), which selects \synvaDepth{} for the main
comparison. The remaining columns report the test trees ($n{=}89$), once with sampled topology and
once with the ground-truth topology of each test tree. With sampled topology, each variant generates
$1000$ trees, and we average over $10$ draws of $48$ (validation) or $89$ (test) trees. With
ground-truth topology, each variant generates one tree per test topology, and we average over $3$
seeds. BifAngle is the angle between the two child directions at a bifurcation, taken as the median
per tree and averaged over the trees. Bold marks the best value per column. The last row shows real
training trees against the test trees and the bifurcation angle of the test trees.}
\label{tab:SM_TreeflowAblation}
\scalebox{0.8}{
\begin{tabular}{l c cc ccc}
\toprule
& Validation & \multicolumn{2}{c}{Test, sampled topology}
& \multicolumn{3}{c}{Test, ground-truth topology} \\
\cmidrule(lr){2-2}\cmidrule(lr){3-4}\cmidrule(lr){5-7}
Variant & MMD$\downarrow$ & MMD$\downarrow$ & $1$-NNA$\!\to\!.5$
& MMD$\downarrow$ & $1$-NNA$\!\to\!.5$ & BifAngle$\!\to\!62.2^\circ$ \\
\midrule
\multicolumn{7}{l}{\textit{TreeFlowNet variants (typed message-passing GNN backbone)}} \\
\synvaBase{} (plain base)                         & $0.072$          & $\mathbf{0.065}$ & $0.740$
  & $0.054$          & $0.745$          & $68.0^\circ$ \\
\textbf{\synvaDepth{}} (depth-warp, selected)     & $\mathbf{0.071}$ & $0.066$          & $\mathbf{0.733}$
  & $\mathbf{0.053}$ & $\mathbf{0.723}$ & $\mathbf{67.9^\circ}$ \\
logit-time                                        & $0.080$          & $0.074$          & $0.751$
  & $0.056$          & $0.745$          & $70.8^\circ$ \\
focal-pos $\lambda{=}2$                           & $0.080$          & $0.074$          & $0.755$
  & $0.056$          & $0.747$          & $72.2^\circ$ \\
focal-pos $\lambda{=}4$                           & $0.079$          & $0.071$          & $0.756$
  & $0.056$          & $0.773$          & $72.6^\circ$ \\
focal-pos $+$ cp-bif                              & $0.083$          & $0.076$          & $0.746$
  & $0.058$          & $0.760$          & $70.3^\circ$ \\
\synvaAlign{} (aligned-$60^\circ$)                & $0.092$          & $0.075$          & $0.750$
  & $0.063$          & $0.768$          & $69.2^\circ$ \\
aligned-$45^\circ$                                & $0.096$          & $0.081$          & $0.775$
  & $0.067$          & $0.772$          & $70.9^\circ$ \\
anti-curl-back                                    & $0.104$          & $0.091$          & $0.803$
  & $0.078$          & $0.830$          & $77.3^\circ$ \\
\midrule
\multicolumn{7}{l}{\textit{TreeFlowNet variants with physiological auxiliary losses}} \\
physio-wide ($\theta^\star{=}120^\circ$)          & $0.213$          & $0.228$          & $0.943$
  & $0.195$          & $0.933$          & $122.6^\circ$ \\
physio-wide, no-warp ($\theta^\star{=}120^\circ$) & $0.214$          & $0.229$          & $0.941$
  & $0.196$          & $0.936$          & $122.8^\circ$ \\
\synvaPhys{} ($\theta^\star{=}50^\circ$)          & $0.198$          & $0.214$          & $0.954$
  & $0.188$          & $0.946$          & $106.4^\circ$ \\
\synvaCombined{} (combined)                       & $0.202$          & $0.214$          & $0.951$
  & $0.187$          & $0.934$          & $105.6^\circ$ \\
\midrule
\multicolumn{7}{l}{\textit{Transformer variant (flat self-attention, no GNN message passing)}} \\
\synvaPhysTx{} ($\theta^\star{=}50^\circ$)        & $0.191$          & $0.155$          & $0.957$
  & $0.139$          & $0.966$          & $101.1^\circ$ \\
\midrule
Real training trees                               & --               & $0.028$          & $0.525$
  & --               & --               & $62.2^\circ$ \\
\bottomrule
\end{tabular}}
\end{table*}

Validation MMD selects \synvaDepth{} ($0.071$) just ahead of \synvaBase{} ($0.072$). The two differ by far less than the standard deviation over validation draws ($0.014$), and on the test trees \synvaBase{} reaches a slightly lower unconditional MMD ($0.065$ vs.\ $0.066$), so both configurations perform on par. All other variants of the first block lose between $0.008$ and $0.033$ in validation MMD. Under ground-truth topology \synvaDepth{} again reaches the lowest test MMD ($0.053$) and $1$-NNA ($0.723$). \synvaBase{} and \synvaDepth{} also
come closest to the bifurcation angle of the test trees ($62.2^\circ$) with $68.0^\circ$ and $67.9^\circ$, while the other modifications of the first block move it further away ($69.2^\circ$ to $77.3^\circ$).
The physiological block moves the bifurcation angle away from the real trees. With $\theta^\star{=}120^\circ$, the angle term and the sibling-cosine term agree, and the angle reaches $123^\circ$. With $\theta^\star{=}50^\circ$, the two terms pull in opposite directions, and the angle settles at $106^\circ$,close to $106.3^\circ$, the angle at which the gradients of the two terms cancel. The sibling-cosine target of $120^\circ$, which lies far above the test-tree angle ($62.2^\circ$), therefore dominates the block, and the block does not test the lower-angle target on its own. MMD rises to $0.19$--$0.23$ in both settings, which is why we exclude this block from the main comparison. Under the same physiological losses, the Transformer variant \synvaPhysTx{} reaches a lower MMD than \synvaPhys{} ($0.155$ vs.\ $0.214$ with sampled topology), but the highest $1$-NNA of all variants (sampled topology: $0.957$, GT topology: $0.966$) and the lowest radius-histogram similarity ($\cos_\text{radius}$ $0.634$ vs.\ $0.856$ with sampled topology, not shown in the table), so a nearest-neighbor classifier still separates its trees from real ones almost perfectly.

\subsubsection{Baselines}
\label{sec:SM_SynVA-V1_Results_Baselines}
\Cref{tab:SM_baseline_sd} completes \cref{tab:synva_v1_main} with the standard deviations of the
histogram and topology metrics. \Cref{tab:SM_baseline_robust} tests whether the ranking depends
on the bandwidth or on the node spacing. With a fixed bandwidth (the median distance among the test trees, scaled by $0.5$, $1$ or $2$) and with the $12$ spacing-invariant features alone, SynVA-V1 keeps the lowest MMD. We also resample the test trees $1000$ times, with replacement for MMD and as subsets of $80\,\%$ for COV and $1$-NNA, which duplicated trees would distort. SynVA-V1 reaches a lower MMD than VesselGPT in $92.8\,\%$ of the resamples ($98.6\,\%$ with fixed bandwidth), a higher COV in $80.9\,\%$ and a $1$-NNA closer to $0.5$ in $82.0\,\%$; against PartVessel and Chebyshev, SynVA-V1 wins MMD and $1$-NNA in every resample. \Cref{tab:SM_vgpt_copies} follows the near-copies of
VesselGPT over training. Their share grows from $3.1\,\%$ after $100$ epochs to $39.4\,\%$ at the checkpoint with the lowest training loss (epoch $998$), while the recipe of the paper, which stops early on the validation likelihood, copies $0.5\,\%$ of its samples.

SynVA-V1 reaches a lower $\text{KL}_\text{degree}$ and Spec.\ than real training trees (\cref{tab:SM_baseline_sd}). Both metrics compare histograms pooled over all trees of a draw. Nucleus sampling with top-$p{=}0.95$ concentrates the sampled topologies on frequent sequences, so the pooled degree histogram of $89$ samples fluctuates less around the mode than that of $89$ real trees. The same truncation produces smaller trees ($73.8$ vs.\ $92.3$ nodes). With top-$p{=}1.0$, the trees grow to $86.0$ nodes, and $\text{KL}_\text{degree}$ ($0.0012$) and Spec.\ ($0.0012$) rise to the level of real trees. We set top-$p{=}0.95$ before training and did not tune it. On the validation trees, top-$p{=}1.0$ reaches a lower MMD ($0.061$ vs.\ $0.071$), and on the test trees a higher one ($0.073$ vs.\ $0.066$). Both differences lie within the standard deviation over draws ($0.014$ to $0.015$ for top-$p{=}1.0$).

\begin{table*}[htb]
\centering
\small
\setlength{\tabcolsep}{4pt}
\mdseries
\caption{Histogram and topology metrics of \cref{tab:synva_v1_main} with standard deviations over
$10$ draws of $89$ samples, and the mean node count. The first row draws $89$ real training trees
and shows what a perfect sampler of the training distribution reaches.}
\label{tab:SM_baseline_sd}
\scalebox{0.85}{
\begin{tabular}{l cccccc}
\toprule
Method & $\cos_\text{radius}\uparrow$ & $\cos_\text{length}\uparrow$ & $\cos_\text{tortuosity}\uparrow$
& $\text{KL}_\text{degree}\downarrow$ & Spec.$\downarrow$ & Nodes \\
\midrule
Real training trees & $0.976 \pm 0.008$ & $0.980 \pm 0.007$ & $0.9997 \pm 0.0001$
  & $0.0012 \pm 0.0005$ & $0.0014 \pm 0.0004$ & $92.3$ \\
\midrule
VesselGPT  & $\mathbf{0.944 \pm 0.016}$ & $0.922 \pm 0.022$ & $0.919 \pm 0.038$
  & $0.0060 \pm 0.0017$ & $0.0056 \pm 0.0011$ & $68.7$ \\
PartVessel & $0.671 \pm 0.019$ & $0.499 \pm 0.010$ & $0.988 \pm 0.001$
  & $0.0005 \pm 0.0003$ & $0.0018 \pm 0.0004$ & $75.8$ \\
Chebyshev  & $0.735 \pm 0.017$ & $0.345 \pm 0.002$ & $0.996 \pm 0.001$
  & $0.0038 \pm 0.0005$ & $0.0029 \pm 0.0006$ & $113.9$ \\
SynVA-V1   & $0.935 \pm 0.028$ & $\mathbf{0.955 \pm 0.020}$ & $\mathbf{0.9994 \pm 0.0004}$
  & $\mathbf{0.0001 \pm 0.0001}$ & $\mathbf{0.0007 \pm 0.0002}$ & $73.8$ \\
\bottomrule
\end{tabular}}
\end{table*}

\begin{table}[htb]
\centering
\small
\setlength{\tabcolsep}{3pt}
\mdseries
\caption{MMD on the test trees ($n{=}89$) under other kernel bandwidths and without the four
spacing-dependent features. $\sigma_\text{test}$ is the median pairwise distance among the test
trees. Values are means over $10$ draws.}
\label{tab:SM_baseline_robust}
\scalebox{0.85}{
\begin{tabular}{l ccccc}
\toprule
& Median & \multicolumn{3}{c}{Fixed bandwidth} & $12$ features \\
\cmidrule(lr){3-5}
Method & heuristic & $0.5\,\sigma_\text{test}$ & $\sigma_\text{test}$ & $2\,\sigma_\text{test}$
& (median) \\
\midrule
VesselGPT  & $0.091$ & $0.118$ & $0.102$ & $0.039$ & $0.106$ \\
PartVessel & $0.336$ & $0.393$ & $0.336$ & $0.129$ & $0.308$ \\
Chebyshev  & $0.217$ & $0.228$ & $0.230$ & $0.101$ & $0.141$ \\
SynVA-V1   & $\mathbf{0.066}$ & $\mathbf{0.091}$ & $\mathbf{0.062}$ & $\mathbf{0.021}$
  & $\mathbf{0.073}$ \\
\bottomrule
\end{tabular}}
\end{table}

\begin{table}[htb]
\centering
\small
\mdseries
\caption{Near-copies of VesselGPT over training. The threshold is the fifth percentile of the distances from the test trees to their nearest training tree, so a model whose samples lie as far from the training trees as the test trees yields about $5\,\%$.}
\label{tab:SM_vgpt_copies}
\scalebox{0.85}{
\begin{tabular}{l c}
\toprule
VesselGPT checkpoint & Near-copies \\
\midrule
Code recipe, epoch $100$ & $3.1\,\%$ \\
Code recipe, epoch $300$ & $23.5\,\%$ \\
Code recipe, epoch $793$ (lowest validation MMD) & $35.3\,\%$ \\
Code recipe, epoch $998$ (lowest training loss, main table) & $39.4\,\%$ \\
Paper recipe (early stopping) & $0.5\,\%$ \\
\bottomrule
\end{tabular}}
\end{table}

\section{GUI for Ostium Selection}
\label{sec:SM_GUI}

\begin{figure*}[htb]
  \centering
  \vspace{-5mm}
  \includegraphics[width=0.8\textwidth]{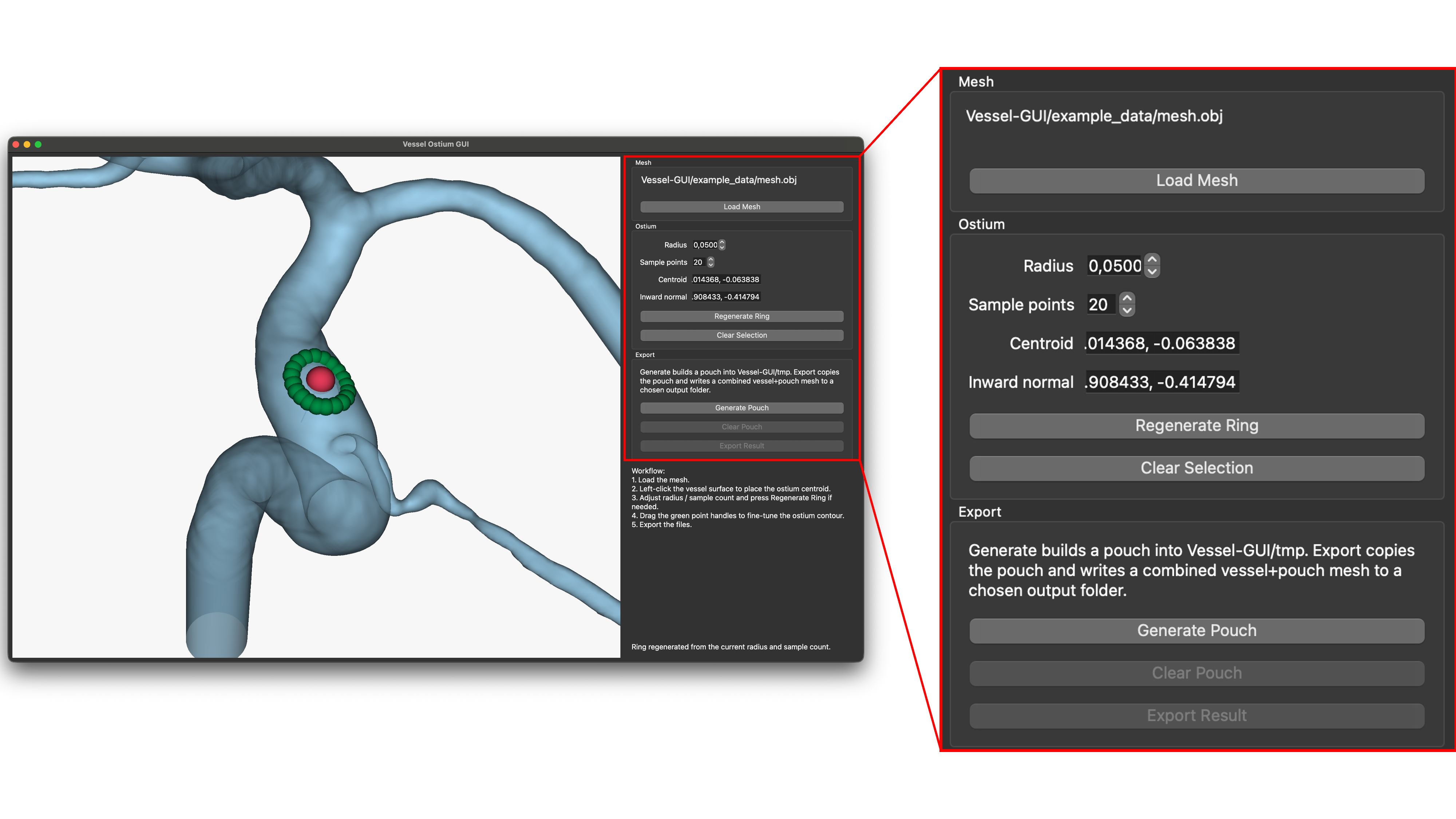}
  \vspace{-8mm}
  \caption{GUI for ostium placement and contour editing. The tool supports interactive ring adjustment, vessel cutting, and exports the ostium centroid and inward normal for downstream processing.}
  \label{fig:SM_GUI}
\end{figure*}

The graphical user interface (GUI) shown in \cref{fig:SM_GUI} enables the user to select a vertex on the vessel surface to define the ostium centroid. A surface normal is estimated from the local geometry, and an editable circular contour with a user-defined radius is generated at this location. The vessel surface is subsequently clipped using \texttt{vtkSelectPolyData} and \texttt{vtkClipPolyData}, resulting in an opening that defines the attachment site of the aneurysm.

The resulting ostium and associated normal vector serve as inputs for generating an aneurysm pouch using the models introduced in this work. To ensure watertight connectivity between the vessel and the synthesized aneurysm, boundary loops are extracted from both geometries and stitched together. The resulting junction is further refined via an \(N\)-hop neighborhood averaging scheme to ensure smooth geometric transitions.
\section{SynVA-A1}
\label{sec:SM_SynVA-A1}

The SynVA-A1 model is based on the concept introduced in AneuG~\cite{ding2025two}, and extends it to conditionally generate aneurysm geometries on pre-existing healthy vascular structures. Conditioning is performed on the ostium as well as on local vessel geometry and morphological parameters, enabling the synthesis of realistic aneurysmal deformations directly on anatomically plausible vessel segments. This design is motivated by the observation that, in nature, aneurysms do not arise in isolation but develop from previously healthy vessels. Consequently, modeling aneurysm formation as a conditional transformation of healthy vasculature not only aligns with the underlying biological process but also enables the systematic conversion of arbitrary healthy vessel geometries into pathological counterparts.

\subsection{Data Preprocessing}
\label{sec:SM_SynVA-A1_Data}
SynVA-A1 is trained on a real-world dataset from which a canonical mesh is first constructed as the basis for graph harmonic deformation~(GHD) fitting. For this purpose, the aneurysm submeshes are used.
Prior to fitting, all target aneurysm sacs are first brought into a shared coordinate system through alignment of their respective ostia. Within this shared space, a canonical aneurysm template is constructed from the entire training distribution by applying a ray-casting procedure in which rays are emitted from a common origin and intersect all aligned aneurysm geometries. For each ray direction, the median intersection distance across all aligned training samples is computed, and this statistic defines the vertex position of the canonical mesh along that ray. In this way, the resulting canonical template constitutes a robust statistical average of the dataset geometry under consistent ostium alignment.
Subsequently, each target aneurysm is rigidly aligned to this canonical mesh. First, a translation is applied to map the target ostium center to the canonical ostium center. Second, a rotation is performed to align the ostium normals, ensuring consistent orientation across all samples. The resulting transformation matrix $\bm{H}$ is stored for postprocessing steps. After alignment, both canonical and target meshes are normalized.

For SynVA-A1 training, the GHD fitting checkpoint provides the target representation. The target vector consists of the fitted GHD coefficients ($K=144$, three coordinate channels) and the fitted scale. The fitted rotation
and translation are not predicted by SynVA-A1; they are used only to define the local GHD coordinate frame and to decode meshes for the training losses. Thus, the encoder does not receive a raw mesh, but the normalized GHD target from the fitting step.

The condition input is extracted from the same case and is centered around the
ostium. SynVA-A1 therefore does not use the complete vessel tree as input, but
only the local vessel geometry that is relevant for aneurysm attachment. To
make this vessel patch consistent with the fitted GHD representation, the
candidate vessel points are first mapped into the same local coordinate frame
as the fitted aneurysm. Distances are then measured from the ostium-ring
centroid, and a neighborhood around the ostium is retained. The neighborhood
radius is chosen from the ostium size and the local distance distribution, so
that it covers the neck region and nearby parent-vessel surface without
including the entire vessel. From this local patch, 256 points are selected by
farthest-point sampling. If the local selection contains too few points, the
full candidate surface is used as a fallback. The remaining conditioning inputs are the ordered ostium ring, compact ostium parameters consisting of center, normal, radius, and eccentricity, dense attachment samples from the segmented ostium region, and the morphology
descriptor vector. The morphology condition consists of $18$ scalar parameters, the independent features $A_\mathrm{A}$, $V_\mathrm{A}$, $A_\mathrm{CH}$, $V_\mathrm{CH}$, $D_\mathrm{max}$, $H_\mathrm{max}$, $W_\mathrm{max}$, $H_\mathrm{ortho}$, and $W_\mathrm{ortho}$, the ostium parameters $A_\mathrm{O1}$, $A_\mathrm{O2}$, $N_\mathrm{max}$, and $N_\mathrm{avg}$, and the derived indices $AR_1$, $AR_2$, $EI$, $NSI$, and $UI$. GHD targets and continuous conditioning variables are normalized with training-set statistics.

\subsection{Architecture}
\label{sec:SM_SynVA-A1_Architecture}
Analogous to AneuG~\cite{ding2025two}, SynVA-A1 consists of two coupled components. First, each training aneurysm is represented via a GHD fit of the aneurysm sac and neck region in a shared canonical eigenspace. Second, SynVA-A1 learns a conditional generative model over the resulting GHD parameters, conditioned on local vessel geometry, ostium information, and global morphological descriptors.

\subsubsection{GHD Fitting}
\label{sec:SM_SynVA-A1_Architecture_GHD}
Our fitting procedure builds upon the GHD representation introduced in AneuG~\cite{ding2025two}, but adapts it from full aneurysm–vessel segments, which include the parent artery and multiple openings, to a localized representation focused on the aneurysm sac. In the present setting, the fitting target is defined by the aneurysm sac in combination with a single ostium, which serves as the sole anatomical interface to the parent vessel.

We retain the core GHD representation from AneuG. Each case is fitted by deforming the canonical sac in a shared graph-harmonic basis and by optimizing a global similarity transform. The target of the fit is therefore not a raw mesh with arbitrary topology, but a compact set of GHD coefficients in the common canonical eigenspace. In our implementation this basis contains $K=144$ graph-harmonic modes.

The general shape terms are also inherited from the original AneuG fitting objective. We use surface CD and surface-normal matching to align the deformed canonical sac with the target sac. We further keep the standard mesh regularizers used in AneuG: Laplacian smoothing, edge-length regularization, normal-consistency regularization, and local rigidity. These terms control the overall sac shape and discourage noisy or folded deformations.

The main methodological change is the treatment of the ostium. Original AneuG supervises several vessel openings and, in its full-vessel setting, can use differentiable occupancy and differentiable centerline losses to constrain the parent-vessel segment. Since SynVA-A1 does not fit the parent vessel, these full vessel losses are disabled in our sac-and-neck fitting mode. Instead, we replace the multi-opening objective by a single-ostium objective. The ostium is represented by an ordered boundary ring and by a reconstructed opening patch from the ostium checkpoint. This allows us to supervise the neck directly: the opening position is matched to the target, the opening normal is aligned using a robust ring-based normal estimate, and an area term prevents collapse of the neck opening. Additional overlap, boundary-smoothness, and relative-volume terms stabilize the vessel--aneurysm interface and the global sac size.

\subsubsection{Conditional Aneurysm Generator}
\label{sec:SM_SynVA-A1_Architecture_Generator}
SynVA-A1 models a conditional prior over the fitted GHD parameters of the aneurysm sac and neck region. The generator therefore does not predict mesh vertices directly, but the GHD representation produced by the fitting step. For each case, the target vector contains the $144\times3$ GHD coefficients and the fitted scale. Compared with the AneuG’s conditional GHD-VAE backbone, SynVA-A1 changes mainly what the model is conditioned on and how the decoded samples are supervised: it adds local vessel--ostium conditioning, morphology conditioning, geometry-aware ostium supervision, and prior-path calibration for direct samples from the VAE prior.

The condition is built from the local anatomy around the prescribed ostium. The local vessel patch, represented by 256 points, is encoded with a PointNet-style MLP branch followed by max pooling. Low-dimensional ostium parameters, consisting of center, normal, radius, and eccentricity, are encoded with a small MLP. The ordered ostium ring is resampled to 20 points, flattened, and encoded with a separate MLP so that the network receives the neck boundary as an explicit geometric signal. These branch embeddings are fused into a 96-dimensional vessel--ostium feature. The selected morphology descriptor vector is appended to this feature, and the resulting condition vector is projected to a 64-dimensional condition embedding. Dense samples from the segmented ostium region are prepared for the same case, but in the final model they are used for interface supervision rather than as an additional condition branch.

The generative backbone is a conditional GHD-VAE. During training, the encoder receives the normalized fitted GHD target and maps it to a Gaussian latent distribution. A latent code is sampled with the standard reparameterization trick and passed to the decoder together with the condition embedding. The decoder then predicts the normalized GHD coefficients and scale. Both encoder and decoder use a linear input projection and one residual MLP block; the hidden dimension is 256 and the latent dimension is 108. At inference time, the fitted target is absent. A latent code is sampled from the standard normal prior and decoded under the same vessel, ostium, and morphology condition.

The additional supervision is tied to this inference setting. The decoded GHD parameters are converted back to a mesh during training, which allows losses at the vessel--aneurysm interface to constrain the generated neck geometry. Moreover, prior-path calibration explicitly decodes samples from $\bm{z}\sim\mathcal{N}(\bm{0},\bm{I})$ during training, matching them to case-level and batch-level target statistics. Thus, the model learns not only to reconstruct fitted targets through the posterior path, but also to produce plausible conditional samples through the same prior path used at test time.

\subsection{Training Details}
\label{sec:SM_SynVA-A1_Training}

\subsubsection{GHD Fitting}
\label{sec:SM_SynVA-A1_Training_GHD}
The GHD fits are optimized independently for each prepared aneurysm case. We
use the precomputed canonical basis with $K=144$ and a single ostium. The
optimization uses AdamW with learning rate $10^{-3}$ for 3000 epochs
and a step scheduler with step size 2500 and decay factor 0.75. All loss weights
increase linearly during the first 300 epochs. The active loss weights are listed in
\cref{tab:SM_SynVA-A1-GHDLossWeights}.

\begin{table}[htb]
\centering
\caption{Loss weights used for the SynVA-A1 GHD fitting.}
\label{tab:SM_SynVA-A1-GHDLossWeights}
\scalebox{0.8}{
\begin{tabular}{lclc}
\toprule
Term & Weight & Term & Weight \\
\midrule
Surface Chamfer & 20.0 & Surface normals & 16.0 \\
Laplacian & 0.1 & Edge length & 0.1 \\
Normal consistency & 350.0 & Local rigidity & 100.0 \\
Opening position & 10.0 & Opening normal & 0.1 \\
Opening area & 1.0 & Opening overlap & 1.0 \\
Boundary smoothness & 0.5 & Relative volume & 0.05 \\
\bottomrule
\end{tabular}}
\end{table}

\subsubsection{SynVA-A1 Training}
\label{sec:SM_SynVA-A1_Training_A1}
We use the same stratified and balanced split as the other models in this
work: 530 training cases, 132 validation cases, and 100 held-out test cases. The CVAE is trained for 4000 epochs with batch size 96 using AdamW with learning rate $10^{-4}$ and weight decay $10^{-4}$. The latent dimension is 108, the hidden dimension is 256, the condition embedding dimension is 64, and the vessel-condition embedding dimension is 96. The VAE and the vessel--ostium condition network are optimized jointly, without condition dropout in the selected run. We use a KL warm-up of 1000 epochs, free bits of 0.02, gradient clipping with maximum norm 1.0, validation every 50 epochs, and select the checkpoint with the lowest validation total loss. The selected SynVA-A1 checkpoint is from epoch 2550.

The CVAE is trained with three complementary loss groups. First, the posterior path reconstructs the fitted GHD target. During this step, the encoder sees the normalized GHD coefficients and scale of a training aneurysm, samples a latent code with the reparameterization trick, and the decoder reconstructs the same GHD vector under the case-specific condition. This reconstruction is supervised in coefficient space, in the scale value, and after GHD decoding in vertex space. The KL term is warmed up during training and uses free bits to prevent posterior collapse.

Second, we add geometry-aware supervision around the ostium. For this loss group, the reconstructed GHD vector is decoded to a mesh, transformed to the case frame, and aligned to the target ostium ring in the same way as in the later placement step. We then evaluate four local interface losses. The attachment loss measures the nearest-surface distance from dense samples of the segmented ostium attachment region to the generated aneurysm and therefore encourages the generated neck to cover the prescribed attachment area. The sac-center loss compares the center of the generated aneurysm sac, excluding the opening vertices, with the center of the fitted target sac and stabilizes the global placement of the aneurysm relative to the neck. The side loss uses the ostium normal and penalizes generated sacs whose center lies on the wrong side of the ostium plane, which prevents growth through the vessel wall. The opening-center loss compares the mean position of the generated opening ring with the target ostium-ring center. In the selected model, these terms are combined as
$\mathcal{L}_{\mathrm{geo}}=5.0\,\mathcal{L}_{\mathrm{attach}}+
10.0\,\mathcal{L}_{\mathrm{center}}+
2.0\,\mathcal{L}_{\mathrm{side}}+
10.0\,\mathcal{L}_{\mathrm{open}}$.
Together, these losses ensure that a good reconstruction is not only close in GHD space, but also compatible with the prescribed vessel--aneurysm interface.

Third, SynVA-A1 explicitly trains the prior sampling path. This is important because inference does not use the encoder; it samples $\bm{z}_{\mathrm{prior}}\sim\mathcal{N}(\bm{0},\bm{I})$ and decodes this latent vector under the given condition. We therefore perform the same operation during training in each mini-batch. The resulting prior-decoded GHD vector is encouraged to stay close to the paired target, while its batch mean and batch standard deviation are matched to those of the fitted targets. The mean term keeps the samples centered on the training distribution, and the standard-deviation term discourages collapse to an average aneurysm. Overall, the objective contains reconstruction, geometry, and prior-calibration terms, with the weights listed in \cref{tab:SM_SynVA-A1-LossWeights}.

\begin{table}[t]
\centering
\caption{CVAE loss weights used for SynVA-A1.}
\label{tab:SM_SynVA-A1-LossWeights}
\scalebox{0.8}{
\begin{tabular}{lclc}
\toprule
Term & Weight & Term & Weight \\
\midrule
Coeff. MSE & 1.0 & Attachment samples & 5.0 \\
Scale Huber & 1.0 & Sac center & 10.0 \\
Vertex MSE & 250.0 & Side & 2.0 \\
KL & 0.01 & Opening center & 10.0 \\
Prior Huber & 0.25 & Prior mean/std & 0.25 / 8.0 \\
\bottomrule
\end{tabular}}
\end{table}

\subsection{Experimental Setting}
\label{sec:SM_SynVA-A1_Experiments}
All ablations are trained and evaluated with the same splits and the same
100-case test set. We compare a sequence of variants that progressively add
the proposed conditioning and training components. \textbf{AneuG-Base} is the
AneuG-style conditional GHD-VAE baseline trained on the fitted aneurysm sac and
neck GHD parameters. It does not use the proposed local vessel--ostium
condition, morphology condition, or prior-path calibration. \textbf{AneuG-Cond}
adds the local vessel--ostium condition, including the local vessel point set,
low-dimensional ostium parameters, and ordered ostium ring.
\textbf{AneuG-Cond-Morph} extends this variant with the selected morphology
descriptor vector. \textbf{AneuG-Cond-Prior} instead adds prior-path
calibration to AneuG-Cond but omits the morphology condition, which isolates
the effect of calibrating direct samples from the VAE prior. the effect of calibrating direct samples from the VAE prior. All variants except AneuG-Base use the geometry-aware ostium supervision. The variants with prior-path calibration use a KL weight of $0.01$ and $0.02$ free bits; AneuG-Cond and AneuG-Cond-Morph use $0.0002$ and $0.01$.

\textbf{SynVA-A1} is the selected model and combines local vessel--ostium
conditioning, morphology conditioning, geometry-aware ostium supervision, and
prior-path calibration. \textbf{SynVA-A1-Ostium} uses the same architecture as
SynVA-A1, but increases the ostium/interface supervision weights and adds a nearest-distance term with weight $10$ to test whether stronger local neck emphasis improves the generated attachment
geometry. 
Evaluation uses the paired surface CD with $10{,}000$ sampled surface points and the mean absolute differences of the morphology descriptors. Lower values are better for both.

\subsection{Additional Results}
\label{sec:SM_SynVA-A1_Results}

\begin{table*}[htb]
\centering
\caption{Paired aneurysm reconstruction results on the 100 held-out test cases. Values are mean absolute differences with standard deviation, except for CD, which is computed from 10k sampled surface points. Lower is better; the best mean value in each row is highlighted.}
\label{tab:SM_SynVA-A1-ResultsSummary}
\scalebox{0.78}{
\begin{tabular}{lcccccc}
\toprule
Mean Absolute Difference & AneuG-Base & AneuG-Cond & AneuG-Cond-Morph & AneuG-Cond-Prior & SynVA-A1 & SynVA-A1-Ostium \\
\midrule
$A_\mathrm{A}$ & 0.2032 $\pm$ 0.2951 & 0.1974 $\pm$ 0.2960 & 0.1711 $\pm$ 0.2807 & 0.1638 $\pm$ 0.2661 & 0.1560 $\pm$ 0.2599 & \textbf{0.1555 $\pm$ 0.2630} \\
$A_\mathrm{CH}$ & 0.2015 $\pm$ 0.2931 & 0.1962 $\pm$ 0.2984 & 0.1744 $\pm$ 0.2871 & 0.1625 $\pm$ 0.2684 & \textbf{0.1539 $\pm$ 0.2631} & 0.1543 $\pm$ 0.2682 \\
$D_{\max}$ & 0.0987 $\pm$ 0.1022 & 0.1068 $\pm$ 0.1109 & 0.1079 $\pm$ 0.1120 & \textbf{0.0842 $\pm$ 0.0992} & 0.0842 $\pm$ 0.0992 & 0.0881 $\pm$ 0.1008 \\
$H_{\max}$ & 0.1044 $\pm$ 0.1004 & 0.1087 $\pm$ 0.1068 & 0.1037 $\pm$ 0.1053 & 0.0814 $\pm$ 0.0914 & \textbf{0.0796 $\pm$ 0.0905} & 0.0826 $\pm$ 0.0924 \\
$H_{\mathrm{ortho}}$ & 0.1008 $\pm$ 0.1012 & 0.0998 $\pm$ 0.1009 & 0.0929 $\pm$ 0.0955 & 0.0746 $\pm$ 0.0834 & \textbf{0.0715 $\pm$ 0.0821} & 0.0743 $\pm$ 0.0844 \\
$V_\mathrm{A}$ & 0.0279 $\pm$ 0.0510 & 0.0267 $\pm$ 0.0510 & 0.0220 $\pm$ 0.0480 & 0.0225 $\pm$ 0.0462 & 0.0214 $\pm$ 0.0449 & \textbf{0.0210 $\pm$ 0.0454} \\
$V_\mathrm{CH}$ & 0.0279 $\pm$ 0.0510 & 0.0267 $\pm$ 0.0510 & 0.0220 $\pm$ 0.0480 & 0.0225 $\pm$ 0.0462 & 0.0214 $\pm$ 0.0449 & \textbf{0.0210 $\pm$ 0.0454} \\
$W_{\max}$ & 0.0698 $\pm$ 0.0837 & 0.0685 $\pm$ 0.0862 & 0.0615 $\pm$ 0.0857 & 0.0592 $\pm$ 0.0796 & 0.0574 $\pm$ 0.0805 & \textbf{0.0569 $\pm$ 0.0806} \\
$W_{\mathrm{ortho}}$ & 0.0646 $\pm$ 0.0818 & 0.0674 $\pm$ 0.0855 & 0.0708 $\pm$ 0.0877 & 0.0592 $\pm$ 0.0771 & \textbf{0.0567 $\pm$ 0.0752} & 0.0603 $\pm$ 0.0783 \\
\midrule
CD & 0.1010 $\pm$ 0.0721 & 0.0981 $\pm$ 0.0735 & 0.0912 $\pm$ 0.0672 & 0.0872 $\pm$ 0.0652 & 0.0864 $\pm$ 0.0652 & \textbf{0.0862 $\pm$ 0.0649} \\
\bottomrule
\end{tabular}}
\end{table*}

\begin{figure}[htbp]
  \centering
  \includegraphics[width=\linewidth]{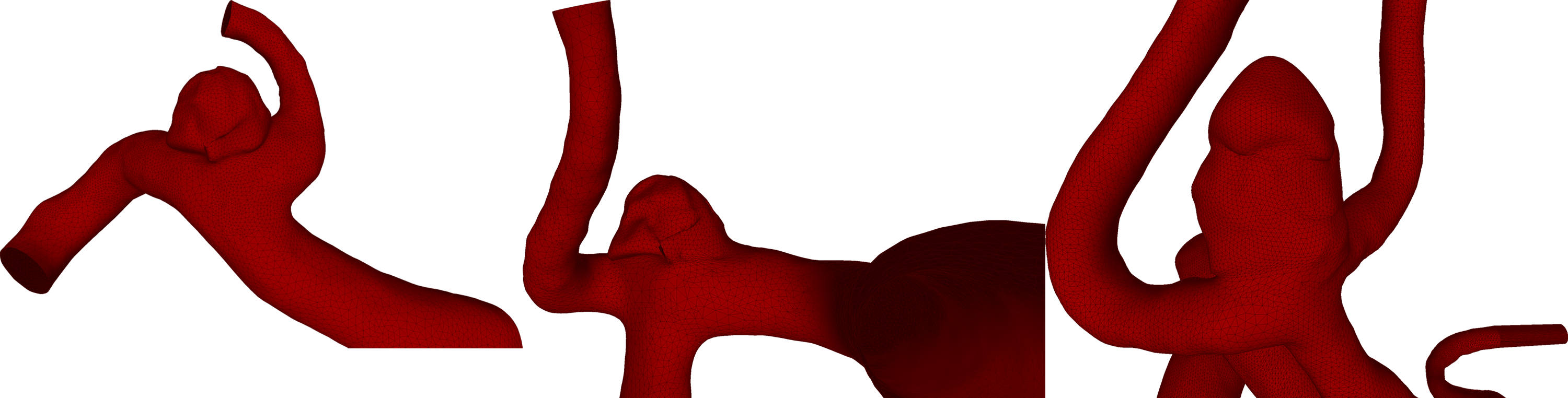}
  \vspace{-0.2cm}
  \caption{Representative SynVA-A1 failure cases.}
  \label{fig:SM_SynVA-A1-FailureCases}
\end{figure}

The ablation~(\cref{tab:SM_SynVA-A1-ResultsSummary}) shows that the individual additions affect different aspects of the generated aneurysms. Morphology conditioning and prior-path calibration each lower the paired CD, and prior-path calibration substantially improves the main size descriptors. SynVA-A1 gives the lowest height and orthogonal-width errors, and remains close to the best model for the other morphology measures. Therefore, we select SynVA-A1 as the final model because it provides the most balanced trade-off across the evaluated criteria. It combines local vessel--ostium conditioning, morphology conditioning, geometry-aware ostium supervision, and prior-path calibration, and it avoids optimizing for a single descriptor at the expense of the overall generated shape. The stronger ostium variant slightly improves surface-area, volume, and maximum-width errors, but the increased interface emphasis led to sharper and more locally jagged surfaces in visual inspection~(\cref{fig:SM_SynVA-A1-FailureCases}). Thus, SynVA-A1 is selected as the best compromise between global sac shape, morphology control, and stable ostium attachment, rather than as the model with the largest number of row-wise best values.

Qualitatively, the generated pouches usually attach to the prescribed ostium,
but representative failure cases can still exhibit sharp edges and locally
jagged surfaces.

\section{SynVA-A2}
\label{sec:SM_SynVA-A2}
MeshAnything and MeshAnything V2 are designed to generate triangle meshes from complete 3D inputs, such as point clouds, which already implicitly encode the full target geometry~\cite{chen2025meshanything,chen2025meshanythingv2}. This paradigm is not directly applicable to anatomy-driven aneurysm synthesis, where the aneurysm geometry must be generated de novo from a pre-existing vessel without access to its full shape representation.

\begin{figure*}[htb]
  \centering
  \includegraphics[width=0.8\textwidth]{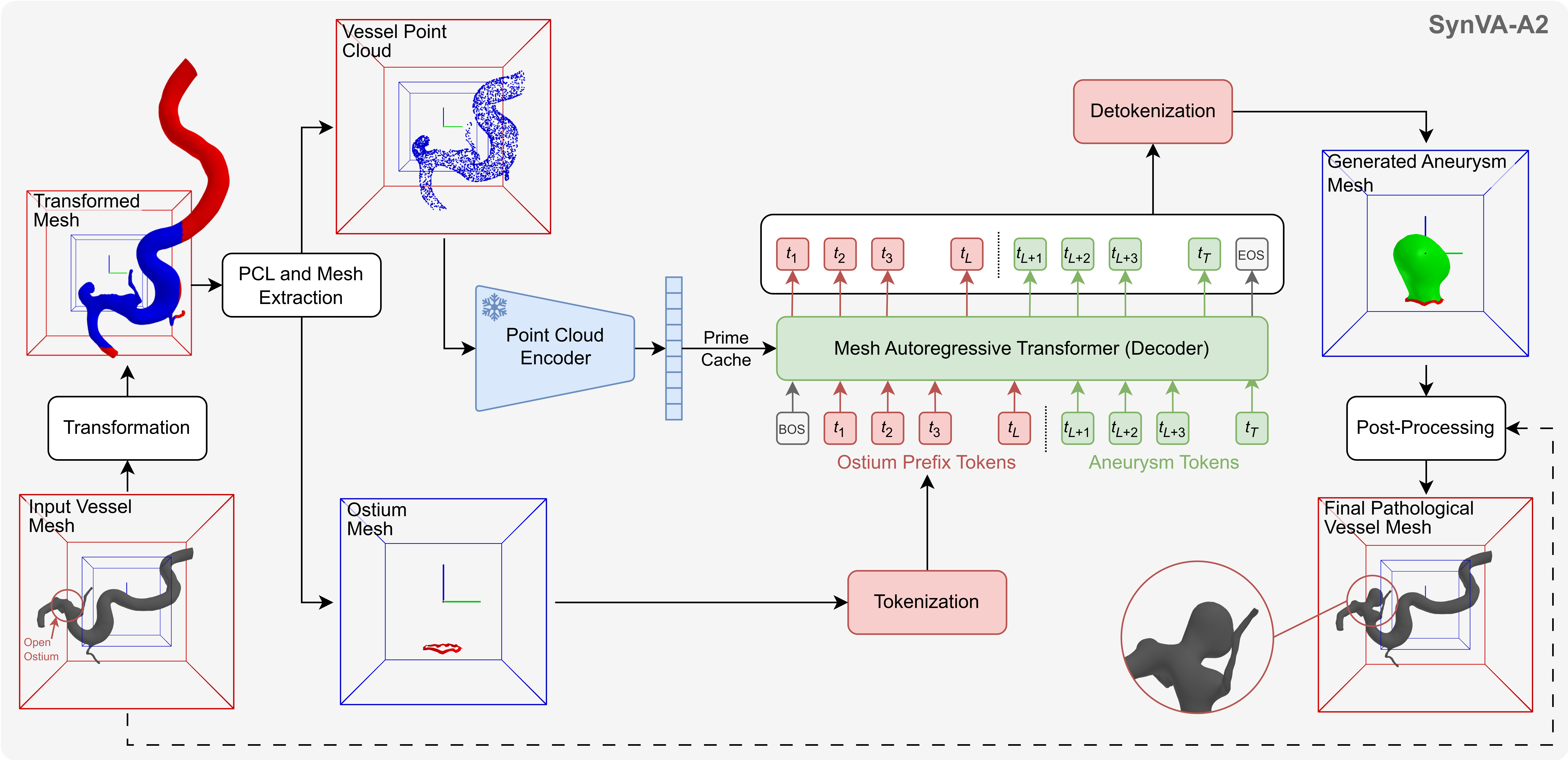}
  \caption{SynVA autoregressive aneurysm generation. A standardized healthy vessel is converted into an incomplete point cloud (conditioning) and an ostium mesh (prefix). Conditioned on both, a decoder-only Transformer autoregressively generates aneurysm mesh tokens, enabling extrapolative growth from the ostium without requiring a full target shape. The output is detokenized, postprocessed, and stitched back to the vessel.}
  \label{fig:SM_SynVA-A2}
\end{figure*}

To address this limitation, we adopt the core autoregressive mesh decoding framework of MeshAnything V2 and extend it to support localized, conditional generation from partial geometric context. Specifically, we introduce (i) ostium-aware prefix conditioning, where ostium boundary tokens are injected into the autoregressive sequence, and (ii) conditioning on incomplete point clouds, representing only the surrounding vessel geometry. During generation, conditioning embeddings derived from the point cloud are first used to initialize the Transformer state, followed by forced decoding of the ostium prefix tokens. The model then autoregressively generates the aneurysm mesh conditioned on this partial context (see \cref{fig:SM_SynVA-A2}).

This formulation enables mesh synthesis that is spatially anchored to existing structures, allowing aneurysms to be generated directly from the ostium rather than requiring a full target shape as input. Consequently, our approach extends MeshAnything V2 from global shape reconstruction to structure-aware, incremental mesh generation, which is essential for anatomically consistent modeling. Further implementation details are provided below.

\subsection{Data Preprocessing}
\label{sec:SM_SynVA-A2_Data}
Both synthetic and real-world data are used for training and undergo an identical preprocessing pipeline. The following steps are therefore applied consistently across both datasets.

First, samples whose aneurysm submesh has fewer than $80$ faces are discarded to remove low-resolution and geometrically unreliable cases. To comply with the maximum input size of MeshAnything (1600 faces), aneurysm meshes exceeding this limit are simplified using quadric edge collapse. During this process, boundaries, normals, and topology are preserved to maintain geometric fidelity, with particular emphasis on retaining the ostium vertices.

To align the data with the model’s input assumptions, meshes are normalized and spatially standardized, depicted by the transformation step in \cref{fig:SM_SynVA-A2}. Since MeshAnything V2 generates meshes within a unit cube $[-0.5,\,0.5]$ and the point cloud encoder operates on inputs in $[-1,\,1]$, both aneurysm and vessel submeshes are transformed to ensure consistent scale, position, and orientation. As illustrated in \cref{fig:SM_SynVA-A2}, the $[-0.5,\,0.5]$ bounding box is shown in blue, while the larger $[-1,\,1]$ region is depicted in red. Specifically, meshes are first translated such that the ostium centroid $C_\mathrm{O}$ coincides with the origin. They are then rotated so that the ostium normal vector aligns with the positive $z$-axis, enforcing a consistent growth direction. Subsequently, meshes are isotropically scaled such that the maximum ostium diameter $N_\mathrm{max}$ is fixed to $0.25$. Finally, meshes are translated by $-0.3$ along the $z$-axis to better utilize the available generation space, as synthesis is restricted to the positive $z$-direction.

For prefix-based conditioning, an explicit ostium mesh is extracted from the vessel submesh by selecting all faces that contain at least one ostium vertex. This mesh represents the boundary ring from which the aneurysm is generated. To construct the conditioning input, the vessel submesh is spatially cropped around the ostium region within a fixed bounding box of $[-1,\,1]$, under the assumption that distant structures have negligible influence on aneurysm shape. As shown in \cref{fig:SM_SynVA-A2}, this corresponds to removing distal regions~(red part of transformed mesh) while retaining the local vessel neighborhood~(blue part of transformed mesh) for anatomically meaningful conditioning. From this cropped region, a point cloud of 20k surface samples is generated.

\subsection{Architecture}
\label{sec:SM_SynVA-A2_Architecture}
Building upon MeshAnything V2~\cite{chen2025meshanythingv2}, we introduce targeted conditioning mechanisms that enable mesh generation from partial geometric context, allowing aneurysm geometries to be synthesized directly from a pre-existing vessel. Specifically, our approach combines (i) conditioning on incomplete point clouds of the healthy vessel and (ii) explicit ostium-prefix token injection, enabling autoregressive mesh growth from a predefined boundary. This extends MeshAnything V2 from full-shape reconstruction to structure-aware, extrapolative mesh generation, which is essential for anatomically consistent aneurysm synthesis.

Both the vessel submesh (represented as a point cloud with normals) and the ostium prefix mesh serve as inputs. Meshes are first canonicalized and reordered, and subsequently tokenized using Adjacent Mesh Tokenization. Vertex coordinates are discretized with a resolution of $256$ (instead of $128$ in the original MeshAnything V2), as the lower resolution led to noticeable surface artifacts (staircasing) under our preprocessing. Consequently, pretrained weights are not used. The resulting token sequence consists of ostium prefix tokens, followed by aneurysm continuation tokens, and standard special tokens~(BOS, EOS, padding).

The conditioning point cloud is encoded using a frozen pretrained ShapeVAE encoder~\cite{zhao2023michelangelo}, producing $257$ tokens (one global and $256$ local tokens) with a latent dimension of $768$. The global token is projected to the Transformer hidden size ($1024$), while each local token is concatenated with additional shape latents derived from the ShapeVAE bottleneck and projected via a linear layer. This design combines complementary geometric representations and improves the expressiveness of the conditioning signal. The final conditioning tensor has shape $B \times 257 \times 1024$, matching the hidden size of the ShapeOPT backbone (OPT-350M~\cite{zhang2022opt}).

The autoregressive model for mesh generation is a decoder-only Transformer (OPT-350M) trained from scratch~\cite{zhang2022opt}. During training, the conditioning embeddings are provided as input embeddings, while the full token sequence (prefix + continuation) is teacher-forced. Crucially, loss masking is applied to the prefix tokens, such that optimization is restricted to the continuation, i.e., the aneurysm generation.

At inference time, the conditioning embeddings are first used to prime the Transformer cache. A BOS token is inserted, after which the ostium prefix tokens are force-fed sequentially, ensuring that the subsequent generation is conditioned on the complete boundary. The aneurysm mesh is then generated autoregressively. To prevent degenerate early termination, EOS prediction is suppressed for a fixed number of steps following the prefix. Token selection is performed via argmax decoding. After removing special tokens, the predicted sequence is deterministically dequantized and detokenized into triangle meshes.

Postprocessing removes invalid or low-quality outputs (e.g., meshes with fewer than $30$ faces, prefix-only artifacts, or degenerate triangles based on normalized area and edge-length criteria). Standard mesh repair operations are applied, including removal of duplicate or unreferenced elements, non-manifold correction, and hole filling (up to a maximum size of $20$). Normals are recomputed, and the mesh is transformed back into the original coordinate system via inverse normalization. Finally, the generated aneurysm is stitched to the vessel mesh. Ostium vertices are matched using a $k$-d tree followed by Hungarian assignment~\cite{kuhn1955hungarian}, and snapped to ensure exact alignment. The meshes are merged, duplicate vertices are consolidated (with label consistency), and the transition region is smoothed.
A generation is considered successful if the autoregressive sequence ends with an EOS token before the token budget is exhausted, in accordance with the completeness criterion of MeshAnything V2~\cite{chen2025meshanythingv2}, and the decoded mesh passes the validity checks described above. The subsequent repair fixes local defects such as small holes or non-manifold edges, but cannot complete an incomplete sequence.

\subsection{Training Details}
\label{sec:SM_SynVA-A2_Training}
We adopt the same data split as used across all models in this work. Due to preprocessing constraints and additional validation checks (ensuring both point clouds and meshes lie within their respective bounding boxes), a small number of samples are excluded.

During training, data augmentation is applied to improve generalization while preserving geometric consistency between inputs and targets. All 3D assets—i.e., the conditioning point cloud, ostium prefix mesh, and aneurysm mesh—are jointly rotated around the z-axis to maintain spatial alignment. Additionally, mild simplex noise is applied exclusively to the aneurysm submesh to encourage robustness to local geometric variations. Ostium vertices are explicitly excluded from this perturbation, and a smooth transition is enforced in the surrounding region to avoid discontinuities at the vessel–aneurysm interface.

Training is performed with a batch size of 12 per GPU for up to 400 epochs. We employ the AdamW optimizer with weight decay $0.1$ and mixed-precision (FP16) training. A cosine learning rate schedule with linear warm-up is used, starting from $1 \times 10^{-6}$. The base and final learning rates are set to $1 \times 10^{-5}$ and $6 \times 10^{-6}$ during base training, and reduced to $1 \times 10^{-6}$ and $6 \times 10^{-7}$ during fine-tuning. The maximum number of triangles is limited to $1600$, and the coordinate discretization resolution is set to $256$ (increased from $128$ in the original MeshAnything V2). Training optimizes a masked next-token cross-entropy loss, where prefix tokens (i.e., ostium tokens) are excluded via loss masking, enforcing learning only on the continuation (aneurysm generation). During evaluation, we report the same loss in addition to CD and deviations in independently computed morphological parameters.

\subsection{Experimental Setting}
\label{sec:SM_SynVA-A2_Experiments}
We conduct three experiments under a unified training protocol. Due to the increased mesh discretization resolution~(256 instead of 128), pretrained MeshAnything~V2 weights are not compatible. Moreover, they are likely suboptimal for our setting, as we extend the task toward extrapolative mesh generation from incomplete point clouds, which differs fundamentally from the original full-shape reconstruction objective. Consequently, all models are trained from scratch.

Evaluation focuses on generalization to real anatomical geometries and is therefore always performed on the real-data test set. Since no separate validation split is available, validation is carried out on the respective training domain. However, to keep the process computationally tractable, validation is restricted to subsets and does not involve full mesh generation for all test samples. Final results are obtained independently by generating aneurysms for the complete test set using the selected model.

The experimental design covers three complementary regimes: (i) training purely on real data, (ii) training purely on synthetic data generated by the SynVA procedural model with cross-domain evaluation on real data, and (iii) synthetic pretraining followed by fine-tuning on real data using a reduced learning rate. 

Model selection is based on a combination of validation loss, morphological parameter deviations~(see \cref{sec:SM_MorphologicParameters}), and generation success rate. This setup enables a controlled analysis of real-only learning, synthetic-only generalization, and synthetic-to-real transfer, while maintaining a consistent and computationally efficient validation strategy.

To isolate the effect of point-cloud conditioning, we conduct an ablation in which both training (on synthetic data) and evaluation (on real data) are performed without point-cloud input.

\subsection{Additional Results}
\label{sec:SM_SynVA-A2_Results}
The results of the three experiments are summarized in this section. The configuration trained exclusively on real data is discarded, as the limited amount of available real samples is insufficient to effectively train the model, resulting in no valid mesh generations on the test set.

\Cref{tab:SM_SynVA-A2-ResultsShape} reports the mean absolute differences of independent morphological parameters with respect to the ground truth, as well as the CD on the test set. For a fair comparison, only the intersection of successfully generated samples is considered ($n=46$).

\begin{figure}[htb]
  \centering
  \includegraphics[width=\linewidth]{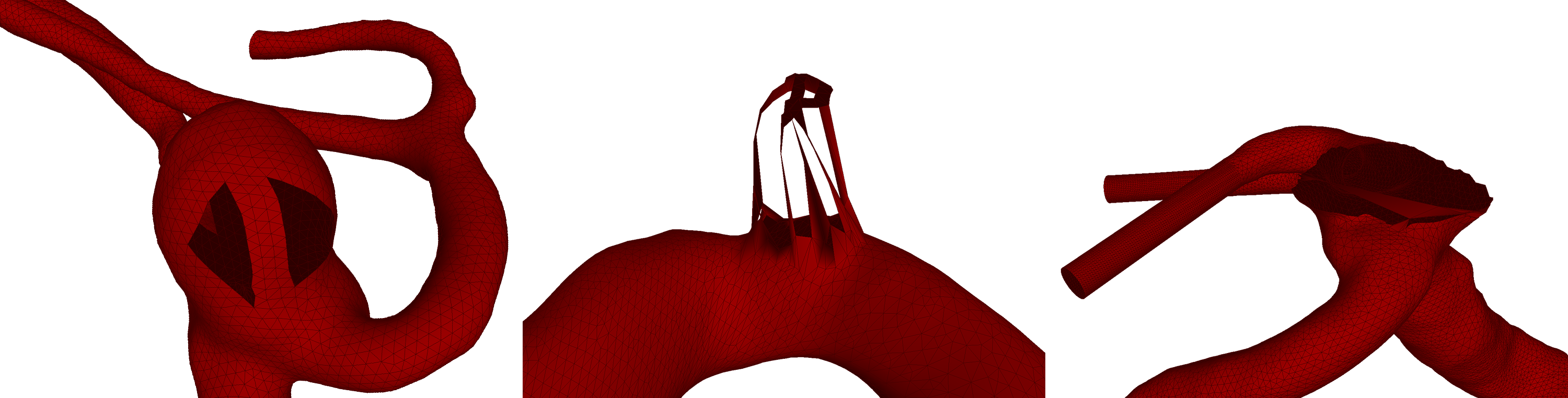}
  \vspace{-0.2cm}
  \caption{Qualitative examples of failure cases in ostium-conditioned aneurysm generation with SynVA-A2.}
  \label{fig:SM_SynVA-A2-FailureCases}
\end{figure}

Fine-tuning on real data consistently improves performance across all evaluated metrics. The validation loss decreases from $0.5592$ (synthetic-only training) to $0.4636$. While the success rate is slightly lower for the fine-tuned model ($66\,\%$) compared to the synthetic-only model ($68\,\%$), it produces more realistic geometries both quantitatively and qualitatively. Consequently, the fine-tuned model is denoted as SynVA-A2 and used for subsequent analyses. Representative failure cases are shown in \cref{fig:SM_SynVA-A2-FailureCases}, including remaining holes and, in rare cases, incomplete or only coarsely approximated aneurysm geometries.

\begin{table}[ht]
\centering
\caption{Comparison of SynVA-A2 training configurations based on the mean absolute differences of independent morphological aneurysm parameters and the CD between generated and ground truth geometries.}
\label{tab:SM_SynVA-A2-ResultsShape}
\scalebox{0.8}{
\begin{tabular}{lcc}
\toprule
Mean Absolute Difference  & Syn ($n=46$) & SynReal ($n=46$) \\
\midrule
$A_\mathrm{A}$ & 0.3858 $\pm$ 0.5941 & \textbf{0.1998 $\pm$ 0.3405} \\
$A_\mathrm{CH}$ & 0.3835 $\pm$ 0.6023 & \textbf{0.2040 $\pm$ 0.3386} \\
$D_{\max}$ & 0.1563 $\pm$ 0.1587 & \textbf{0.1004 $\pm$ 0.1071} \\
$H_{\max}$ & 0.1827 $\pm$ 0.1668 & \textbf{0.1059 $\pm$ 0.1067} \\
$H_{\mathrm{ortho}}$ & 0.1786 $\pm$ 0.1591 & \textbf{0.1032 $\pm$ 0.1020} \\
$V_\mathrm{A}$ & 0.0671 $\pm$ 0.1253 & \textbf{0.0362 $\pm$ 0.0662} \\
$V_\mathrm{CH}$ & 0.0604 $\pm$ 0.1236 & \textbf{0.0267 $\pm$ 0.0584} \\
$W_{\max}$ & 0.1080 $\pm$ 0.1307 & \textbf{0.0779 $\pm$ 0.1042} \\
$W_{\mathrm{ortho}}$ & 0.1001 $\pm$ 0.1273 & \textbf{0.0769 $\pm$ 0.0969} \\
\midrule
Chamfer Loss & 0.1370 $\pm$ 0.1095 & \textbf{0.1054 $\pm$ 0.0739} \\
\bottomrule
\end{tabular}
}
\end{table}

\begin{figure}[htb]
  \centering
  \includegraphics[width=\linewidth]{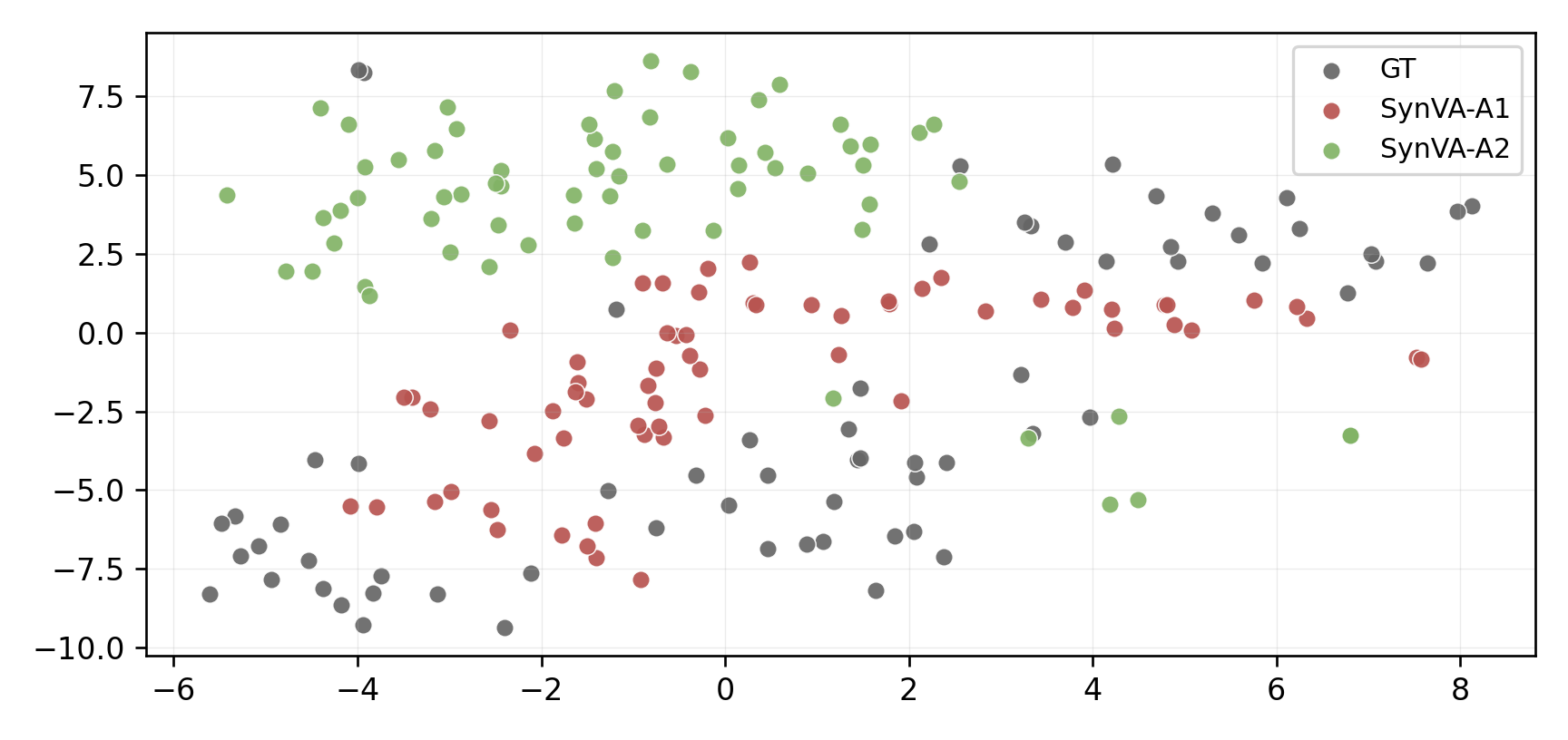}
  \vspace{-0.6cm}
  \caption{t-SNE embeddings of ground-truth and synthetic aneurysm meshes generated by SynVA-A1 and SynVA-A2.}
  \label{fig:SM_SynVA-A-TSNE}
\end{figure}

The ablation yields a 0\,\% success rate on the real-data test set, as no aneurysm could be generated successfully. This result strongly supports the necessity of local point-cloud conditioning for autoregressive extrapolative generation in SynVA-A2 (adapted from MeshAnything V2~\cite{chen2025meshanythingv2}). The t-SNE plot (\cref{fig:SM_SynVA-A-TSNE}) illustrates the distribution of generated and real aneurysms in latent space. While both models~(A1 and A2) exhibit overlap with the ground-truth distribution, SynVA-A1 aligns more closely with real geometries than SynVA-A2.

\Cref{tab:SM_SynVA-A-ExpertAgreement,tab:SM_SynVA-A-ExpertWilcoxon} supplement the expert evaluation from \cref{sec:Exp-Results_SynVA-AP} by showing the agreement between the two neuroradiologists as well as pairwise significance tests between Ground Truth, SynVA-A1, and SynVA-A2.

\begin{table}[htb]
\centering
\caption{Inter-rater agreement between the two neuroradiologists per condition ($n=40$ cases each) and pooled over all conditions~($n=120$). Exact agr.: fraction of cases with identical ratings.}
\label{tab:SM_SynVA-A-ExpertAgreement}
\scalebox{0.75}{
\begin{tabular}{lcccc}
\toprule
Condition & ICC(2,1) & Cohen's $\kappa$ & Spearman $\rho$ & Exact agr. \\
\midrule
GT       & $0.242$  & $0.175$  & $0.301$ ($p=0.059$)  & 62\,\% \\
SynVA-A1 & $-0.021$ & $-0.003$ & $-0.038$ ($p=0.816$) & 25\,\% \\
SynVA-A2 & $0.159$  & $0.174$  & $0.196$ ($p=0.226$)  & 45\,\% \\
\midrule
All ($n=120$) & $0.127$ & $0.153$ & $0.189$ ($p=0.038$) & 44\,\% \\
\bottomrule
\end{tabular}}
\end{table}

\begin{table}[htb]
\centering
\caption{Paired Wilcoxon signed-rank tests on the per-case expert-averaged ratings ($n=40$), Holm-corrected across the three comparisons. $\Delta$ mean: difference of mean ratings on the 0--5 scale.}
\label{tab:SM_SynVA-A-ExpertWilcoxon}
\scalebox{0.75}{
\begin{tabular}{lcccc}
\toprule
Comparison & $\Delta$ mean & $p$ (raw) & $p$ (Holm) & Significant \\
\midrule
GT vs.\ SynVA-A1       & $+0.49$ & $0.0060$ & $0.018$ & yes \\
GT vs.\ SynVA-A2       & $+0.30$ & $0.0141$ & $0.028$ & yes \\
SynVA-A1 vs.\ SynVA-A2 & $-0.19$ & $0.2143$ & $0.214$ & no \\
\bottomrule
\end{tabular}}
\end{table}

Inter-rater agreement is low ($\mathrm{ICC}=0.13$, $\kappa=0.15$ across all 120 ratings). Given that both experts predominantly assign high scores, the range of values used is severely limited, which systematically reduces measures of agreement such as $\mathrm{ICC}$ and $\kappa$. In 77\,\% of cases, the two ratings differ by no more than one point. Ground-truth aneurysms are rated as significantly more realistic than SynVA-A1 ($p=0.018$) and SynVA-A2 ($p=0.028$); however, the absolute differences are small, ranging from 0.3 to 0.5 points. There is no significant difference between SynVA-A1 and SynVA-A2 ($p=0.214$), which is consistent with the qualitative feedback that multiple aneurysm shapes may be plausible for the same anatomy.

\section{SynVA-P1: Procedural Model}
\label{sec:SM_SynVA-P1}
\begin{figure*}[htb]
  \centering
  \includegraphics[width=0.8\textwidth]{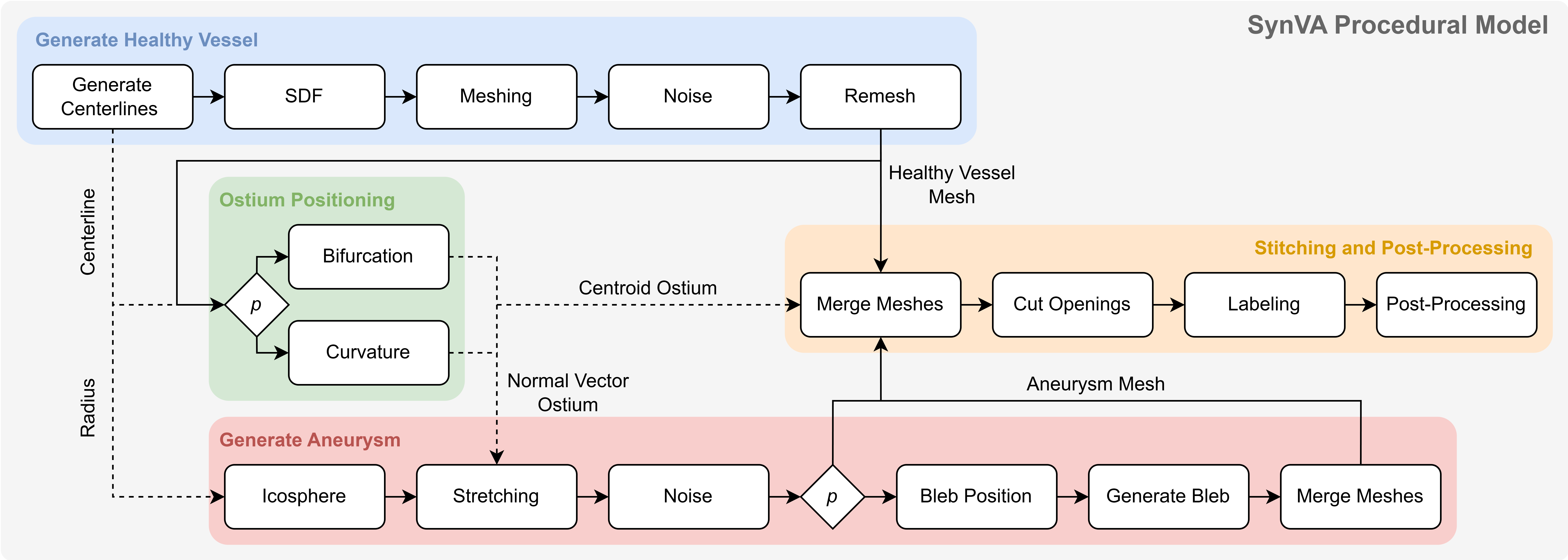}
  \caption{The SynVA-P1 procedural model for generating synthetic vessels with aneurysms. The main components are the healthy vessel~(blue) and aneurysm generation (red), which are linked by the ostium selection (green) and the stitching and postprocessing (orange).}
  \label{fig:SM_SynVA-P1}
\end{figure*}

The SynVA-P1 procedural model is designed to generate parameterized meshes of vascular structures. By incorporating biological constraints, statistical measures derived from studies on real patient data, and stochastic variability, the model enables the controlled generation of a large number of diverse yet physiologically plausible vessel geometries. At its current stage, the model is limited to a single bifurcation of a parent vessel into two child vessels, along with the modeling of an aneurysm. Additional underlying assumptions are described in detail in the subsequent sections. \Cref{fig:SM_SynVA-P1} illustrates the overall model and its individual components.

\begin{figure*}[htb]
  \centering
  \includegraphics[width=0.6\textwidth]{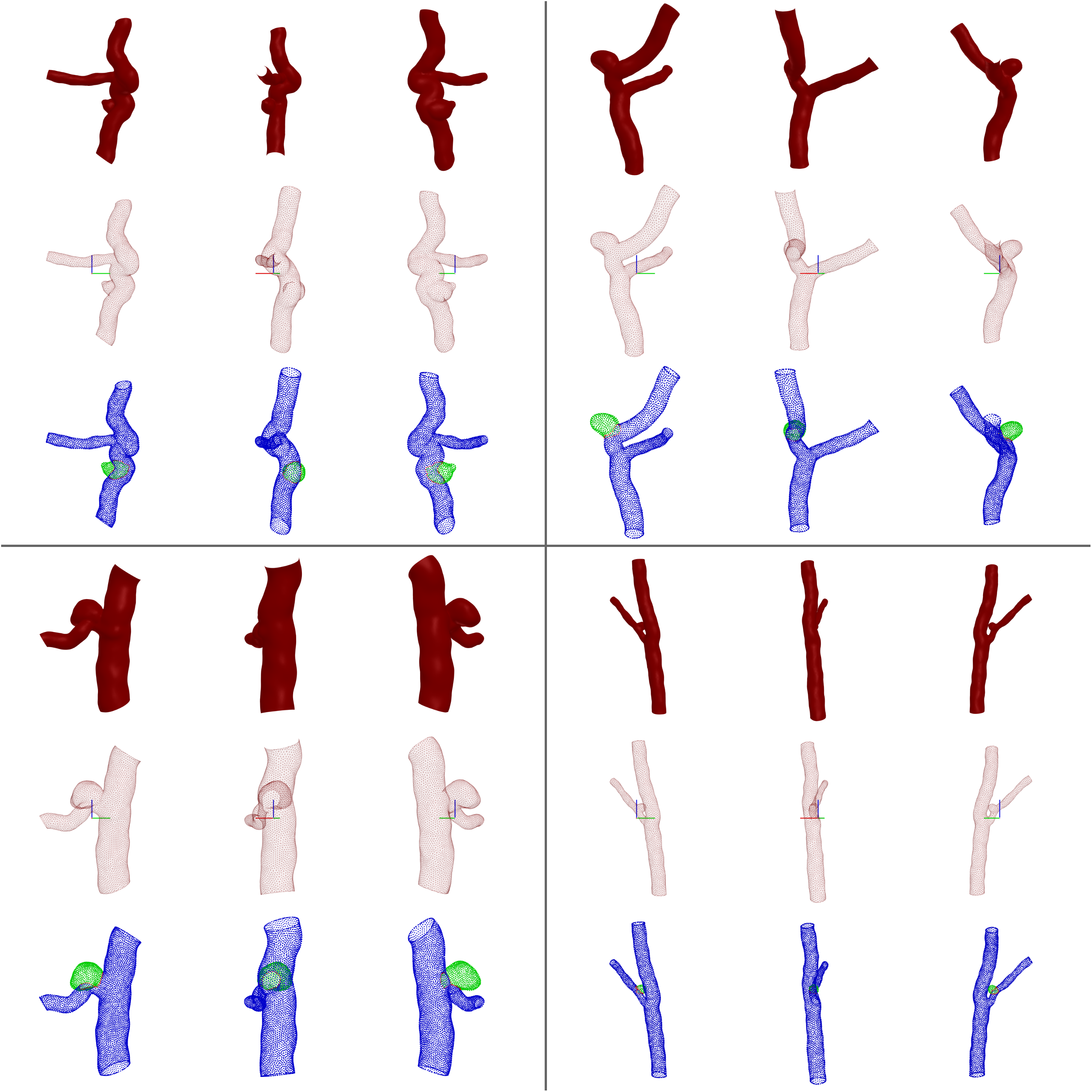}
  \caption{Qualitative examples of the SynVA procedural model. Four instances are visualized under different rotations, including mesh representations, wireframes, and labeled point clouds.}
  \label{fig:SM_SynVA-P1_Results}
\end{figure*}

The first main component is the healthy vessel generation, which constructs the topology of a healthy vessel featuring a bifurcation without an aneurysm (see \cref{sec:SM_SynVA-P1_HealthyVessel}). Subsequently, the centroid of the ostium, defined as the center of the intersection surface between the aneurysm and the healthy vessel, is determined within the ostium selection component (see \cref{sec:SM_SynVA-P1_Ostium}). In addition to the centroid, the corresponding surface normal is computed. The second main component of the SynVA procedural model is the conditional generation of the aneurysm pouch, guided by the previously determined ostium normal vector. A detailed description of the individual processing steps of this component is provided in \cref{sec:SM_SynVA-P1_Aneurysm}. The identified ostium centroid serves as the reference for the stitching process (\cref{sec:SM_SynVA-P1_PostProcessing}), in which both main components (healthy vessel and aneurysm) are merged using appropriate mesh operations. In addition, consistent labeling of the resulting structure as well as the computation of relevant morphological parameters are performed during this process. The model produces the following outputs:
\begin{itemize}
    \item \textbf{Complete mesh:} A unified mesh representation of the vascular structure including the aneurysm, defined by vertices, faces, and associated surface normals~(see \cref{fig:SM_SynVA-P1_Results}).
    \item \textbf{Vertex labels:} An index-based label file assigning each vertex a semantic class: $0$ (healthy vessel), $1$ (aneurysm), and $2$ (ostium).
    \item \textbf{Point cloud:} A labeled point cloud representation of the complete vascular geometry, where class labels are encoded in the color attribute~(see \cref{fig:SM_SynVA-P1_Results}).
    \item \textbf{Sub-point clouds:} Class-specific point clouds obtained by partitioning the full point set according to labels, resulting in separate point clouds for the healthy vessel, aneurysm, and ostium.
    \item \textbf{Submeshes:} Two sub-meshes corresponding to the healthy vessel and the aneurysm, respectively. Both submeshes share vertices at the ostium interface to facilitate consistent stitching and downstream processing.
    \item \textbf{Submesh labels:} Index-based vertex labels for each submesh, following the same class notation as above.
    \item \textbf{Additional outputs:} In addition to geometric representations and labels, the model provides the ostium centroid, the corresponding ostium normal vector, vessel centerlines with associated radii, and a set of derived morphological parameters (see \cref{sec:SM_MorphologicParameters}).
\end{itemize}
Example results from the SynVA-P1 procedural model can be seen in \cref{fig:SM_SynVA-P1_Results}. The model was tested on an NVIDIA GeForce RTX 4090 and a 13th Gen Intel Core i9-13900K CPU.

\subsection{Healthy Vessel Generation}
\label{sec:SM_SynVA-P1_HealthyVessel}
The healthy vessel generation component constructs a closed vascular mesh featuring a single bifurcation and no aneurysms. It serves as the foundational geometry upon which subsequent processing steps are built. In the following, we first describe the underlying model design and then outline the corresponding parameter space exploration.

\subsubsection{Model Design}
\label{sec:SM_SynVA-P1_HealthyVessel_Model}
The healthy vessel generation component consists of five core steps that collectively generate a randomized vascular mesh with a single bifurcation~(see \cref{fig:SM_SynVA-P1}). The individual steps are described in the following.

\textbf{Generate Centerlines}
In this step, parametric centerline splines and their associated radii are defined for the three vessel branches. The bifurcation point is initialized at the origin of the coordinate system. The parent vessel diameter $d_0$ is then sampled uniformly at random from the interval~$[2,\,4]$. Based on $d_0$, the diameters of the two child vessels, $d_1$ and $d_2$, are determined according to Murray’s law,
\begin{equation}
    \label{eq:SM_MurrayLaw}
    d_0^3 = d_1^3 + d_2^3,
\end{equation}
which describes the physiological relationship between vessel diameters at a bifurcation~\cite{murray1926physiological}. To obtain a concrete parameterization, a random factor $a \in [0.1,\,0.9]$ is introduced, which defines the relative contribution of one child vessel with respect to the parent vessel in terms of volumetric flow. This yields
\begin{equation}
    \label{eq:SM_d_1}
    d_1 = \sqrt[3]{a \cdot d_0^3},
\end{equation}
\begin{equation}
    \label{eq:SM_d_2}
    d_2 = \sqrt[3]{d_0^3 - d_1^3}.
\end{equation}

In addition to the vessel diameters, Murray’s law further allows the derivation of the bifurcation angles $\theta_1$ and $\theta_2$. These are computed as
\begin{equation}
    \label{eq:SM_theta1}
    \cos(\theta_1) = \frac{d_0^4 + d_1^4 - \left(d_0^3 - d_1^3\right)^{\frac{4}{3}}}{2 d_0^2 d_1^2},
\end{equation}
\begin{equation}
    \label{eq:SM_theta2}
    \cos(\theta_2) = \frac{d_0^4 + d_2^4 - \left(d_0^3 - d_2^3\right)^{\frac{4}{3}}}{2 d_0^2 d_2^2},
\end{equation}
following established formulations in the literature~\cite{murray1926physiological_angles,lee2010murray,revellin2009extension}. To ensure numerical stability, the arguments of the inverse cosine function are clipped to the interval $[-1,,1]$ prior to applying $\arccos$.

To introduce three-dimensional variability and determine the orientation of the vessel branches, a rotation-based formulation is employed. Specifically, three rotation matrices corresponding to rotations about the principal axes are defined as
\begin{equation}
    \label{eq:SM_R_x}
    \bm{R_x}=\begin{bmatrix}
        1 & 0 & 0 \\
        0 & \cos(tilt_x) & -\sin(tilt_x) \\
        0 & \sin(tilt_x) & \cos(tilt_x) \\
    \end{bmatrix},
\end{equation}
\begin{equation}
    \label{eq:SM_R_y}
    \bm{R_y} = \begin{bmatrix}
        \cos(tilt_y) & 0 & \sin(tilt_y) \\
        0 & 1 & 0 \\
        -\sin(tilt_y) & 0 & \cos(tilt_y)
    \end{bmatrix},
\end{equation}
\begin{equation}
    \label{eq:SM_R_z}
    \bm{R_z} = \begin{bmatrix}
        \cos(rot_z) & -\sin(rot_z) & 0 \\
        \sin(rot_z) & \cos(rot_z) & 0 \\
        0 & 0 & 1
    \end{bmatrix}.
\end{equation}
These matrices are combined into a single rotation matrix
\begin{equation}
    \label{eq:SM_R}
    \bm{R} = \bm{R_z} \cdot \bm{R_y} \cdot \bm{R_x}.
\end{equation}

The tilt angles $tilt_x$ and $tilt_y$ represent rotations around the $x$- and $y$-axes, respectively, and are sampled uniformly~(in radians) from the interval $[-10^\circ,\,10^\circ]$. The in-plane rotation $\text{rot}_z$ is sampled uniformly from $[0,\,2\pi]$. The direction of the parent vessel is defined along the negative $z$-axis as
\begin{equation}
    \label{eq:SM_dir_0}
    \bm{d}_0 = \begin{bmatrix}
        0 & 0& -1
    \end{bmatrix}.
\end{equation}
Given the bifurcation angles $\theta_1$ and $\theta_2$, the initial (unrotated) direction vectors of the child vessels are defined in the $xz$-plane. Applying the combined rotation matrix $\bm{R}$ yields the final direction vectors
\begin{equation}
    \label{eq:SM_dir_1}
    \bm{d}_1 = \bm{R} \cdot
    \begin{bmatrix}
        \sin(\theta_1) & 0 & \cos(\theta_1)
    \end{bmatrix},
\end{equation}
\begin{equation}
    \label{eq:SM_dir_2}
    \bm{d}_2 = \bm{R} \cdot
    \begin{bmatrix}
        -\sin(\theta_2) & 0 & \cos(\theta_2)
    \end{bmatrix}.
\end{equation}

For each direction vector, a randomized centerline curve is generated using a set of control points. The number of control points $n_{\text{control}}$ is sampled uniformly from $[5,\,9]$, the jitter parameter $j$ from~$[0.7,\,1.7]$, and the total curve length $l$ from $[6,\,14]$. These parameter ranges were determined based on qualitative analysis; a systematic parameter space exploration is provided in \cref{sec:SM_SynVA-P1_HealthyVessel_PSE}. 

Starting from the bifurcation point at the origin, the curve is constructed by iteratively advancing along the previously defined direction vector. At each step, the local direction is updated based on the preceding direction, followed by smoothing and the addition of jitter to introduce controlled variability. To ensure anatomically plausible behavior near the junction, the jitter magnitude is modulated during the initial segment of the curve: specifically, for the first half of the control points, the jitter is scaled by a linearly increasing factor in the range $[0.5,\,1]$. This enforces that the vessel initially follows the prescribed direction before gradually introducing curvature.

Given the resulting sequence of control points, a parametric spline is fitted to obtain a smooth centerline representation. The spline degree is defined as $k = 3$, and the curve is uniformly sampled at $n = 120$ points with a smoothing factor of $s = 0.3$. The final output is a smooth centerline for each branch with an associated constant radius.

\textbf{SDF}
We construct an implicit representation of the vascular geometry in the form of a signed distance function~(SDF) based on the generated centerlines and their associated radii. Each centerline is interpreted as a piecewise linear curve (polyline), such that the vessel geometry is modeled as the continuous union of cylindrical segments with spherical end caps. For a query point $\bm{p} \in \mathbb{R}^3$, we first compute the minimal Euclidean distance to the closest centerline segment, ensuring a continuous and sampling-invariant approximation of the vessel structure. Subsequently, the corresponding vessel radius is subtracted, yielding the signed distance to the vessel surface. Formally, this defines an implicit function of the form
\begin{equation}
    \label{eq:SM_SDF}
    \phi(\bm{p}) = \min_k \left( \min_i \left\|\bm{p}-\bm{q_{k,i}}\right\| - r_k \right),
\end{equation}
where $\bm{q}_{k,i}$ denotes the closest point on segment $i$ of centerline $k$, and $\|\cdot\|$ is the Euclidean norm. The combination of multiple vessel branches is achieved via a pointwise minimum over the individual distance fields, corresponding to the union of their respective volumes. The resulting function $\phi$ defines a signed distance field whose zero level set represents the vessel surface and can be directly used for subsequent mesh extraction.

\textbf{Meshing}
The mesh is extracted from the previously computed signed distance field using the Marching Cubes algorithm~\cite{lorensen1987marching}. The volumetric domain was bounded by~$[-20,\,20]$ along each axis, and a grid resolution of 200 was chosen, providing a suitable trade-off between geometric fidelity and computational efficiency. The initial triangular mesh produced by Marching Cubes was subsequently cleaned to ensure geometric and topological consistency. This cleaning procedure comprises: merging vertices in close proximity, removing degenerate and duplicate faces as well as unreferenced vertices, and discarding any infinite or invalid values. Additionally, vertex normals were recomputed to guarantee correct surface orientation.

\textbf{Noise}
Since the current mesh exhibits a uniform radius along each branch, we introduce geometric variability by iteratively superimposing noise at multiple scales and amplitudes relative to the local vessel radius. In our implementation, the number of noise iterations is set to 3. For each iteration, the noise scale increases proportionally to
$1/r$, where $r$ is the reference radius, while the noise amplitude for each scale is sampled randomly from a range between 0 and $r/a_r$. The reduction factor $a_r$ is incremented for successive iterations to progressively decrease the amplitude of higher-frequency noise. The specific numerical ranges were determined empirically to produce visually plausible vessel geometries. For each scale and its associated amplitude, OpenSimplex noise is evaluated at the vertex positions, which are subsequently displaced along their normals. To mitigate artifacts arising from high-frequency displacement, two iterations of Laplacian smoothing are applied after each noise iteration~\cite{vollmer1999improved}. This procedure produces a realistically perturbed mesh while maintaining overall topological consistency, enhancing the biological plausibility of synthetic vascular structures.

\textbf{Remesh}
As the final step in the healthy vessel generation pipeline, the mesh is further refined using isotropic explicit remeshing, which improves the triangle size and aspect ratio across the surface. This procedure is conceptually inspired by \citet{hoppe1993mesh} and ensures a more uniform and well-conditioned discretization suitable for subsequent processing.

The output of this fifth step is a closed, clean surface mesh representing the vascular geometry, including a single bifurcation. Together with the corresponding centerlines and idealized radii, this mesh forms the basis for all downstream components and analyses within the pipeline.

\subsubsection{Parameter Space Exploration}
\label{sec:SM_SynVA-P1_HealthyVessel_PSE}
\begin{figure}[!ht]
  \centering
  \includegraphics[width=\linewidth]{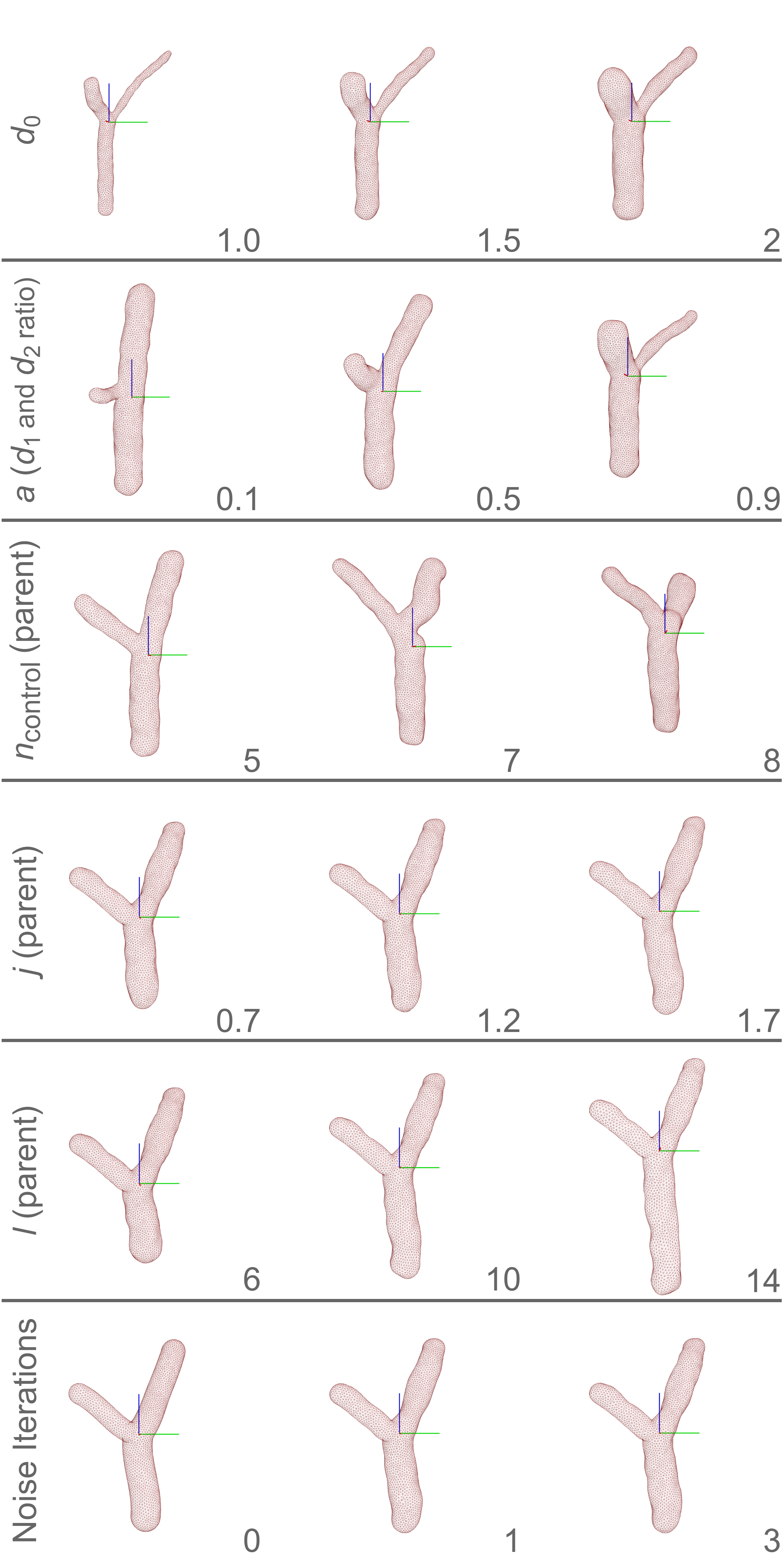}
  \caption{Parameter space exploration of the healthy vessel generation pipeline. Each row corresponds to a different parameter: the parent vessel diameter $d_0$, the branching ratio factor $a$ defining the relative diameters of the child vessels, the number of control points $n_\text{control}$, the jitter amplitude $j$, the vessel length $l$, and the number of noise iterations. For each parameter, three representative values spanning the respective range are shown.}
  \label{fig:SM_SynVA-P1_PSE}
\end{figure}

To assess the effects of individual parameters, a parameter space exploration was conducted. A fixed random seed was used to ensure that only the parameter under investigation varied, while all other parameters were held constant. Representative examples of relevant parameters for the healthy vessel model are shown in \cref{fig:SM_SynVA-P1_PSE}.

It can be observed that increasing the parent vessel diameter $d_0$ leads to a corresponding increase in the diameters of the child vessels. This behavior is governed by Murray’s law, which couples the vessel radii across bifurcations~\cite{murray1926physiological}. The variation of the factor $a$ reveals two key effects. First, for small or large values of $a$, the diameters of the child vessels become increasingly asymmetric, whereas for $a = 0.5$ the diameters are approximately equal. Second, the branching angles are also affected. This follows from the fact that Murray’s law links vessel diameters to optimal branching angles (see \cref{eq:SM_theta1,eq:SM_theta2}). Increasing the number of control points $n_\text{control}$ results in a more curved parent branch, as the underlying spline representation gains additional degrees of freedom. In contrast, increasing the jitter parameter $j$ has only a minor effect on the overall trajectory, primarily introducing small local perturbations that are most noticeable towards the distal end of the parent branch. The parameter $l$ controls the length of the parent branch while leaving the child branch lengths unchanged. Finally, the number of noise iterations directly affects the surface characteristics of the mesh. Without noise, the geometry exhibits a uniform thickness. As the number of noise iterations increases, the surface becomes progressively more irregular due to the superposition of noise functions at increasing scales.

\subsection{Ostium Positioning}
\label{sec:SM_SynVA-P1_Ostium}
Building upon the outputs of the healthy vessel generation, the ostium point is determined (see \cref{fig:SM_SynVA-P1}). This point corresponds to a vertex on the surface mesh. Additionally, a normal vector is estimated, which can be interpreted as the initial growth direction of the aneurysm. With probability $p = 0.5$, the ostium point is located at the bifurcation, i.e., at the junction where the blood flow from the parent vessel splits into the two child vessels. This choice is motivated by the increased hemodynamic stress typically observed at such locations~\cite{schievink1997intracranial,forsting2006intracranial,ringer2018intracranial}. In the remaining cases, the ostium point is determined based on vessel curvature, reflecting the assumption that aneurysms are more likely to occur in highly curved regions than along straight vessel segments.

\textbf{Bifurcation}
For the bifurcation-based selection, a local direction vector of the parent branch at the junction is first estimated by averaging the direction vectors of the first five centerline points. This vector approximates the flow direction towards the bifurcation. Next, the surface curvature of the mesh is computed using the algebraic point set surfaces~(APSS) method on a per-vertex basis~\cite{guennebaud2007algebraic,guennebaud2008dynamic}. Within a predefined search radius around the junction point, all vertices and their associated surface curvatures are evaluated. The selected ostium point is the vertex with minimal curvature that deviates by at most $45^\circ$ from the estimated direction vector. The corresponding normal vector is then defined as the normalized vector between the junction point and the selected surface vertex.

\textbf{Curvature}
For curvature-based selection, the centerline curvature is first computed from the gradients of the spline points and multiplied by the local radius to obtain a scalar score. The point with the maximum score is selected, excluding regions within a safety margin near the ends of the centerline to avoid anatomically implausible placements. Within a predefined search radius relative to the vessel radius, the surface vertices are evaluated based on their curvature. Surface curvature is computed analogously to the bifurcation case. In $90\,\%$ of cases, the vertex with maximal curvature~(i.e., outward-facing) is selected, reflecting regions of increased hemodynamic wall stress, which have been associated with aneurysm formation and are also observed in clinical data. In the remaining $10\,\%$ of cases, however, the vertex with minimal curvature (i.e., inward-facing) is chosen. This accounts for clinically observed scenarios in which aneurysms occur opposite to the primary curvature of the vessel, such as in the internal carotid or ophthalmic artery (e.g.,~\cite{sim2017discrepancy,xu2022incomplete}). The adopted sampling strategy was qualitatively validated against real clinical data to ensure plausible spatial distributions.
The normal vector is then defined as the normalized vector between selected surface vertex (ostium point) and corresponding centerline point.

\begin{figure}[htb]
  \centering
  \includegraphics[width=\linewidth]{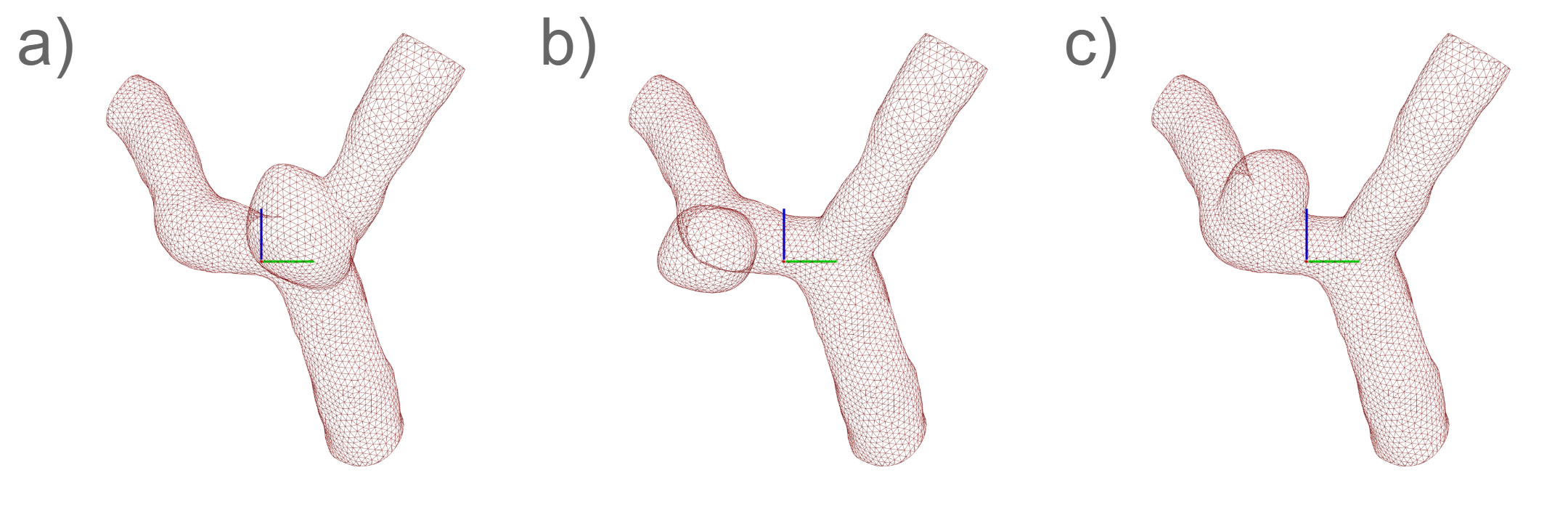}
  \caption{The three ostium location strategies of the SynVA procedural model. a) Placement at the bifurcation, b) placement at a location of maximal surface curvature, and c) placement at a location of minimal surface curvature.}
  \label{fig:SM_SynVA-P1_PSE-Ostium}
\end{figure}

The three cases—bifurcation-based, outward curvature-based, and inward curvature-based selection—are illustrated in \cref{fig:SM_SynVA-P1_PSE-Ostium}.

\subsection{Aneurysm Generation}
\label{sec:SM_SynVA-P1_Aneurysm}
The aneurysm generation component constructs a closed aneurysm mesh conditioned on the ostium normal vector and the local vessel radius. It builds upon the previously defined geometric context to ensure coherent integration with the vascular structure. In the following, we first describe the underlying model design and then outline the corresponding parameter space exploration.

\subsubsection{Model Design}
\label{sec:SM_SynVA-P1_Aneurysm_Model}
The model for generating the aneurysm mesh essentially consists of 6 steps (see \cref{fig:SM_SynVA-P1}). The individual steps are listed below.

\textbf{Icosphere}
In the first step, an icosphere with four subdivisions is generated using a prescribed radius. This radius is defined as the vessel radius at the previously determined ostium location, scaled by a random factor in the range~$[0.8,\,2.0]$. This formulation introduces controlled variability in aneurysm size while maintaining a consistent relationship to the underlying vessel geometry.

\textbf{Stretching}
In the second step, the mesh is anisotropically stretched along the ostium normal vector. The stretching is performed via a directional scaling operation, where each vertex is displaced proportionally to its projection onto the normalized direction vector. The stretch magnitude is controlled by a factor sampled uniformly from the range~$[0,\,0.7]$, resulting in elongation along the normal direction while preserving the orthogonal structure of the mesh.

\textbf{Noise}
Subsequently, the mesh is iteratively perturbed using OpenSimplex noise with varying scales and amplitudes. The implementation follows the noise formulation introduced in the healthy vessel generation and is therefore not described in further detail.

\textbf{Bleb Position}
With probability $p = 0.3$, a bleb, defined as a localized surface bulge, is added to the aneurysm geometry. Bleb formation is clinically relevant, as it is associated with an increased risk of rupture. This design choice is motivated by \citet{salimi2021blebs}, who report the presence of one or multiple blebs in 36\,\% of cases. In the current procedural SynVA model, we restrict the generation to a single bleb. To determine the centroid of the bleb on the aneurysm surface, a ray is cast from the mesh centroid in the direction of the ostium normal vector, and the corresponding surface intersection point is identified. Within a predefined local search region around this intersection, a vertex is randomly selected and defined as the bleb centroid. The associated normal vector of the bleb is then computed as the normalized vector between the mesh centroid and the selected bleb centroid.

\textbf{Generate Bleb}
The bleb itself is generated analogously to the aneurysm base shape. Instead of using the ostium centroid and normal vector, the bleb centroid and its corresponding normal vector are employed to guide the mesh transformation. The bleb radius is scaled by a factor randomly sampled from the range $[0.2,\,0.4]$, introducing controlled variability in bleb size relative to the parent aneurysm.

\textbf{Merge Meshes}
In the sixth and final step of the aneurysm mesh generation, the aneurysm mesh is merged with the bleb mesh. The bleb is first translated to align with the bleb centroid. A mesh cleaning operation is then performed (as previously described), followed by a Boolean union of the meshes, which is accompanied by remeshing, additional cleaning, and Laplacian smoothing. The output of the aneurysm generation pipeline is thus a finalized aneurysm mesh, either with or without an attached bleb.

\subsubsection{Parameter Space Exploration}
\label{sec:SM_SynVA-P1_Aneurysm_PSE}
\begin{figure}[htb]
  \centering
  \includegraphics[width=\linewidth]{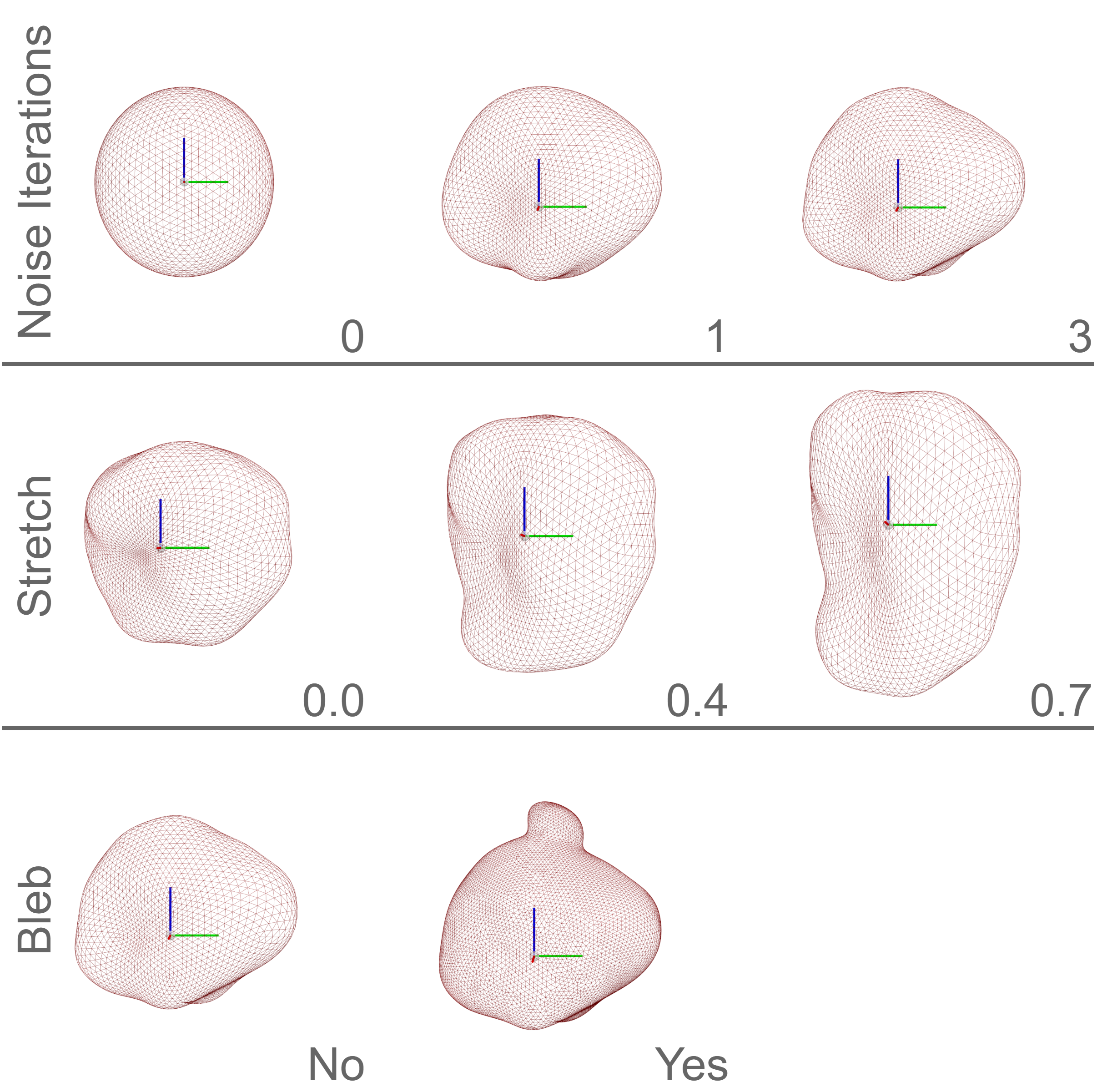}
  \caption{Parameter space exploration of the aneurysm generation pipeline. Each row corresponds to a different parameter: the noise iterations, the stretching factor and presence or absence of a bleb.}
  \label{fig:SM_SynVA-P1_PSE-Aneurysm}
\end{figure}

To assess the effects of individual parameters, a parameter space exploration was conducted. A fixed random seed was used to ensure that only the parameter under investigation varied, while all other parameters were held constant. Representative examples of relevant parameters for the aneurysm model are shown in \cref{fig:SM_SynVA-P1_PSE-Aneurysm}.

It can be observed that increasing the number of noise iterations from 1 to 3 primarily introduces finer surface perturbations. The effect of stretching along the normal vector, here aligned with the blue $z$-axis, is pronounced, resulting in substantial mesh deformation. The bleb is added in the positive direction of the corresponding normal vector.

\subsection{Stitching and Postprocessing}
\label{sec:SM_SynVA-P1_PostProcessing}
The stitching and postprocessing stage constitutes the final pipeline component, combining the two meshes and preparing them for storage in a standardized data structure. This stage consists of four sequential steps, described below.

\textbf{Merge Meshes}
This step is largely analogous to the merging procedure used for combining the aneurysm and bleb meshes. Here, the aneurysm is first translated to align with the ostium centroid. A Boolean union is then performed with the healthy vessel and aneurysm, followed by remeshing and mesh cleaning. The resulting mesh represents the pathological vessel.

\textbf{Cut Openings}
Since the generated mesh remains watertight, it is opened in this step to enable downstream applications such as fluid simulations. To this end, the terminal endpoints of each centerline branch are identified, representing the inlet and outlet locations of the vessel. For each branch, a local cutting operation is performed. Specifically, a tangent vector at the selected endpoint is used to define a cutting plane, whose normal is aligned with the outward vessel direction. A local submesh is then extracted within a radius of $2r$ around the endpoint, and the slicing operation is applied exclusively to this region. This localized processing ensures that only the intended vessel openings are created while preserving surrounding geometry. After slicing, the modified local region is recombined with the unaffected remainder of the mesh. A mesh cleaning step (as described previously) and normal correction are subsequently applied.

\textbf{Labeling}
Although the aneurysm and vessel submeshes are known from the generation process, the preceding merging and remeshing operations no longer preserve any explicit vertex correspondence. Consequently, vertex labels must be recomputed. To this end, a ray-casting-based approach is employed, which proved to be robust in practice. For each vertex of the pathological vessel mesh, a ray is cast from the aneurysm centroid toward the vertex. The first intersection of this ray with the aneurysm surface is determined and compared to the Euclidean distance between the centroid and the respective vertex. If both distances agree within a tolerance threshold, defined relative to the aneurysm size, the vertex is classified as belonging to the aneurysm surface. In cases where the ray-intersection procedure fails, a fallback strategy based on nearest-neighbor queries using a $k$-d tree is applied. Specifically, vertices are labeled as aneurysm if their distance to the aneurysm mesh falls below the same tolerance. Empirically, this method was found to be less accurate than the ray-casting approach. After the initial binary classification into vessel and aneurysm vertices, ostium vertices are identified. A vertex is labeled as ostium if it belongs to the aneurysm class and is connected via mesh edges to at least one vessel vertex, thereby marking the interface between both structures. Finally, a set of consistency checks and corrective operations is applied to ensure label coherence. These include, for example, the removal of isolated misclassified vertices and the enforcement of a consistent ostium topology, such that the interface forms a well-defined boundary between aneurysm and vessel.

\textbf{Postprocessing}
In the final step of the SynVA procedural model (see \cref{fig:SM_SynVA-P1}), additional consistency checks are performed to identify and discard invalid samples. Specifically, it is verified whether a single coherent ostium ring is present and whether the number of ostium vertices exceeds a minimal threshold. If these conditions are not met, the aneurysm is regenerated; after multiple unsuccessful attempts, the healthy vessel geometry is also resampled. Following validation, the aneurysm neck region is locally smoothed around the ostium vertices using Taubin smoothing to ensure a geometrically consistent and smooth transition between vessel and aneurysm. The mesh is then normalized to the range $[-1,\,1]$, and a labeled point cloud is extracted from the mesh vertices. Additionally, separate sub-point clouds are generated for each label class. Subsequently, the ostium centroid and its associated normal vector are recomputed based on the ostium vertices. This is achieved by estimating the center of mass and fitting a local plane to the ostium region. This recomputation is necessary, as the irregular geometry of the aneurysm can introduce deviations from the initial ostium estimation. Finally, separate submeshes for the vessel and aneurysm are extracted, and a set of morphological parameters, described in \cref{sec:SM_MorphologicParameters}, is computed. All outputs are then stored in a standardized data structure for downstream processing.

\section{Dataset}
\label{sec:SM_Dataset}
\begin{figure*}[ht]
  \centering
  \includegraphics[width=0.65\textwidth]{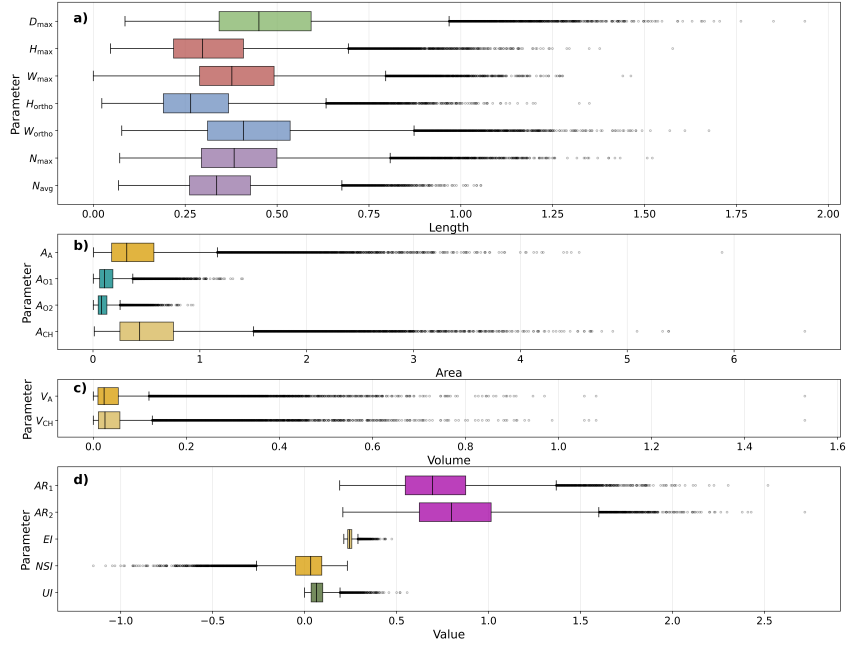}
  \vspace{-2mm}
  \caption{Distribution of morphological parameters across the SynVA procedural dataset, illustrated as boxplots over the sampled parameter space. The parameters are grouped into four categories: a) length-based metrics, b) surface area measures, c) volumetric measures, and d) ratios and derived indices.}  
  \label{fig:SM_Dataset-MorphologyDistribution}
\end{figure*}

The proposed SynVA-P1 procedural model enables the scalable generation of arbitrarily large datasets of labeled vascular meshes, including corresponding point clouds and auxiliary information. Such data can be directly utilized for training and evaluating downstream tasks. In addition to our model code, we provide a pre-generated SynVA dataset comprising 50,000 synthetic vessel samples, allowing immediate use without requiring execution of the procedural pipeline. Representative examples are shown in \cref{fig:SM_SynVA-P1_Results}. To further characterize the dataset, summary statistics of the morphological parameters are reported in \cref{fig:SM_Dataset-MorphologyDistribution}, providing insights into its structural variability and distribution of the aneurysms.

Across all subplots, a concentration of samples within characteristic value ranges can be observed, accompanied by a non-negligible number of outliers. This indicates both consistency and variability in the generated dataset, highlighting the diversity of the procedural model.

For the length-based parameters (\cref{fig:SM_Dataset-MorphologyDistribution}a), aneurysms exhibit consistently smaller heights ($H_\mathrm{max}$ and $H_\mathrm{ortho}$) compared to their corresponding widths ($W_\mathrm{max}$ and $W_\mathrm{ortho}$), suggesting predominantly laterally expanded geometries. In addition, the ostium-related measures ($N_\mathrm{max}$ and $N_\mathrm{avg}$) indicate relatively large neck extents, implying that aneurysms in the dataset tend to have broad necks rather than strongly constricted ones.

Regarding surface areas (\cref{fig:SM_Dataset-MorphologyDistribution}b), the aneurysm area $A_\mathrm{A}$ and its convex hull counterpart $A_\mathrm{CH}$ lie within similar value ranges, with $A_\mathrm{CH}$ being consistently larger due to its enclosing definition. In contrast, the ostium areas ($A_\mathrm{O1}$ and $A_\mathrm{O2}$) are comparatively small and exhibit low variance.

For volumetric measures (\cref{fig:SM_Dataset-MorphologyDistribution}c), only minor differences between $V_\mathrm{A}$ and $V_\mathrm{CH}$ are observed. This can be attributed both to their inherent geometric similarity and to the employed fallback strategy, where $V_\mathrm{CH}$ is used when $V_\mathrm{A}$ cannot be reliably computed due to incomplete closure of the ostium.

Finally, the ratio-based and shape descriptors (\cref{fig:SM_Dataset-MorphologyDistribution}d) show that $AR_1$ and $AR_2$ occupy similar value ranges. The ellipticity index ($EI$) is concentrated within a narrow interval, whereas the non-sphericity index ($NSI$) and undulation index ($UI$) exhibit broader distributions. It should be noted that, similar to the volumetric measures, the fallback strategy for $V_\mathrm{A}$ may introduce bias into $NSI$ and $UI$.

\subsection{Dataset Limitations and Intended Use}
\label{sec:SM_Dataset_Limitations}
The SynVA-P1 dataset is fully synthetic. It is generated by the procedural model described in \cref{sec:SM_SynVA-P1} and contains no patient data or personal information. Its parameter distributions follow physiological principles such as Murray's law and statistical findings from the cited studies, and all labels for vessel, aneurysm, and ostium are produced automatically by the generation pipeline. The dataset therefore reflects the assumptions of the procedural model rather than biases of clinical image acquisition. Each sample contains a single bifurcation, and the dataset does not cover the full variability of patient-level cerebrovascular anatomy, rare aneurysm morphologies, or demographic differences. Models trained on it may generalize less well to underrepresented anatomies, and performance estimates on synthetic data alone may be overly optimistic. Hemodynamic quantities such as wall shear stress are not validated, and some meshes may require additional preprocessing before CFD analysis. The dataset is intended for research, including pretraining, data augmentation, benchmarking, and method development where annotated real data are scarce. It is not intended for diagnosis, treatment planning, rupture-risk estimation, or other clinical decisions without validation on real patient data.
\section{Downstream Tasks}
\label{sec:SM_Downstream}
To determine whether the data generated by SynVA-P1 provides transferable geometric prior knowledge for learning on real data, we examine two complementary downstream tasks: point-wise semantic segmentation of vessels, aneurysms, and ostia as a local task, and the estimation of morphological aneurysm parameters as a global, shape-related task. 
Both tasks follow the same protocol: The models are either trained exclusively on subsets of the real training data or pretrained on SynVA-P1 and then fine-tuned on the same subsets.
In addition, we compare SynVA-P1 with the data-driven generators SynVA-A1 and SynVA-A2 as pretraining sources on the segmentation task (\cref{sec:SM_Downstream_SynDataComparison}).
All models are evaluated on the same held-out real test set of 100 cases, which is used consistently throughout this work and is not used in either pretraining or training. 
The main segmentation experiments are conducted on a single NVIDIA A100 (80\,GB PCIe) GPU and an Intel Xeon Gold 6330 CPU; the additional downstream experiments use a single NVIDIA B200 GPU. The training of SynVA-A2 serves as further, indirect evidence: When trained exclusively on the available real data, the model does not produce valid generations on the real test set; it becomes usable only after pretraining on SynVA-P1.

\subsection{Aneurysm Head Identification}
\label{sec:SM_Downstream_Segmentation}
The task of aneurysm head identification is formulated as a semantic segmentation problem, where each point in the vascular structure is classified as vessel, aneurysm, or ostium. This setup allows us to assess whether synthetic data can support training and generalize to real test cases.

\subsubsection{Architecture}
\label{sec:SM_Downstream_Segmentation_Architecture}
For the aneurysm head identification task, we used a point-wise semantic segmentation model based on Point Transformer V3 (PTv3) \cite{wu2024ptv3}, implemented within the Pointcept framework \cite{pointcept2023}. PTv3 was selected because it is designed for efficient and scalable point cloud processing, making it suitable for vascular geometries represented as 3D point clouds. The downstream task was formulated as a three-class semantic segmentation problem. For each input point, the model predicts one of three anatomical labels: vessel, aneurysm, or ostium. The input to the network consisted of six channels, corresponding to the 3D point coordinates and surface normals.

The PTv3 backbone follows an encoder-decoder structure similar to a U-Net architecture. The encoder progressively downsamples the point representation while increasing the feature dimension from 32 to 512 channels. The decoder then restores the point resolution through unpooling and skip connections, reducing the feature representation back to 64 channels before classification. Each Transformer block uses serialized self-attention, residual connections, pre-normalization, sparse convolution based positional encoding, and a feed-forward MLP. PTv3 handles unordered point clouds by serializing points into structured sequences using different space filling curve orderings. In our configuration, Z order, transposed Z order, Hilbert order, and transposed Hilbert order were used, with order shuffling enabled during training. This allows the model to capture spatial relationships from different ordering perspectives. The final model contained 46.17 million trainable parameters.

\subsubsection{Training Details}
\label{sec:SM_Downstream_Segmentation_Training}
For the downstream task, we use PTv3 across all experiments and vary only the training data regime to assess the impact of synthetic data. Real and synthetic datasets follow the same train/test splits as defined in this work.

Point clouds are preprocessed using grid sampling with a voxel size of $0.01$ and centered to ensure a consistent spatial representation. During training, data augmentation is applied to improve robustness: random point dropout (ratio $0.2$, probability $0.2$), random rotations around the $z$-axis by an angle in~$[-\pi,\,\pi]$~(probability $0.5$), small rotations around $x$ and $y$ by an angle in~$[-\pi/64,\,\pi/64]$ (probability $0.5$ each), random scaling in~$[0.9,\,1.1]$, random flipping (probability $0.5$), and Gaussian jittering with $\sigma=0.005$ (clipped at $0.02$). Validation is performed without stochastic augmentation. At test time, the predictions are aggregated over five deterministic test-time augmentations (scaling by $0.9$, $1.0$, $1.05$, and $1.1$, and a flipped copy).

In total, $11$ experiments are conducted. Five models are trained from scratch on real data using $10\,\%$, $25\,\%$, $50\,\%$, $75\,\%$, and $100\,\%$ of the available training set, corresponding to $67$, $166$, $331$, $497$, and $662$ samples, respectively. For each setting, $10\,\%$ of the selected subset is used for validation and $90\,\%$ for training. In addition, five models are first pretrained on the full synthetic training set~($39{,}940$ samples generated with SynVA-P1, with $10{,}000$ synthetic samples used for testing) and subsequently fine-tuned on the same real-data subsets. Finally, one model is trained exclusively on synthetic data. No joint mixed-batch training is performed, allowing a clear analysis of the individual contributions of synthetic and real data. All models are evaluated on the same real test set of $100$ samples.

Training is performed for $100$ epochs using AdamW. A batch size of $4$ is used in all settings except for the $100\,\%$ real-data experiment, where a batch size of $6$ is feasible. The base learning rate is set to $5 \times 10^{-4}$, with a reduced rate of $5 \times 10^{-5}$ for Transformer block parameters. Weight decay is $0.05$, and a OneCycleLR schedule with cosine annealing is applied. Gradients are clipped at a maximum norm of $1.0$, and training uses bfloat16 mixed precision. The PTv3 backbone employs stochastic depth with $\text{drop\_path}=0.3$.

The loss function combines cross-entropy and Lovász loss with equal weighting, using an index ignoring of $-1$. This setup balances point-wise classification accuracy and segmentation overlap, which is particularly important for smaller structures such as the ostium. Evaluation is performed using mean Intersection over Union~(mIoU), mean class accuracy~(mAcc), overall accuracy~(allAcc), and per-class IoU and accuracy for vessel, aneurysm, and ostium.

\subsubsection{Results}
\label{sec:SM_Downstream_Segmentation_Results}
The quantitative results of this downstream task, used to assess the usefulness of the procedurally generated SynVA-P1 dataset (49,940 samples), are summarized in \cref{tab:SM_Downstream_Segmentation} across all 11 training configurations.

\begin{table*}[htb]
\centering
\caption{Quantitative evaluation of point-wise semantic segmentation performance for aneurysm head identification on the held-out real test set. Metrics are reported in percent.}
\label{tab:SM_Downstream_Segmentation}
\scalebox{0.9}{
\begin{tabular}{lcccccc}
\toprule
Training data & mIoU $\uparrow$ & mAcc $\uparrow$ & allAcc $\uparrow$ & Vessel IoU $\uparrow$ & Aneurysm IoU $\uparrow$ & Ostium IoU $\uparrow$ \\
\midrule
10\,\% real data & 50.41 & 55.52 & 96.72 & 96.73 & 51.35 & 3.16 \\
25\,\% real data & 60.66 & 73.33 & 98.21 & 98.51 & 77.49 & 5.97 \\
50\,\% real data & 63.79 & 80.06 & 98.61 & 98.89 & 84.31 & 8.17 \\
75\,\% real data & 64.13 & 83.22 & 98.61 & 98.96 & 84.44 & 9.00 \\
100\,\% real data & 64.84 & 84.95 & 98.66 & 98.98 & 86.22 & 9.33 \\
\midrule
Synthetic pretraining + 10\,\% real data & 63.88 & 78.35 & 98.67 & 98.90 & 82.94 & 9.80 \\
Synthetic pretraining + 25\,\% real data & 66.42 & 82.23 & 98.99 & 99.18 & 88.58 & 11.49 \\
Synthetic pretraining + 50\,\% real data & \textbf{67.11} & 86.10 & \textbf{99.03} & \textbf{99.29} & \textbf{89.24} & 12.82 \\
Synthetic pretraining + 75\,\% real data & 66.50 & 85.52 & 98.93 & 99.17 & 88.81 & 11.52 \\
Synthetic pretraining + 100\,\% real data & 66.93 & \textbf{87.86} & 98.95 & 99.20 & 88.45 & \textbf{13.13} \\
\midrule
100\,\% synthetic data & 36.78 & 42.09 & 92.24 & 92.15 & 17.80 & 0.38 \\
\bottomrule
\end{tabular}%
}
\end{table*}

The results indicate that synthetic pretraining consistently improves downstream performance across all real-data fractions. For the 10\,\% real-data setting, the real-only model achieves an mIoU of 50.41, whereas synthetic pretraining followed by fine-tuning on the same fraction increases mIoU to 63.88. Similar gains are observed for 25\,\%, 50\,\%, 75\,\%, and 100\,\% real data, indicating that the synthetic dataset provides useful geometric priors, especially in low-data regimes.

The best overall mIoU is obtained with synthetic pretraining followed by fine-tuning on 50\,\% real data, reaching 67.11. This setting also yields the highest aneurysm IoU of 89.24. The highest ostium IoU is achieved with synthetic pretraining followed by fine-tuning on 100\,\% real data, reaching 13.13. Notably, increasing the amount of real fine-tuning data beyond 50\,\% does not further improve mIoU, suggesting diminishing returns for this task.

By contrast, the model trained only on synthetic data performs substantially worse on the real test set, with an mIoU of 36.78, an aneurysm IoU of 17.80, and an ostium IoU of 0.38. This is expected given the domain gap, since SynVA-P1 currently generates simplified single-bifurcation anatomies, whereas the real data contain more complex vascular trees with multiple branches. Overall, the results show that synthetic data alone is not sufficient for generalization, but serves as an effective pretraining source.

\subsection{Morphological Parameter Estimation}
\label{sec:SM_Downstream_Regression}
Morphological parameters such as size, neck width, and aspect ratio are established features used in rupture risk assessment~\cite{dhar2008morphology,mocco2018aneurysm,niemann2018rupture}. However, their computation requires an aneurysm that is clearly separated from the parent vessel and has a defined ostium, which is usually determined manually or semi-automatically when working with real-world data~\cite{saalfeld2018semiautomatic}. 
We therefore formulate the task as a direct regression of 13 morphological parameters from the unsegmented point cloud of the pathological vessel, without access to aneurysm or ostium labels. The parameters calculated on the annotated meshes serve as a reference. 
The 13 targets are aneurysm surface area, two variants of ostium area, maximum diameter, maximum height and width, orthogonal height and width, maximum and average neck diameter, two aspect ratios, and convex hull area.
We exclude volume-based descriptors (aneurysm volume, convex hull volume, NSI, UI, and EI), because the sign of the aneurysm volume depends on the face orientation, which differs between the real meshes.
This task supplements point-based segmentation with a global target metric and thus tests whether synthetic pretraining transfers knowledge about the overall geometry of the aneurysm beyond local label information.

\subsubsection{Architecture}
\label{sec:SM_Downstream_Regression_Architecture}
We use a simplified PointNet~\cite{qi2017pointnet} without the input and feature transformation networks. The network takes $2{,}048$ points sampled from the vessel point cloud and uses only their 3D coordinates, without surface normals or class labels. 
Before passing the points to the network, we center each cloud at its centroid and scale it by its maximum absolute coordinate. 
A shared per-point MLP, implemented as $1\times1$ convolutions with ReLU activations, maps the coordinates to feature representations with 64, 128, 256, and 512 channels. 
Global max pooling then aggregates these features into a 512-dimensional shape descriptor.
A regression head with 256 and 128 hidden units and ReLU activations predicts the 13 parameters, with dropout of $0.2$ applied after the first hidden layer. 

\subsubsection{Training Details}
\label{sec:SM_Downstream_Training}
For the downstream task, we use PointNet across all experiments and vary only the training data regime to assess the impact of synthetic data. Real and synthetic datasets follow the same train/test splits as defined in this work. We randomly permute the $662$ real training cases using a fixed seed and select the first $66$, $166$, $331$, $496$, or $662$ cases, corresponding to 10\%, 25\%, 50\%, 75\,\%, and 100\% of the training set, respectively.

As in the segmentation task, we reserve a validation split at every training stage for model selection. Pretraining runs for up to $100$ epochs on the 39{,}940 SynVA-P1 samples, with 10\,\% $(3{,}994)$ held out for validation. This stage uses no real data. For training on the real data, we reserve 10\,\% of the selected real subset for validation, ranging from $7$ of $66$ cases at the 10\,\% fraction to $66$ of $662$ cases at the 100\,\% fraction. Both configurations run for up to 200 epochs. Fine-tuning starts from all weights of the selected pretraining checkpoint. After each epoch, we compute the validation mean squared error and retain the checkpoint with the lowest value for evaluation on the test set. The real test cases remain excluded from training and model selection.

We optimize all models with Adam using a learning rate of $10^{-3}$ and a batch size of $32$. The objective is the mean squared error between predictions and z-standardized targets. To keep target scaling consistent across training regimes, we standardize all targets using the mean and standard deviation of the complete real training set. Each sample contains a fixed set of $2{,}048$ points, sampled once, with no data augmentation. For evaluation, we invert the standardization and compute $R^2$ and MAE separately for each parameter on the $100$ real test cases, then average each metric over the $13$ parameters. Because the morphological parameters are computed on normalized meshes, MAE is expressed in normalized units rather than millimeters.

We train each configuration with five seeds ($0$--$4$), each with its own pretraining run, subset permutation, and initialization. In addition to the ten configurations, we evaluate each pretrained model directly on the real test set using its selected pretraining checkpoint, without fine-tuning~(synthetic only). All runs use a single NVIDIA B200 GPU.

\subsubsection{Results}
\label{sec:SM_Downstream_Regression_Results}
The quantitative results of the estimation of morphological parameters are summarized in \cref{tab:SM_Downstream_Regression} for all eleven training configurations. $R^2$ and MAE are averaged across the 13 parameters and reported as the mean $\pm$ standard deviation across five seeds.

\begin{table}[htb]
\centering
\caption{Quantitative evaluation of morphological parameter estimation on the held-out real test set. $R^2$ and MAE are averaged over the 13 morphological parameters; MAE in normalized mesh units. Values are shown as mean $\pm$ standard deviation~(5 seeds).}
\label{tab:SM_Downstream_Regression}
\scalebox{0.73}{
\begin{tabular}{lcc}
\toprule
Training data & $R^2$ $\uparrow$ & MAE $\downarrow$ \\
\midrule
10\,\% real data & $-0.074 \pm 0.114$ & $0.249 \pm 0.006$ \\
25\,\% real data & $0.112 \pm 0.054$ & $0.225 \pm 0.011$ \\
50\,\% real data & $0.101 \pm 0.086$ & $0.217 \pm 0.011$ \\
75\,\% real data & $0.341 \pm 0.068$ & $0.189 \pm 0.010$ \\
100\,\% real data & $0.423 \pm 0.047$ & $0.178 \pm 0.008$ \\
\midrule
Synthetic pretraining + 10\,\% real data & $0.451 \pm 0.038$ & $0.177 \pm 0.007$ \\
Synthetic pretraining + 25\,\% real data & $0.487 \pm 0.045$ & $0.172 \pm 0.007$ \\
Synthetic pretraining + 50\,\% real data & $0.505 \pm 0.028$ & $0.169 \pm 0.005$ \\
Synthetic pretraining + 75\,\% real data & $0.545 \pm 0.037$ & $0.161 \pm 0.006$ \\
Synthetic pretraining + 100\,\% real data & $\mathbf{0.562 \pm 0.021}$ & $\mathbf{0.160 \pm 0.005}$ \\
\midrule
100\,\% synthetic data & $0.347 \pm 0.050$ & $0.204 \pm 0.010$ \\
\bottomrule
\end{tabular}}
\end{table}

Synthetic pretraining improves the estimation for all fractions of real data. The effect is most pronounced in the scenario with the least amount of data. With 10\,\% of the real training data, the model trained exclusively on real data achieves a negative $R^2$ of $-0.074$. Its predictions are thus, on average, worse than a constant prediction of the mean. After pretraining on SynVA-P1 and fine-tuning on the same proportion, $R^2$ rises to $0.451$, and the MAE decreases by 29\,\% from $0.249$ to $0.177$. At the same time, variation across seeds decreases substantially (standard deviation of $R^2$: $0.114$ versus $0.038$), stabilizing the training. 

With pretraining and only 10\,\% or 25\,\% of real data, the results ($R^2 = 0.451$ and $0.487$, respectively) already reach and on average exceed the model trained exclusively on all real data ($0.423$). 
All pretrained models outperform this benchmark. 
The best result is achieved by pretraining followed by fine-tuning on 100\,\% real data ($R^2 = 0.562$, MAE $= 0.160$). 
With pretraining, $R^2$ increases monotonically with the amount of real data. Without pretraining, $R^2$ stays at about $0.1$ up to 50\,\% real data and only increases clearly from 75\,\% on.
Unlike segmentation, whose mIoU stops increasing once 50\,\% of the real data is reached, parameter estimation continues to benefit from additional real data up to the full real dataset. 
The benefit of pretraining generally decreases as the amount of real data increases (from $+0.525$ in $R^2$ at 10\,\%), but remains substantial even at 100\,\% ($+0.139$ in $R^2$).
The pretrained model without any real data reaches $R^2 = 0.347$, outperforming real-only training with up to 50\,\% of the real data and matching real-only training at 75\,\% ($0.341$).

Overall, the results show that SynVA-P1-generated data convey not only pointwise label information but also transferable knowledge about global aneurysm geometry. In the low-data regime, they make the task learnable in the first place.

\subsection{Comparison of Synthetic Data Sources}
\label{sec:SM_Downstream_SynDataComparison}
The experiments in \cref{sec:SM_Downstream_Segmentation,sec:SM_Downstream_Regression} show that pretraining on procedurally generated data improves transfer to real data. It remains unclear which type of synthetic data best complements real training data. SynVA provides two fundamentally different sources for this purpose. The procedural generator SynVA-P1 operates entirely without patient-derived geometry. The data-driven generators SynVA-A1 and SynVA-A2 generate aneurysms on real parent vessels. We compare these sources on the segmentation task of \cref{sec:SM_Downstream_Segmentation} at comparable corpus sizes.

\subsubsection{Synthetic Corpora}
\label{sec:SM_Downstream_SynDataComparison_SynCorpora}
To compare the data sources while ensuring comparable data volumes, we create a corpus of approximately 4k valid samples for each generator. 
SynVA-P1 generates its samples independently of real data; here, we use a random subset of 4.0k samples (10\,\% of the P1 training set). 
SynVA-A1 and SynVA-A2 generate aneurysms on existing vessels according to their intended use. Only vessels from the real-world training split serve as parent vessels, ensuring that the test set remains completely separate. 
However, repeated generations on the same parent vessel share the same anatomy and the same ostium. For both generators, we therefore additionally apply ostium jitter: For each generation, the ostium ring is independently scaled by up to $\pm15\,\%$ and tilted by up to $\pm 10^\circ$ about a random axis, and all parameters derived from the ostium are recalculated from the perturbed ring. 
This introduces geometric variation while keeping the conditioning internally consistent.

For SynVA-A1, this results in 4.2k samples without jitter (eight per parent vessel, $530$ of the $662$ training vessels) and 5.3k samples with jitter (eight per parent vessel, all 662 training vessels). 
For SynVA-A2, the failure rate limits the corpus without jitter to 1.4k valid samples, since only 65\,\% of the training vessels develop a valid aneurysm ($427$ of $662$). 
With jitter, this case coverage increases to 99\,\% ($657$ of $662$) because cases that previously failed are successfully generated with a perturbed ostium. The dataset thus reaches a comparable size of 4.0k samples. 
The A1 corpus with jitter is therefore about one third larger than the P1 and A2
corpora.
All samples undergo the postprocessing and validity check of the respective generator (\cref{sec:Methodology_SynVA-A}, \cref{sec:SM_SynVA-A1,sec:SM_SynVA-A2,sec:SM_SynVA-P1}) and carry labels for vessel, aneurysm, and ostium.

\subsubsection{Training Details}
\label{sec:SM_Downstream_SynDataComparison_Training}
Architecture, optimization, augmentation, and evaluation remain consistent with the segmentation protocol described in \cref{sec:SM_Downstream_Segmentation}, except for the batch size: 16 for pretraining and 6 for all fine-tuning and real-only runs, regardless of the real-data fraction. Thus, the models differ solely in their pretraining data. PTv3 is pretrained on the respective corpus and then fine-tuned on 10\,\%, 25\,\%, 50\,\%, 75\,\%, or 100\,\% of the real training data. In addition, for each corpus, one model is trained exclusively on synthetic data. 
All models in this comparison, including the real-only baseline, were trained in a dedicated experiment under identical conditions, with one training run per setting (seed 42), and evaluated on the same real test set of 100 cases. Since this experiment was conducted separately from the main segmentation experiment, its baseline values differ from those in \cref{tab:SM_Downstream_Segmentation}.
As a control for the repeated exposure to the same real vessels, we additionally train real-only models on the real training subset repeated six times, without any synthetic data.

\subsubsection{Results}
\label{sec:SM_Downstream_SynDataComparison_Results}
The results of the comparison are summarized in \cref{tab:SM_Downstream_SynDataComparison}.

\begin{table*}[htb]
\centering
\caption{Segmentation results on the held-out real test set for pretraining on size-matched synthetic corpora from SynVA-P1, SynVA-A1, and SynVA-A2 (with ostium jitter), followed by fine-tuning on the given fraction of real data. All models were trained in the same run under identical conditions. Metrics are reported in percent; best value per column in bold.}
\label{tab:SM_Downstream_SynDataComparison}
\scalebox{0.9}{
\begin{tabular}{lcccccc}
\toprule
Training data & mIoU $\uparrow$ & mAcc $\uparrow$ & allAcc $\uparrow$ & Vessel IoU $\uparrow$ & Aneurysm IoU $\uparrow$ & Ostium IoU $\uparrow$ \\
\midrule
\multicolumn{7}{l}{\textit{Trained from scratch on:}} \\
\quad 10\,\% real data  & 48.38 & 52.77 & 96.44 & 96.44 & 45.10 & 3.61 \\
\quad 25\,\% real data  & 58.72 & 70.62 & 97.90 & 98.16 & 72.25 & 5.75 \\
\quad 50\,\% real data  & 62.44 & 78.77 & 98.36 & 98.65 & 81.21 & 7.47 \\
\quad 75\,\% real data  & 63.26 & 80.67 & 98.54 & 98.89 & 81.98 & 8.90 \\
\quad 100\,\% real data & 64.42 & 84.55 & 98.65 & 99.01 & 84.55 & 9.70 \\
\midrule
\multicolumn{7}{l}{\textit{Pretrained on SynVA-P1 (4.0k), fine-tuned on:}} \\
\quad 10\,\% real data  & 59.17 & 68.87 & 98.06 & 98.23 & 72.79 & 6.50 \\
\quad 25\,\% real data  & 62.80 & 75.97 & 98.49 & 98.67 & 82.65 & 7.07 \\
\quad 50\,\% real data  & 63.03 & 78.68 & 98.56 & 98.89 & 81.82 & 8.38 \\
\quad 75\,\% real data  & 63.87 & 81.64 & 98.62 & 98.94 & 82.78 & 9.90 \\
\quad 100\,\% real data & 65.34 & 85.07 & 98.78 & 99.08 & 86.45 & 10.49 \\
\quad no fine-tuning (synthetic only) & 35.68 & 37.41 & 95.00 & 94.98 & 11.58 & 0.49 \\
\midrule
\multicolumn{7}{l}{\textit{Pretrained on SynVA-A1 with ostium jitter (5.3k), fine-tuned on:}} \\
\quad 10\,\% real data  & 61.60 & 74.63 & 98.35 & 98.57 & 77.23 & 8.99 \\
\quad 25\,\% real data  & 61.89 & 76.67 & 98.35 & 98.58 & 77.14 & 9.96 \\
\quad 50\,\% real data  & 65.73 & 85.33 & 98.84 & 99.12 & 86.64 & 11.44 \\
\quad 75\,\% real data  & 62.75 & 81.47 & 98.33 & 98.53 & 77.68 & 12.04 \\
\quad 100\,\% real data & \textbf{66.53} & \textbf{87.47} & \textbf{98.91} & \textbf{99.18} & \textbf{87.91} & \textbf{12.51} \\
\quad no fine-tuning (synthetic only) & 45.59 & 47.73 & 96.35 & 96.33 & 36.84 & 3.60 \\
\midrule
\multicolumn{7}{l}{\textit{Pretrained on SynVA-A2 with ostium jitter (4.0k), fine-tuned on:}} \\
\quad 10\,\% real data  & 61.94 & 77.61 & 98.35 & 98.63 & 78.14 & 9.06 \\
\quad 25\,\% real data  & 64.33 & 82.64 & 98.64 & 98.92 & 83.35 & 10.72 \\
\quad 50\,\% real data  & 64.18 & 84.16 & 98.60 & 98.93 & 83.85 & 9.76 \\
\quad 75\,\% real data  & 65.36 & 85.95 & 98.74 & 99.02 & 85.94 & 11.11 \\
\quad 100\,\% real data & 66.13 & 87.08 & 98.84 & 99.12 & 87.24 & 12.04 \\
\quad no fine-tuning (synthetic only) & 34.87 & 37.02 & 94.71 & 94.75 & 7.71 & 2.15 \\
\bottomrule
\end{tabular}%
}
\end{table*}

All synthetic sources improve segmentation for all fractions of real data, with one exception: SynVA-A1 with jitter at 75\,\% ($62.75$ versus $63.26$ mIoU).  
The effect is strongest when 10\,\% of the data is real: the mIoU increases from $48.38$ without pretraining to $59.17$ (SynVA-P1), $61.60$ (SynVA-A1), and $61.94$ (SynVA-A2), respectively. At 100\,\%, the results converge ($65.34–66.53$ versus $64.42$). The challenging ostium class benefits particularly. With 10\,\% real data, its IoU increases from $3.61$ to between $6.50$ and $9.06$, and the data-driven models thus reach the level of the model trained on 75\,\% real data ($8.90$).

The repetition control shows that the gain in the low-data regime is not explained by repeated exposure to the real vessels: with 10\,\% real data, the sixfold repetition reaches $56.21$ mIoU, below all three synthetic sources ($59.17$--$61.94$). At 75\,\% and 100\,\%, however, it reaches $66.68$ and $66.67$ mIoU and thus exceeds all synthetic sources; at these fractions, the remaining gains cannot be attributed to the synthetic data.

The data-driven generators achieve the highest transfer performance in most settings. SynVA-A1 also exhibits the smallest domain gap: when trained exclusively on synthetic data, it achieves $45.59$ mIoU, compared to $35.68$ (P1) and $34.87$ (A2).
In this experiment, pretraining on the full P1 corpus (39.9k samples) yields $58.58$ mIoU at 10\,\% real data, with no advantage over the size-matched subset ($59.17$). This comparison therefore provides no evidence for a benefit from larger P1 corpora.

Since each setting has been trained once, we base our conclusions on the large, consistent effects. Additional seeds for the 75\,\% setting illustrate the run-to-run variation: for SynVA-A1 with jitter, four seeds range from $62.75$ to $65.88$ mIoU, and for SynVA-A1 without jitter from $65.15$ to $66.37$. 
We therefore do not interpret differences of a few points at high real-data fractions as a ranking.
Overall, the two types of synthetic data complement each other. 
The data-driven generators produce pathologies that are closely tied to real-world vascular anatomy. SynVA-P1 does not require patient-derived geometry and scales to arbitrary corpus sizes.

\end{document}